\documentclass[11pt]{article} 
\usepackage{times} 
\usepackage[margin=1in]{geometry} 
\usepackage{booktabs}
\usepackage{graphicx}
\usepackage{amsmath}
\usepackage{amssymb}
\usepackage{subcaption}
\usepackage[hidelinks]{hyperref}
\usepackage{array}
\usepackage[numbers,sort&compress]{natbib}
\usepackage{xcolor}
\usepackage{listings}
\newcommand{\piRLinf}{\pi_{\mathrm{RLinf}}}
\title{\texorpdfstring{%
\begin{tabular}{@{}c@{\hspace{0.9em}}c@{}}
\raisebox{-0.4\height}{\includegraphics[height=3em]{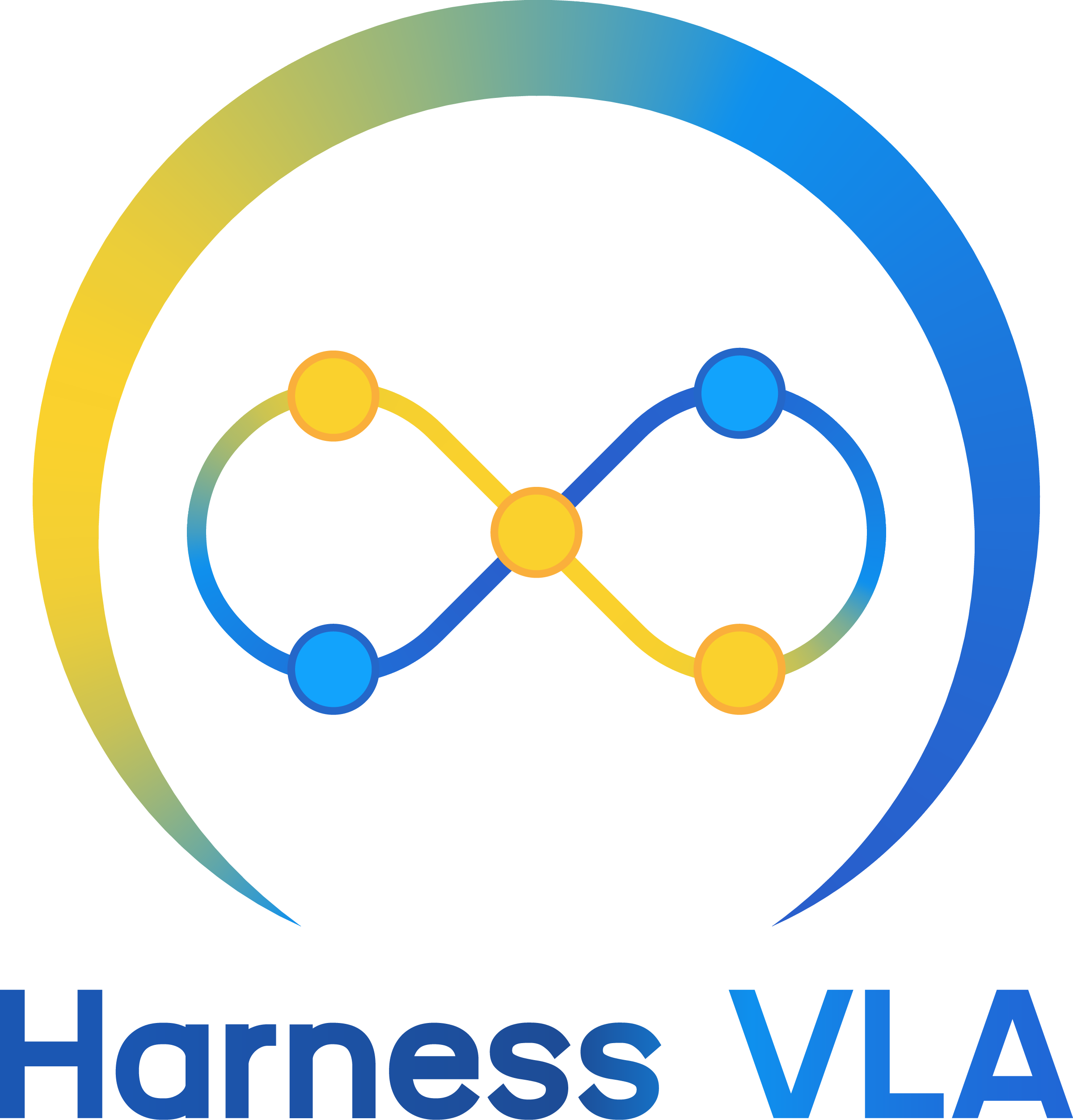}} &
\begin{minipage}{0.80\textwidth}
\centering
Harness VLA: Steering Frozen VLAs into Reliable\\
Manipulation Primitives via Memory-Guided Agents
\end{minipage}
\end{tabular}%
}{Harness VLA: Steering Frozen VLAs into Reliable Manipulation Primitives via Memory-Guided Agents}}

\author{
\begin{tabular}{@{}c@{}}
Yixian Zhang$^{1,*}$ \enspace Huanming Zhang$^{1,*}$ \enspace Feng Gao$^2$ \enspace Xiao Li$^3$ \enspace Zhihao Liu$^4$ \enspace
Yi Nie$^{1,2}$  \\ Chunyang Zhu$^5$ \enspace Jiaxing Qiu$^5$ \enspace Yuchen Yan$^5$ \enspace Jiyuan Liu$^7$ \enspace Wenhao Tang$^1$ \enspace Jiaji Rao$^1$   \\
Zhengru Fang$^6$ \enspace Changxu Wei$^1$ \enspace Yu Wang$^1$ \enspace Wenbo Ding$^{1,\dagger}$ \enspace Chao Yu$^{1,\dagger}$ \\[0.45em]
{\small $^1$Tsinghua University \enspace $^2$Striding AI \enspace $^3$Purdue University} \\
{\small $^4$Institute of Automation, Chinese Academy of Sciences \enspace $^5$Infinigence AI} \\
{\small $^6$Hong Kong University of Science and Technology \enspace $^7$Zhongguancun Academy} \\
{\small $^*$Equal contribution. \enspace $^\dagger$Corresponding author} \\[-0.25em]
{\small Website: \url{https://harnessvla.github.io/}} \\
{\small Code: \url{https://github.com/RLinf/RPent}}
\end{tabular}
}

\date{}

\makeatletter
\renewcommand{\@maketitle}{%
  \newpage
  \null
  \vskip 1em%
  \begin{center}%
  \let \footnote \thanks
    {\LARGE \@title \par}%
    {\large
      \lineskip .5em%
      \begin{tabular}[t]{c}%
        \@author
      \end{tabular}\par}%
    \vskip 1em%
    {\large \@date}%
  \end{center}%
  \par
  \vskip 1.5em}
\makeatother

\begin{document}
\maketitle
\vspace{-3em}
\begin{figure}[htbp]
\centering
\includegraphics[width=1\linewidth]{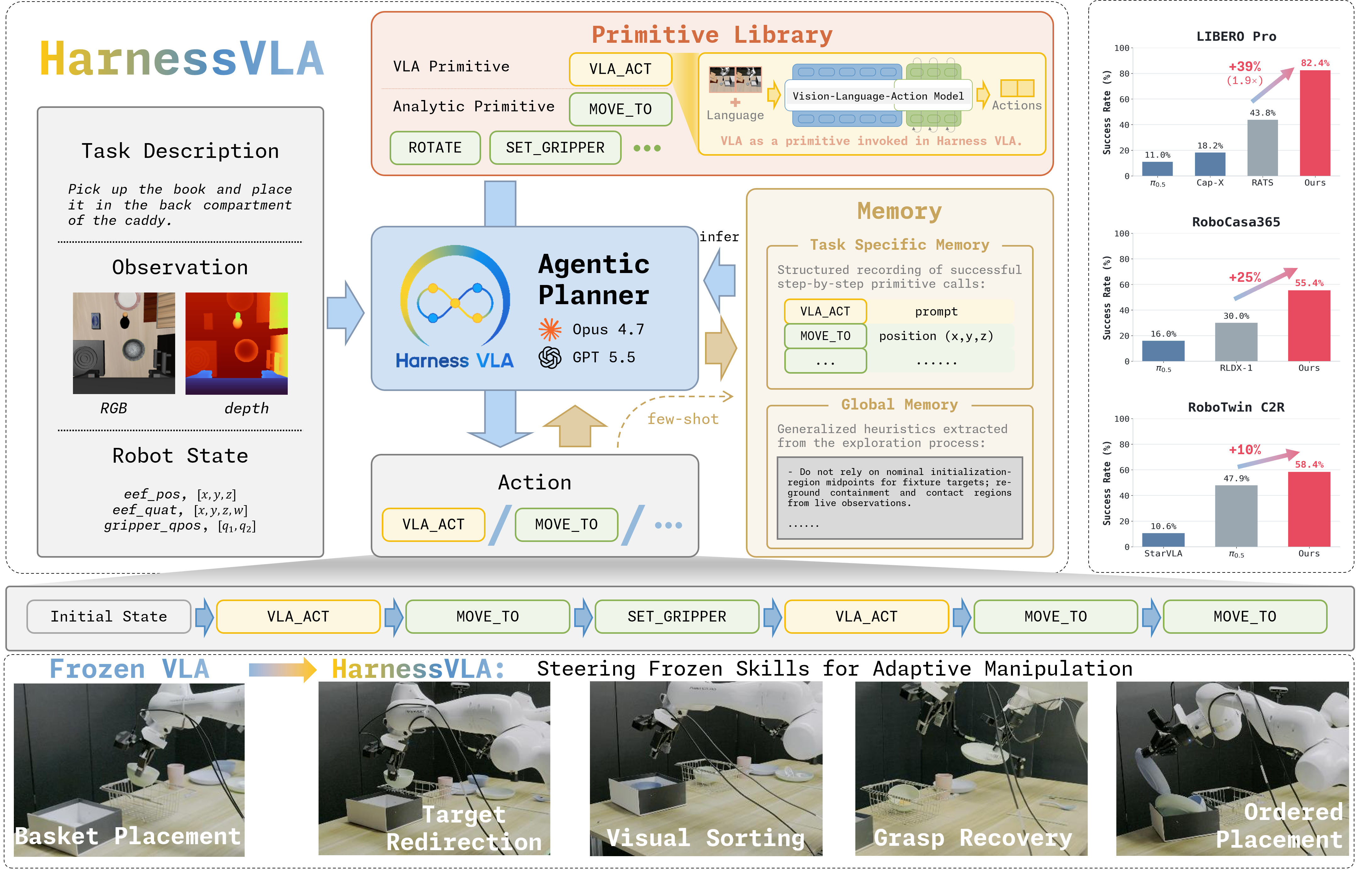}
\caption{\textbf{Harness VLA system overview.} Given a task description, RGB-D observations, and robot state, the agentic planner selects structured calls from a fixed primitive library rather than emitting low-level actions directly. The library exposes the frozen VLA as \textsc{vla\_act} for contact-rich behaviors and uses analytic primitives such as \textsc{move\_to}, \textsc{rotate}, and \textsc{set\_gripper} for perception-conditioned staging, transport, posture adjustment, and release. Task Specific Memory stores successful command traces from reference-seed exploration for few-shot re-grounding, while Global Memory stores reusable success rules and failure models. The right panels summarize benchmark results, and the command strip illustrates the interleaving of VLA calls and analytic control. The bottom row contrasts frozen-VLA basket placement with target redirection, visual sorting, grasp recovery, and ordered placement by Harness VLA on dual-Franka robots.}
\label{fig:scheme}
\end{figure}
\clearpage

\begin{abstract}
Language-conditioned manipulation requires both precise contact-rich control and robust reasoning over language, scenes, and long horizons. End-to-end Vision-Language-Action (VLA) models provide strong local visuomotor skills, but they are trained on in-distribution task trajectories and often degrade under deployment perturbations such as semantic retargeting, goal re-binding, spatial-layout shifts, and unstable local contacts. LLM coding agents provide complementary semantic and compositional reasoning, but purely analytic primitives struggle with irregular grasping, constrained placement, and articulated-object interaction. We present Harness VLA, a memory-augmented agentic framework that exposes a frozen VLA as a retryable contact-rich primitive and composes it with a small fixed library of analytic primitives for grounding, staging, transport, navigation, and release. Rather than expanding the skill library, the harness learns the operating range of these fixed primitives from task-specific execution traces, global success rules, and failure models. By lifting semantic re-grounding, non-contact execution, and VLA re-staging to the planner while reserving the frozen VLA for local contact-rich phases, Harness VLA extends pretrained VLAs beyond their original trajectory distribution without fine-tuning. Across perturbed tabletop, household kitchen, and clean-to-randomized bimanual manipulation, Harness VLA improves over the strongest relevant baselines by $38.6$ and $27.1$ percentage points on LIBERO-Pro and RoboCasa365, respectively, and reaches $58.4\%$ on RoboTwin C2R. Real-world demonstrations on dual-Franka robots further show target redirection, grasp recovery, and new task compositions with the same frozen VLA. Code is available at \url{https://github.com/RLinf/RPent}.
\end{abstract}

\section{Introduction}
\label{sec:intro}

\begin{figure}[htbp]
\centering
\includegraphics[width=0.9\linewidth]{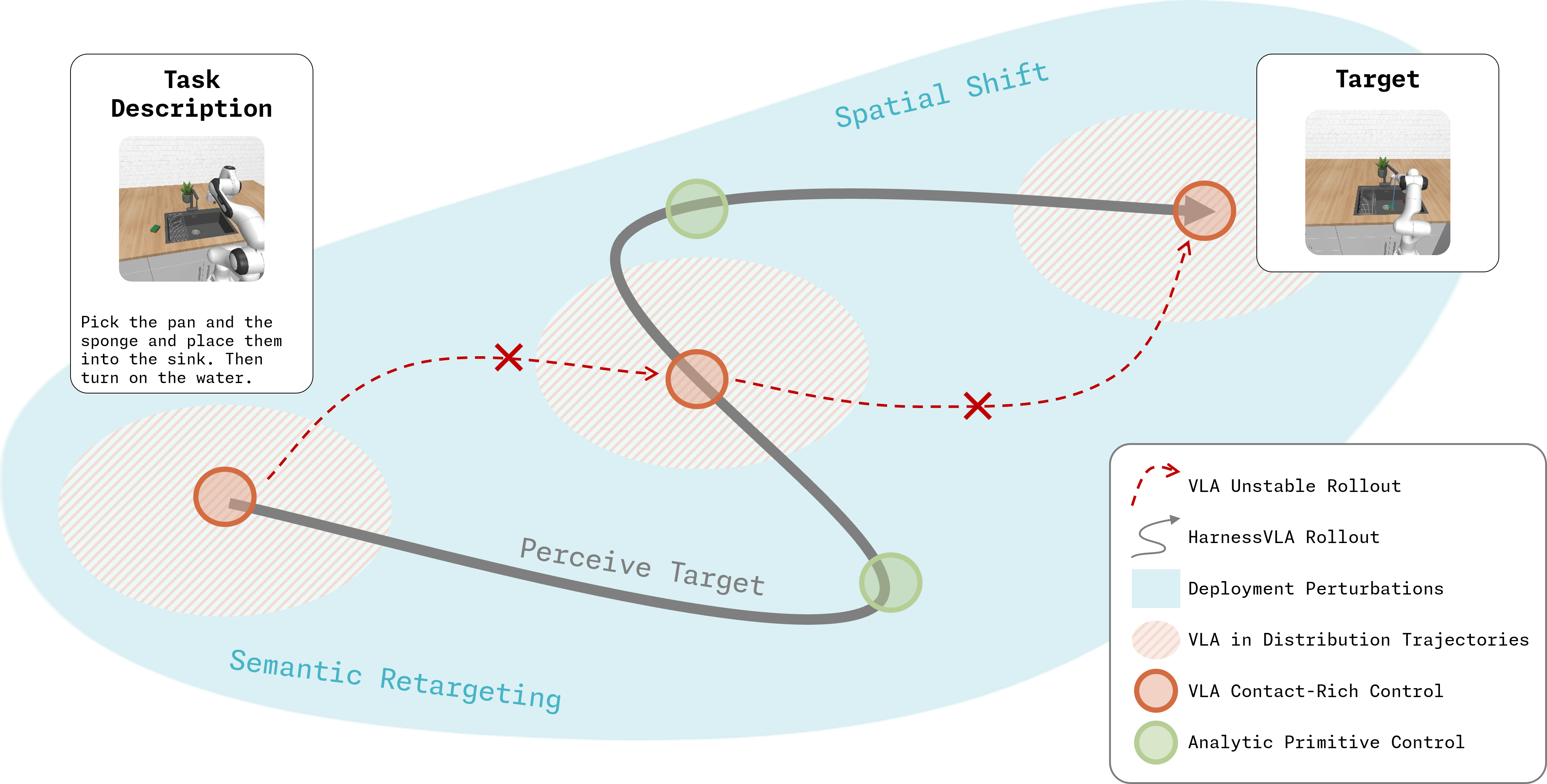}
\caption{\textbf{Primitive composition extends a frozen VLA beyond its trajectory distribution.} Deployment perturbations expand the possible task configurations beyond the in-distribution trajectories covered by the frozen VLA. A direct VLA rollout may attempt to bridge the perturbed space and fail before reaching the target. Harness VLA instead decomposes the task into local contact-rich VLA invocations and analytic primitive control: analytic primitives perceive the current target, re-ground task bindings, and move the robot between VLA-compatible local regions, while \textsc{vla\_act} is invoked only for contact-rich phases inside those regions.}
\label{fig:teaser}
\end{figure}

    A long-standing goal of robotic manipulation is a system that reliably executes free-form natural-language instructions across changing objects, layouts, and embodiments. Two dominant paradigms approach this goal from opposite directions. End-to-end Vision-Language-Action (VLA) models learn contact-rich visuomotor control directly from robot trajectories, while LLM coding agents use language-model reasoning to compose explicit perception-and-control APIs. Each paradigm is powerful, but each assigns the wrong component too much responsibility: monolithic VLAs must absorb language grounding, long-horizon composition, and low-level control inside a single policy, whereas coding agents must realize physically delicate interactions through hand-designed or agent-generated APIs. Figure~\ref{fig:teaser} visualizes our response: use analytic primitives to traverse deployment perturbations and invoke the VLA only inside local contact-rich regions where its training distribution is informative.

    \textbf{End-to-end VLA models} have advanced rapidly, from generalist robot policies~\citep{brohan2023rt2, kim2024openvla} to flow-matching and action-reasoning architectures~\citep{black2025pi0, black2025pi05, hung2025nora, molmoact2025}. Their strength is local, image-conditioned contact: grasping irregular objects, placing with tight tolerances, or actuating fixtures that are brittle for analytic controllers. Their weakness is deployment outside the trajectory distribution on which they were trained. A model trained on in-distribution task trajectories may know how to grasp a milk carton or turn a faucet, yet fail when semantic targets are redirected, goal predicates are re-bound, object layouts shift, or short skills must be composed into longer routines. Under such deployment perturbations, the policy may repeat a familiar training-time behavior even when the instruction or scene binding has changed~\citep{kim2024openvla, black2025pi05, pi0-experiment-wild}; a single unstable contact failure can also derail the entire monolithic rollout.

    \textbf{LLM coding agents and harnesses} provide complementary semantic and compositional reasoning. Systems such as Code as Policies and ProgPrompt synthesize executable programs over curated perception and control APIs~\citep{liang2023code, singh2023progprompt}, and recent multimodal or agentic variants extend this idea with richer perception, tool use, feedback, and persistent execution state~\citep{gupta2023visprog, shi2025maestro, fu2026cap, zhang2026playful}. More broadly, coding-agent harnesses wrap model outputs in a structured runtime with tool interfaces, memory, validators, execution loops, and feedback channels, allowing an agent to revise decisions, write successful traces or failure diagnoses back into memory, and orchestrate heterogeneous tools under a common control surface~\citep{wang2024codeact, yang2024sweagent, wang2025openhands, ning2026code}. Yet in robot manipulation, scaling such systems often still means expanding the primitive or skill library, while purely analytic primitives--deterministic kinematic or model-based controllers such as IK transport, wrist rotation, base motion, gripper opening, and release--remain poorly suited to irregular grasping, constrained placement, and articulated-object manipulation.

    \textbf{Harness VLA} instantiates this coding-agent harness view for robot manipulation: keep the primitive library fixed and small, and let the agent learn how to orchestrate it. The planner composes analytic primitives for non-contact structure such as target grounding, free-space transport, posture adjustment, mobile staging, re-staging after failed attempts, and release. For contact-rich phases, it invokes a frozen VLA through a single learned primitive, \textsc{vla\_act}. This converts the VLA from a monolithic trajectory policy into a reusable contact specialist, extending it to the tasks outside its original trajectory distribution without fine-tuning or deployment-time primitive expansion.
    The key is not only exposing \textsc{vla\_act}, but learning when and how to use it. Harness VLA treats VLA execution as a retryable local attempt: the planner can stage the robot into a favorable local observation, invoke the VLA, inspect the contact outcome, and re-stage if needed. Two memory modules support this process inside the agentic harness~\citep{wang2024codeact, yang2024sweagent, wang2023voyager}: task-specific traces store successful primitive compositions for few-shot re-grounding, while global memory stores reusable success rules and failure models. Rather than adding more skills, the harness teaches the planner the operating range of each fixed primitive: which subproblems should be handled analytically, when \textsc{vla\_act} is appropriate, and how failed contact attempts should be re-staged. Our core contributions are:
\begin{itemize}
    \item A memory-augmented agentic framework for using a frozen VLA as a primitive. Harness VLA composes \textsc{vla\_act} with fixed analytic primitives, extending a pretrained VLA from local contact-rich control to long-horizon, perturbed manipulation without fine-tuning the VLA or expanding the primitive vocabulary at deployment time.
    \item An empirical analysis showing why a small fixed primitive library is sufficient when the planner learns how to use it. Repeated planner-staged invocations can reframe brittle VLA attempts, while analytic primitives solve much of the non-contact structure around each contact-rich phase.
    \item Strong benchmark results across standard and perturbed tabletop manipulation, household kitchen manipulation, and clean-to-randomized transfer. Harness VLA preserves competitive standard LIBERO performance, improves over the strongest relevant baselines by $38.6$ and $27.1$ percentage points on LIBERO-Pro and RoboCasa365, respectively, and reaches $58.4\%$ on RoboTwin C2R. We further conduct real-world experiments on dual-Franka robots to illustrate the mechanisms of Harness VLA.
\end{itemize}

\section{The Harness VLA Framework}
\label{sec:method}

	Our agentic framework for language-conditioned manipulation follows the system view in Figure~\ref{fig:scheme}. A task description, RGB-D observations, and robot state are passed to an agentic planner, which reasons over a fixed primitive library and retrieves context from Task Specific Memory and Global Memory. The agentic harness (Section~\ref{sec:method:harness}) couples this planner to the simulator through a JSON-serialized primitive interface, drives the turn-based execution loop, and writes successful exploration traces into Task Specific Memory while committing generalized heuristics to Global Memory. The primitive library (Section~\ref{sec:method:primitives}) defines the only operations the planner is allowed to invoke: a small set of analytic primitives together with a structurally special VLA primitive that encapsulates a pretrained visuomotor policy for contact-rich interactions. Section~\ref{sec:method:formulation} first formalises the task and the iterative execution loop on which these components are built.

	\subsection{Problem Formulation and Agentic Execution Loop}
	\label{sec:method:formulation}

	\textbf{Task setup.} We consider language-conditioned robotic manipulation within an environment $\mathcal{E}$ driven by a rigid-body physics engine (e.g., MuJoCo via Robosuite). At each timestep $t$, the environment exposes a multimodal observation tuple $o_t = (I_t^{\text{rgb}}, I_t^{\text{d}}, q_t)$ comprising an agent-view RGB image $I_t^{\text{rgb}}$, a co-aligned metric depth map $I_t^{\text{d}}$, and a robot proprioceptive state $q_t$ (concatenating the end-effector pose and gripper state). A task is defined by a natural-language description $\ell$ alongside a binary completion predicate $\mathcal{G}$, exposed solely as a sparse success signal at episode termination.

	\textbf{Agentic execution loop.} As illustrated by the rollout strip in Figure~\ref{fig:scheme}, a task rollout is formulated as an autoregressive, turn-based interaction between a high-level agentic planner $\Pi$ and the underlying physics engine. Instead of treating the visuomotor policy as a separate hierarchical tier, we unify all low-level control mechanisms---including the frozen pretrained VLA $f_\theta$ and all deterministic operational-space controllers---into a single, predefined primitive library $\mathcal{P}$.

	At each execution turn $t$, the planner $\Pi$ processes the current multimodal observation $o_t$, the task description $\ell$, and the retrieved context from both the Task Specific Memory and Global Memory. Operating as the sole cognitive orchestrator, $\Pi$ emits a structured JSON invocation for a selected primitive $c_t \in \mathcal{P}$. The physics engine directly receives this invocation and executes the corresponding physical motions in the simulator until the primitive's internal post-condition is met. Upon primitive termination, the engine yields the subsequent observation $o_{t+1}$ and updated robot state $q_{t+1}$. This environment-planner loop iterates continuously until the goal predicate $\mathcal{G}$ is satisfied or a maximum step budget is exhausted.

	\subsection{The Harness VLA Architecture}
	\label{sec:method:harness}

Motivated by recent coding agents, which make model decisions executable through harnessed execution-feedback loops~\citep{wang2024codeact, yang2024sweagent, wang2025openhands}, Harness VLA packages robot manipulation in the same REPL-style form. The harness is the runtime contract between the planner and the environment: it exposes primitive schemas, serializes decisions as JSON commands, executes primitives, refreshes RGB-D and proprioceptive observations, logs traces, retrieves Task Specific Memory and Global Memory, enforces reset and budget policies, and checks progress through the benchmark predicate~\citep{ning2026code}. Appendix~\ref{app:harness} details the REPL protocol, and Appendix~\ref{app:agent_prompt} specifies the agent prompts.

Because this harness delegates all fine-grained execution to the primitive library, the agentic planner $\Pi$ is freed to focus entirely on compositional reasoning. To do so, it relies heavily on the multimodal observation channel: the RGB image supports qualitative scene reasoning (e.g., clutter, semantic identity), while the co-aligned depth map and proprioception supply metric spatial data for precise localization.

We structure the agent's lifecycle within this harness into two distinct phases: an exploratory bootstrapping phase and a rigorous deployment evaluation phase.

\textbf{Exploratory Bootstrapping Phase.} Operating on a single reference instantiation of a task, the agent autonomously interacts with the environment to discover a working solution. During this phase, the planner $\Pi$ is uniquely granted access to a \textsc{reset} primitive and operates under a generous wall-clock budget. Because the primitive vocabulary is fixed, the exploration focuses entirely on iterative composition: discovering the optimized orchestration of the learned VLA primitive and the deterministic analytic primitives. The planner $\Pi$ repeatedly trials different staging orders, pre-contact poses, invocation timings for \textsc{vla\_act}, and early-return termination thresholds. It observes the physical effects of each primitive call and corrects course upon failure.

Upon successful task completion, the agent systematically abstracts its experience into the two memory modules shown in Figure~\ref{fig:scheme}. First, the verified sequence of primitive invocations is serialized into a JSONL format. This file explicitly records the successful step-by-step primitive calls, parameterizing them by replacing concrete spatial coordinates with symbolic perception queries to make the sequence reusable across different spatial layouts. This parameterized JSONL trace is stored in the \textbf{Task Specific Memory} to serve as a structural prior for subsequent generalization tests. Second, the agent extracts generalized heuristics from the exploration process and commits them to a persistent \textbf{Global Memory}. This shared repository explicitly aggregates \emph{success rules}, such as optimal prompting strategies that utilize the full task instruction. It concurrently documents critical \emph{failure models}, including the identification of empty-grasp executions and false success detections. This aggregation ensures the planner avoids repeating historical pitfalls across different tasks.

\textbf{Deployment Evaluation Phase.} During formal evaluation on unseen environment variations, including position swaps, instruction redirections, and testing across multiple initial state seeds, the harness imposes a strict execution regime. The \textsc{reset} primitive is completely disabled, and the operational step budget is significantly shortened. To solve the perturbed tasks, the planner $\Pi$ retrieves the pre-computed JSONL trace from the \textbf{Task Specific Memory} and grounds it dynamically using the live RGB-D observation. By referencing the success rules and failure models accumulated in the \textbf{Global Memory}, the agent executes the trajectory deterministically. The performance achieved under this strict phase directly constitutes our reported benchmark results, validating the overall effectiveness of the Harness VLA framework.

	\subsection{Unified Primitive Interface}
	\label{sec:method:primitives}

\begin{table}[htbp]
\centering
\caption{Primitive vocabulary. The same primitive names are used throughout the paper; RoboCasa365 additionally uses mobile-base primitives for kitchen-scale staging.}
\label{tab:primitive_vocabulary}
\small
\setlength{\tabcolsep}{4pt}
\begin{tabular}{p{0.20\linewidth}p{0.18\linewidth}p{0.50\linewidth}}
\toprule
Primitive & Type & Role \\
\midrule
\textsc{move\_to} & Composite & Move an end-effector to a world-frame Cartesian target using the environment's embedded solver. \\
\textsc{move\_pose} & Composite & Move the end-effector while co-varying pose variables such as pitch for reach-limited configurations. \\
\textsc{rotate\_wrist} & Atomic & Apply a wrist-yaw set-point while holding the current spatial position. \\
\textsc{rotate\_pitch} & Atomic & Apply a wrist-pitch set-point while holding the current spatial position. \\
\textsc{set\_gripper} & Atomic & Drive the gripper to an open or closed set-point for a fixed number of steps. \\
\textsc{release} & Atomic & Open the gripper under a release post-condition. \\
\textsc{vla\_act} & VLA & Execute a frozen VLA in short bursts for local contact-rich interaction. \\
\midrule
\textsc{navigate\_to} & Composite\newline (RoboCasa365) & Drive the mobile base to a world-frame location for kitchen-scale staging. \\
\textsc{move\_base} & Atomic\newline (RoboCasa365) & Apply an open-loop local base-velocity set-point for fine repositioning. \\
\bottomrule
\end{tabular}
\end{table}

The primitive library $\mathcal{P}$ is the only action interface exposed to the planner. Each primitive is invoked by a single JSON object, executes inside the environment until an internal post-condition is reached, and then returns control together with a refreshed observation. Example JSON invocations are shown at the end of this subsection. The planner therefore never emits low-level torques, joint targets, or action chunks directly; it selects a primitive and binds its arguments from language, RGB-D observations, proprioception, and memory.

We organize $\mathcal{P}$ into two manipulation families. \emph{Analytic primitives} are deterministic, model-based controllers specified from robot kinematics and require no training data. They split into \emph{composite} primitives, which take a world-frame spatial goal and run an embedded solver to coordinate multiple degrees of freedom, and \emph{atomic} primitives, which drive one intrinsic channel such as wrist orientation, gripper state, or base velocity to a parametric set-point. The \emph{VLA primitive}, \textsc{vla\_act}, is a learned policy call that maps a prompt and live cameras to action chunks for local contact-rich behavior. The exploratory \textsc{reset} utility is used only during bootstrapping and is not counted as a manipulation primitive.

Table~\ref{tab:primitive_vocabulary} gives the primitive vocabulary used throughout the paper. The shared manipulation interface contains six analytic primitives and one VLA primitive; RoboCasa365 additionally uses two mobile-base primitives, \textsc{navigate\_to} and \textsc{move\_base}, for kitchen-scale staging. Details of RoboTwin bimanual execution are provided in Appendix~\ref{app:primitives}. Crucially, the primitive vocabulary is fixed before evaluation; the planner cannot invent new primitives at deployment time.

The compact JSON contract below illustrates the shared interface; Appendix~\ref{app:primitives} gives the benchmark-specific availability and implementation notes using the same primitive names.

\begin{lstlisting}[frame=single,framerule=0.5pt]
{"action": "move_to",     "xyz": [<x>,<y>,<z>], ...}
{"action": "move_pose",   "xyz": [<x>,<y>,<z>], "pose": <orientation>, ...}
{"action": "rotate_wrist","target_yaw": <float>, ...}
{"action": "rotate_pitch","target_pitch": <float>, ...}
{"action": "set_gripper", "gripper": <open|close>, ...}
{"action": "release",     ...}
{"action": "navigate_to", "xy": [<x>,<y>], ...}
{"action": "move_base",   "forward": <float>, "lateral": <float>, "turn": <float>, ...}
{"action": "vla_act",     "prompt": <str>, "max_chunks": <int>,
                          "stop": <predicate>}
\end{lstlisting}

\paragraph{VLA-backed contact primitive.}
\textsc{vla\_act} is the learned primitive for contact-rich interaction. Across benchmarks, \textsc{vla\_act} covers grasping, constrained placement, fixture actuation, button pressing, drawer or door manipulation, insertion, and embodiment-specific contact behaviors. The planner supplies a task-conditioned prompt and an early-return predicate $\tau$. The frozen VLA $f_\theta$ then emits action chunks until $\tau$ is satisfied or the chunk budget is exhausted. This keeps the VLA as a local contact specialist while semantic grounding, spatial re-binding, navigation, re-staging, and long-horizon composition remain under planner control.

\section{Experiments}
\label{sec:result}

We organize the empirical study around two deployment regimes and three mechanism analyses. In the \textbf{few-shot} regime, Harness VLA follows the memory-backed workflow in Section~\ref{sec:method:harness}: the agent performs task-level bootstrapping on one reference seed, stores the successful primitive trace in Task Specific Memory, and re-grounds that trace under new seeds or perturbations. In the \textbf{zero-shot} regime, the agent must solve without retrieving Task Specific Memory or Global Memory for the target setting, testing how far online planner reasoning and the frozen primitive interface transfer without task-specific harness memory. Section~\ref{sec:exp:setup} details the configuration, Section~\ref{sec:exp:overall} presents few-shot and zero-shot benchmark performance, and Section~\ref{sec:exp:analysis} analyzes the mechanisms behind the gains. We additionally evaluate Harness VLA with GPT-6 Astra as the planner backbone; Appendix~\ref{app:astra_results} reports the results.

\subsection{Experimental Setup}
\label{sec:exp:setup}

We evaluate Harness VLA on four benchmark families. The tabletop suites are LIBERO~\citep{liu2023libero} and LIBERO-Pro~\citep{Lagrange1788}; the household and bimanual suites are RoboCasa365~\citep{robocasa365} and RoboTwin C2R~\citep{mu2025robotwin}, respectively. We defer benchmark-specific task splits and seed protocols to Appendix~\ref{app:benchmarks}, and focus here on the main empirical outcomes. Across all evaluations, the planner operates over the same frozen primitive vocabulary $\mathcal{P}$ (Section~\ref{sec:method:primitives}) and is not allowed to introduce new primitives at deployment time. The VLA primitive is instantiated with benchmark-specific frozen policies while preserving a unified \textsc{vla\_act} interface: the RLinf-released \texttt{pi05\_libero130\_fullshot} $\pi_{0.5}$-SFT checkpoint, denoted $\piRLinf$~\citep{yu2025rlinf}, for LIBERO and LIBERO-Pro, the frozen RLDX-1 RoboCasa checkpoint~\citep{kim2026rldx} for RoboCasa365, and our post-trained LingBot-VLA checkpoint~\citep{wu2026pragmatic} for RoboTwin C2R. In the tables below, \textbf{Harness VLA (Codex)} and \textbf{Harness VLA (CC)} denote the same harness instantiated with Codex and Claude Code planners, respectively; \textbf{CC} abbreviates Claude Code. The $\piRLinf$, RLDX-1, and LingBot-VLA rows serve as direct frozen-VLA baselines for their corresponding benchmarks. Appendix~\ref{app:vla_instantiations} details the VLA instantiations.

\subsection{Overall Benchmark Performance}
\label{sec:exp:overall}

\paragraph{Few-shot evaluation with Task Specific Memory.}
We first evaluate Harness VLA after task-level bootstrapping has populated Task Specific Memory. This setting tests whether the harness can reuse the primitive organization discovered on a reference seed while grounding all spatial arguments from the current observations. We evaluate whether this memory-backed execution preserves strong in-distribution manipulation performance on standard LIBERO, remains robust under LIBERO-Pro instruction-redirection (\textbf{T}) and position-swap (\textbf{S}) perturbations, and extends the same primitive interface to household kitchen manipulation in RoboCasa365.

\textbf{Standard LIBERO.} Table~\ref{tab:libero_main} reports results on the four standard LIBERO suites. Harness VLA (CC) achieves an aggregate success rate of $\mathbf{96.0\%}$ ($384/400$), including $100.0\%$ on \textsc{Object} and $93.0\%$ on \textsc{LIBERO-10}. Compared with the frozen $\piRLinf$ checkpoint used inside \textsc{vla\_act}, which obtains $95.3\%$ overall, Harness VLA preserves competitive standard-suite performance while exposing the same policy through a controllable primitive interface for the perturbed evaluations below.

\begin{table}[htbp]
\centering
\caption{Success rate (\%) on standard LIBERO. $\piRLinf$~\citep{yu2025rlinf} and Harness VLA (CC) are evaluated by us on $100$ trials per suite ($10$ tasks $\times$ $10$ seeds). Harness VLA uses the same RLinf-released \texttt{pi05\_libero130\_fullshot} $\pi_{0.5}$-SFT checkpoint inside \textsc{vla\_act}. Bold marks the best method in each suite or overall column.}
\label{tab:libero_main}
\small
\setlength{\tabcolsep}{5pt}
\begin{tabular}{lccccc}
\toprule
Method                                   & Spatial      & Object        & Goal         & LIBERO-10    & \textbf{Overall} \\
\midrule
OpenVLA~\citep{kim2024openvla}           & $84.7$       & $88.4$        & $79.2$       & $53.7$       & $76.5$ \\
NORA~\citep{hung2025nora}                & $85.6$       & $89.4$        & $80.0$       & $63.0$       & $79.5$ \\
$\pi_0$~\citep{black2025pi0}             & $96.8$       & $98.8$        & $95.8$       & $85.2$       & $94.2$ \\
$\piRLinf$                               & $\mathbf{99.0}$ & $96.0$     & $97.0$       & $89.0$       & $95.3$ \\
AtomVLA~\citep{sun2026atomvla}           & $96.4$       & $99.6$        & $\mathbf{97.6}$ & $94.4$    & $\mathbf{97.0}$ \\
\textbf{Harness VLA (CC)}           & $97.0$       & $\mathbf{100.0}$ & $94.0$    & $93.0$       & $96.0$ \\
\bottomrule
\end{tabular}
\end{table}

\begin{table}[htbp]
\centering
\caption{Aggregate LIBERO-Pro success rates (\%) across \textsc{Spatial}, \textsc{Object}, \textsc{Goal}, and \textsc{LIBERO-10} under instruction-redirection (\textbf{T}) and position-swap (\textbf{S}) perturbations. Each cell aggregates $100$ trials ($10$ tasks $\times$ $10$ seeds); ``/'' indicates an unavailable or unreported cell. Cap-X and RATS report only the six non-\textsc{LIBERO-10} cells, so their overall values average over reported cells only. $\piRLinf$ and Harness VLA are evaluated by us using the RLinf-released \texttt{pi05\_libero130\_fullshot} $\pi_{0.5}$-SFT checkpoint~\citep{yu2025rlinf}; Harness VLA exposes this frozen checkpoint through the \textsc{vla\_act} primitive. Bold marks the best reported method in each evaluation cell or overall column.}
\label{tab:libero_pro_perturbation}
\small
\setlength{\tabcolsep}{4pt}
\begin{tabular}{l cccccccc c}
\toprule
Method & Spat-T & Spat-S & Obj-T & Obj-S & Goal-T & Goal-S & L10-T & L10-S & \textbf{Overall} \\
\midrule
OpenVLA~\citep{kim2024openvla}           & 0.0 & 0.0 & 0.0 & 0.0 & 0.0 & 0.0 & 0.0 & 0.0 & 0.0 \\
$\pi_0$~\citep{black2025pi0}             & 0.0 & 0.0 & 0.0 & 2.0 & 0.0 & 0.0 & 0.0 & 0.0 & 0.3 \\
$\pi_{0.5}$~\citep{black2025pi05}        & 1.0 & 20.0 & 1.0 & 17.0 & 2.0 & 38.0 & 1.0 & 8.0 & 11.0 \\
MolmoAct~\citep{molmoact2025}            & 0.0 & 0.0 & 0.0 & 6.0 & 0.0 & 0.0 & 6.0 & 0.0 & 1.5 \\
NORA~\citep{hung2025nora}                & 0.0 & 0.0 & 0.0 & 0.0 & 0.0 & 0.0 & 0.0 & 0.0 & 0.0 \\
X-VLA~\citep{zheng2025xvla}              & 0.0 & 0.0 & 8.0 & 2.0 & 9.0 & 1.0 & 10.0 & 0.0 & 3.8 \\
AtomVLA~\citep{sun2026atomvla}           & 1.0 & 16.0 & 0.0 & 10.0 & 11.0 & 2.0 & 9.0 & 1.0 & 6.3 \\
\midrule
Cap-X~\citep{fu2026cap}                  & 14.0 & 12.0 & 18.0 & 22.0 & 17.0 & 26.0 &  / &  / & 18.2 \\
RATS~\citep{zhang2026playful}            & 31.0 & 29.0 & 63.0 & 61.0 & 36.0 & 43.0 &  / &  / & 43.8 \\
\midrule
$\piRLinf$                               & 42.0 & 59.0 & 71.0 & 78.0 & 45.0 & 42.0 & 49.0 & 14.0 & 50.0 \\
\textbf{Harness VLA (Codex)}             & 81.0 & 69.0 & \textbf{94.0} & \textbf{91.0} & 75.0 & 66.0 & 52.0 & 49.0 & 72.1 \\
\textbf{Harness VLA (CC)}                & \textbf{94.0} & \textbf{80.0} & 88.0 & 90.0 & \textbf{87.0} & \textbf{87.0} & \textbf{71.0} & \textbf{62.0} & \textbf{82.4} \\
\bottomrule
\end{tabular}
\end{table}

\textbf{LIBERO-Pro.} Table~\ref{tab:libero_pro_perturbation} reports aggregate results on LIBERO-Pro, spanning \textsc{Spatial}, \textsc{Object}, \textsc{Goal}, and \textsc{LIBERO-10} under instruction-redirection (\textbf{T}) and position-swap (\textbf{S}) perturbations. Existing end-to-end VLA models degrade sharply under these distribution shifts, and RATS is the strongest reported prior baseline with $43.8\%$ overall on its reported cells. Harness VLA reaches $72.1\%$ with Codex and $82.4\%$ with CC, improving over RATS by $38.6$ percentage points in the headline comparison. The direct $\piRLinf$ baseline reaches $50.0\%$ overall under our protocol, showing that the gain does not simply come from the frozen VLA backbone. The gains across both instruction-redirection and position-swap settings show that the few-shot harness has learned a reusable division of labor over the fixed primitive library: the planner can re-bind targets, use analytic primitives to re-stage the scene and handle non-contact execution, and invoke the VLA only for local contact-rich manipulation.

\textbf{RoboCasa365.} RoboCasa365 extends the evaluation from tabletop manipulation to household kitchen tasks with mobile staging, articulated fixtures, and longer composite routines. Table~\ref{tab:robocasa_main} compares Harness VLA against results reported in prior RoboCasa365 papers and the RLDX-1 baseline used for the headline comparison. RLDX-1 reaches an overall success rate of $30.0\%$, while Harness VLA reaches $57.1\%$ with Codex and $48.6\%$ with CC; the Codex instantiation therefore improves over RLDX-1 by $27.1$ percentage points. These gains are consistent with the intended decomposition: the planner handles navigation, staging, and re-staging after local failures, while the frozen VLA remains the local contact-rich primitive.

\begin{table}[htbp]
\centering
\caption{RoboCasa365 success rates (\%). Baseline rows above the separator are results reported in the corresponding prior papers on \textsc{Atomic-Seen}, \textsc{Composite-Seen}, and \textsc{Composite-Unseen}, with RLDX-1 evaluated as the direct frozen-VLA baseline under our protocol. Harness VLA uses one reference seed only for bootstrapping; reported evaluation uses five held-out seeds $s_1$--$s_5$ per task in all three splits, totaling $250$ rollouts. Bold marks the best method in each split.}
\label{tab:robocasa_main}
\small
\setlength{\tabcolsep}{7pt}
\begin{tabular}{l cccc}
\toprule
Method & Atomic-Seen & Composite-Seen & Composite-Unseen & \textbf{Overall} \\
\midrule
RLDX-1~\citep{kim2026rldx} & $60.0$ & $21.3$ & $5.0$ & $30.0$ \\
WorldDreamer~\citep{wang2024worlddreamer} & $66.3$ & $26.7$ & $9.0$ & $35.3$ \\
$\pi_{0.5}$~\citep{black2025pi05} & $39.6$ & $7.1$ & $1.2$ & $16.9$ \\
$\pi_0$~\citep{black2025pi0} & $34.6$ & $6.1$ & $1.1$ & $14.8$ \\
\midrule
\textbf{Harness VLA (Codex)} & $\mathbf{92.2}$ & $\mathbf{61.3}$ & $13.8$ & $\mathbf{57.2}$ \\
\textbf{Harness VLA (CC)} & $80.0$ & $47.5$ & $\mathbf{15.0}$ & $48.8$ \\
\bottomrule
\end{tabular}
\end{table}

\paragraph{Zero-shot evaluation without bootstrapped harness memory.}
\textbf{LIBERO-Pro Goal.} To separate online planner reasoning from bootstrapped harness memory, we evaluate \textsc{LIBERO-Pro Goal} in a strict zero-shot setting where the agent does not retrieve the target-setting Task Specific Memory or the corresponding Global Memory. Table~\ref{tab:zeroshot_vs_capx_transposed} shows that zero-shot Harness VLA (CC) outperforms Cap-X across both perturbation regimes, reaching $31.0\%$ on position-swap (Pos-S) and $79.0\%$ on instruction-redirection (Task-T), compared with Cap-X's $25.6\%$ and $16.8\%$. The comparison with the few-shot \textsc{Goal} cells in Table~\ref{tab:libero_pro_perturbation} clarifies what bootstrapped harness memory contributes. Without this memory, the planner retains much of its semantic re-binding ability under instruction redirection ($79.0\%$ zero-shot versus $87.0\%$ few-shot on Goal-T), but drops substantially under position swaps ($31.0\%$ zero-shot versus $87.0\%$ few-shot on Goal-S). This gap indicates that spatially perturbed manipulation benefits strongly from the task-specific primitive organization discovered during exploration: analytic primitives supply localization, staging, transport, and release around the contact-rich phase, while \textsc{vla\_act} is invoked at the learned interaction points and can be re-staged after failures.

\begin{table}[htbp]
\centering
\caption{Per-task success rate (\%) on \textsc{LIBERO-Pro Goal}: zero-shot Harness VLA (CC) (no Task Specific Memory retrieval, $10$ seeds/task) vs Cap-X~\citep{fu2026cap}. \textbf{Pos} = swap (S), \textbf{Task} = instruction-redirection (T). Bold marks the winner within each setting and task or average column.}
\label{tab:zeroshot_vs_capx_transposed}
\small
\resizebox{\textwidth}{!}{%
\begin{tabular}{ll cccccccccc c}
\toprule
\textbf{Setting} & \textbf{Method} & \textbf{Task 0} & \textbf{Task 1} & \textbf{Task 2} & \textbf{Task 3} & \textbf{Task 4} & \textbf{Task 5} & \textbf{Task 6} & \textbf{Task 7} & \textbf{Task 8} & \textbf{Task 9} & \textbf{Average} \\
\midrule
\textbf{Pos (S)}
& Cap-X   & $0.0$ & $4.0$ & $0.0$ & $\mathbf{36.0}$ & $22.0$ & $\mathbf{60.0}$ & $4.0$ & $2.0$ & $62.0$ & $\mathbf{66.0}$ & $25.6$ \\
& Harness VLA (CC) & $0.0$ & $\mathbf{10.0}$ & $0.0$ & $20.0$ & $\mathbf{90.0}$ & $0.0$ & $\mathbf{10.0}$ & $\mathbf{80.0}$ & $\mathbf{100.0}$ & $0.0$ & $\mathbf{31.0}$ \\
\midrule
\textbf{Task (T)} 
& Cap-X   & $0.0$ & $0.0$ & $10.0$ & $38.0$ & $12.0$ & $4.0$ & $34.0$ & $12.0$ & $40.0$ & $18.0$ & $16.8$ \\
& Harness VLA (CC) & $\mathbf{10.0}$ & $\mathbf{100.0}$ & $\mathbf{90.0}$ & $\mathbf{100.0}$ & $\mathbf{20.0}$ & $\mathbf{80.0}$ & $\mathbf{90.0}$ & $\mathbf{100.0}$ & $\mathbf{100.0}$ & $\mathbf{100.0}$ & $\mathbf{79.0}$ \\
\bottomrule
\end{tabular}%
}
\end{table}

\textbf{RoboTwin C2R.} RoboTwin C2R evaluates zero-shot clean-to-randomized transfer: the agent obtains a Task Specific Memory trace from the clean setting and transfers it directly to randomized task instances, with no randomized-setting bootstrapping, additional task-level exploration, or VLA fine-tuning. The \textsc{vla\_act} backend here is LingBot-VLA, our RoboTwin-specialized VLA checkpoint; after post-training, it is frozen for both the direct VLA baseline and Harness VLA evaluation. Table~\ref{tab:robotwin_c2r} compares C2R success rates against representative VLA baselines. Direct LingBot-VLA reaches $50.4\%$, while Harness VLA raises the same frozen backend to $58.0\%$ with Codex and $58.4\%$ with CC. The table also reports external VLA baselines for context.

\begin{table}[htbp]
\centering
\caption{RoboTwin C2R success rates (\%). LingBot-VLA is our post-trained RoboTwin VLA checkpoint evaluated directly without agent-level decomposition, and is also the frozen \textsc{vla\_act} backend used by Harness VLA. Other VLA rows are representative external baselines. Harness VLA is evaluated on $50$ tasks with $5$ randomized seeds per task. Bold marks the best method.}
\label{tab:robotwin_c2r}
\small
\setlength{\tabcolsep}{6pt}
\resizebox{\textwidth}{!}{%
\begin{tabular}{lcccccc}
\toprule
Benchmark & GR00T-N1.7~\citep{bjorck2025gr00t} & $\pi_{0.5}$~\citep{black2025pi05} & StarVLA~\citep{community2026starvla} & LingBot-VLA~\citep{wu2026pragmatic} & \textbf{Harness VLA (Codex)} & \textbf{Harness VLA (CC)} \\
\midrule
RoboTwin C2R & $20.7$ & $47.9$ & $10.6$ & $50.4$ & $58.0$ & $\mathbf{58.4}$ \\
\bottomrule
\end{tabular}%
}
\end{table}

\subsection{Experiment Analysis}
\label{sec:exp:analysis}
\begin{figure}[!htb]
\centering
\scriptsize
\setlength{\tabcolsep}{1pt}
\begin{tabular}{cccccc}
\textbf{Obj Std} & \textbf{Obj-T $\piRLinf$} & \textbf{Obj-T Ours} &
\textbf{Goal Std} & \textbf{Goal-S $\piRLinf$} & \textbf{Goal-S Ours} \\
\includegraphics[width=0.16\linewidth]{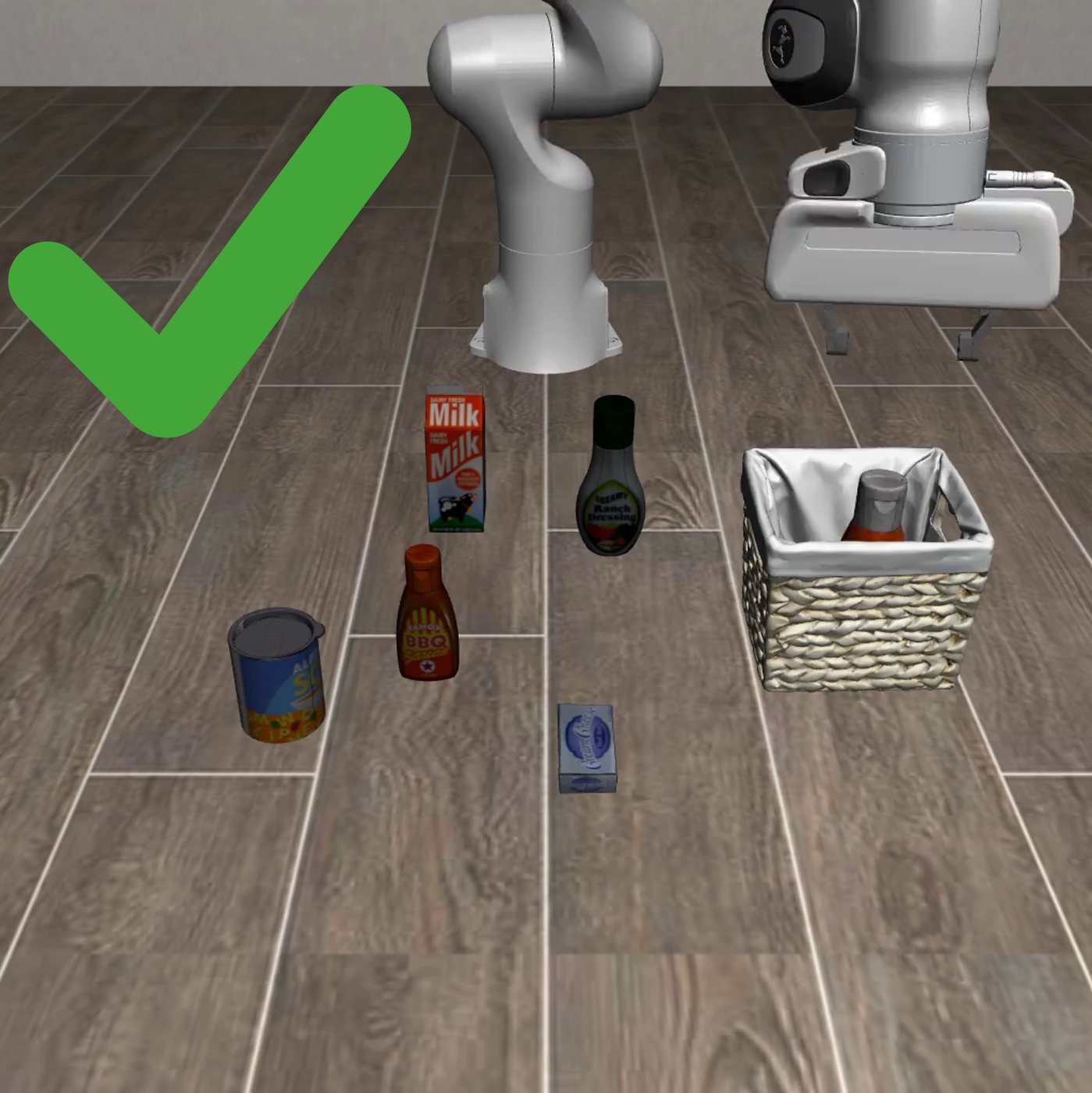} &
\includegraphics[width=0.16\linewidth]{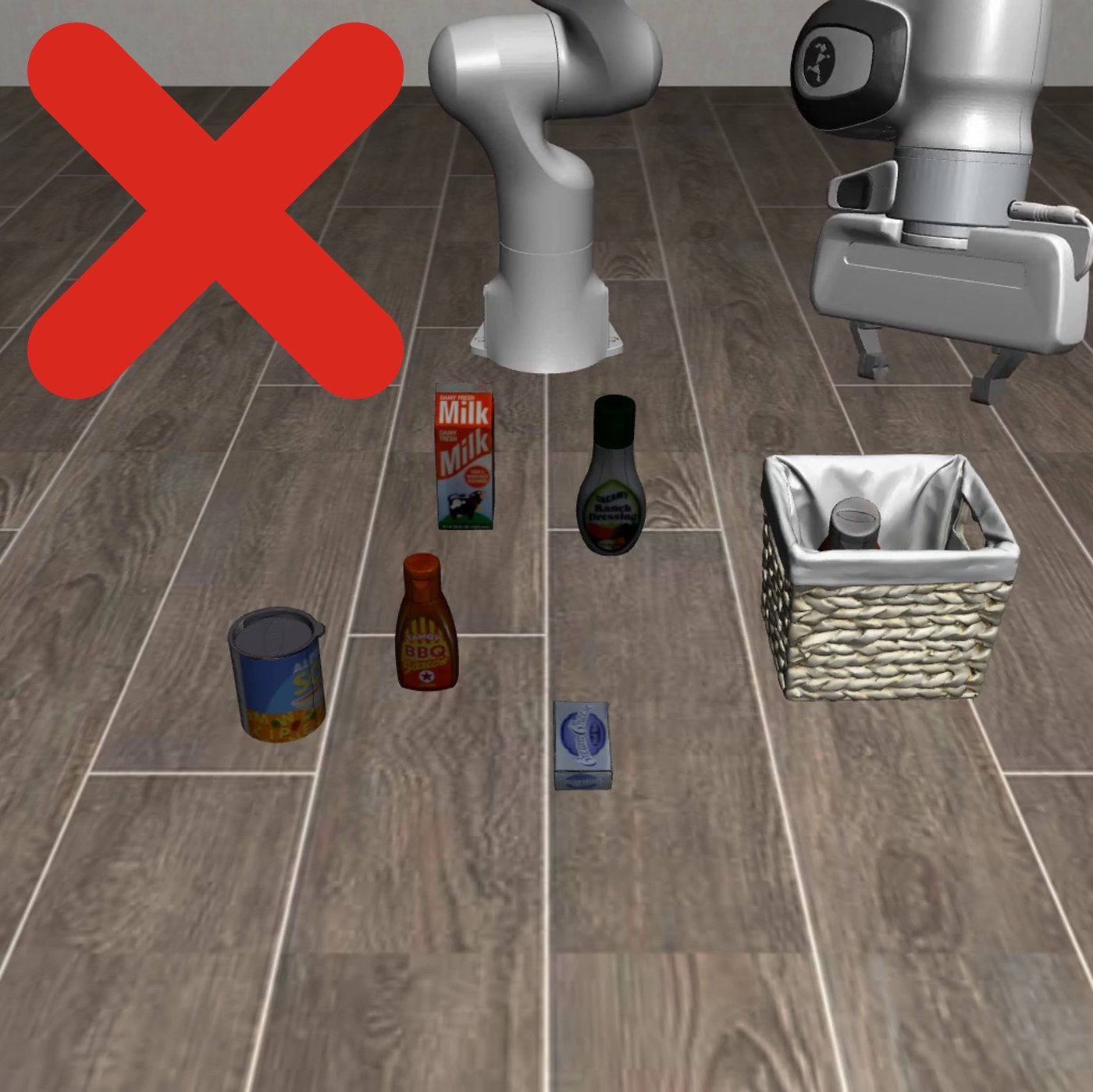} &
\includegraphics[width=0.16\linewidth]{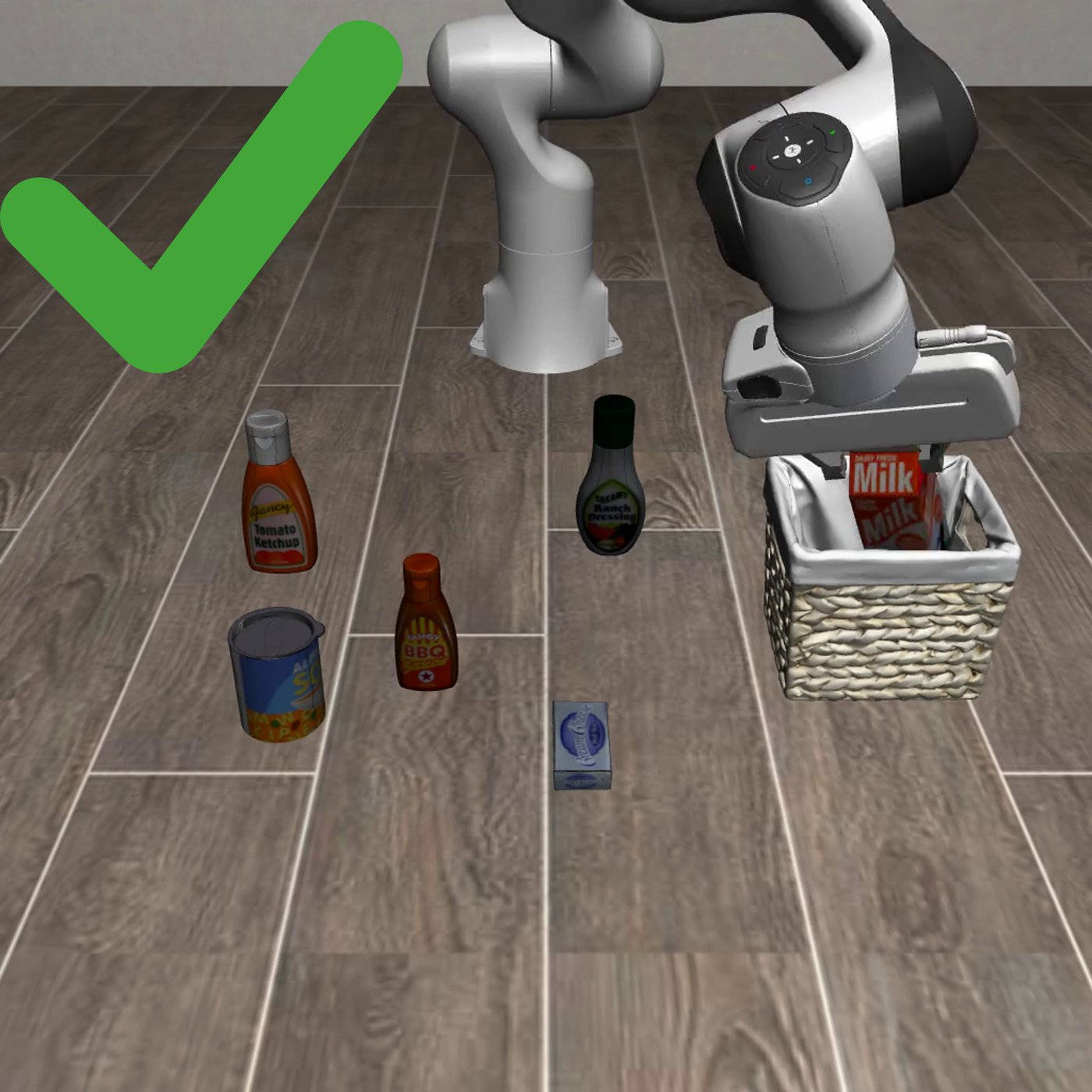} &
\includegraphics[width=0.16\linewidth]{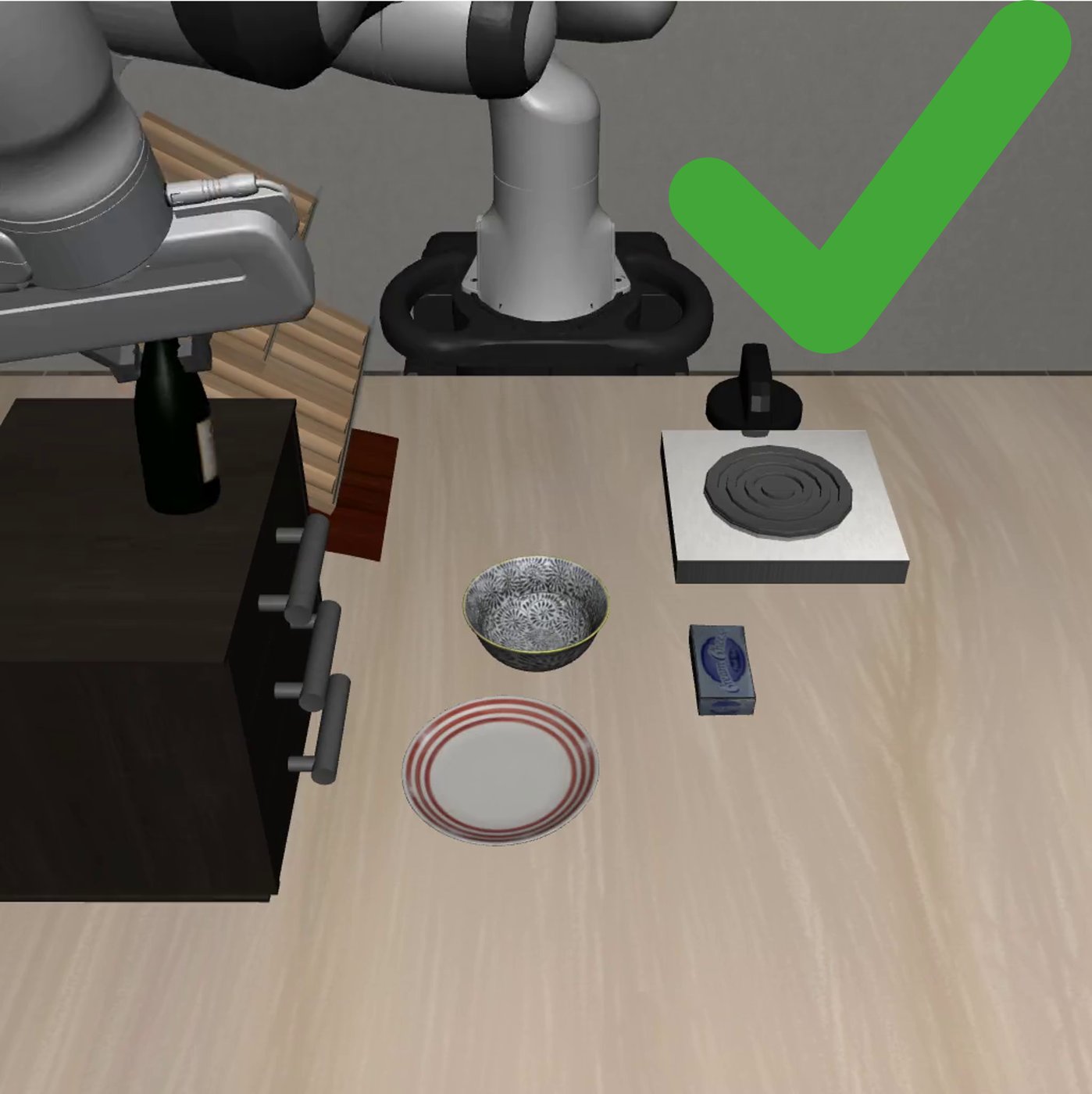} &
\includegraphics[width=0.16\linewidth]{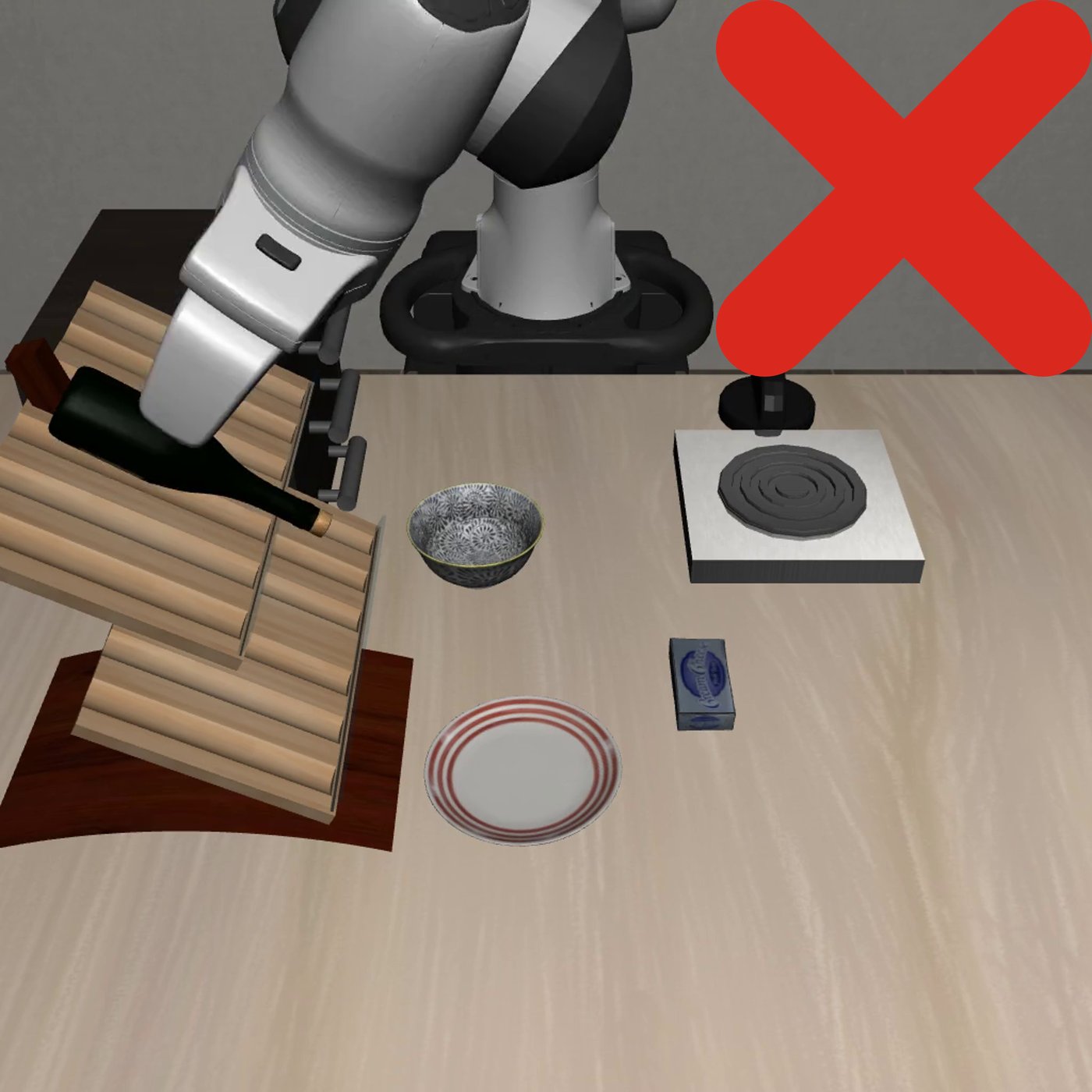} &
\includegraphics[width=0.16\linewidth]{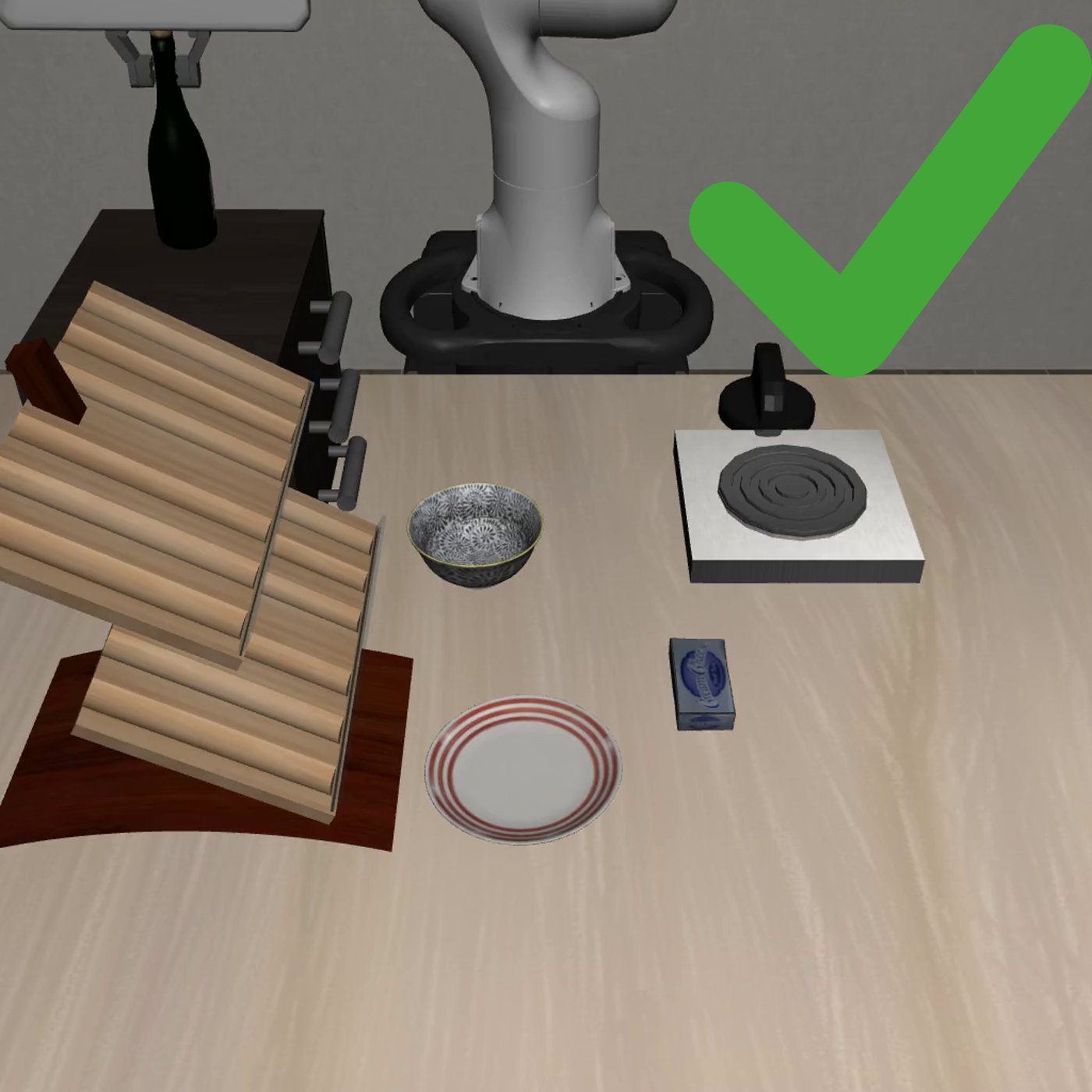}
\end{tabular}
\caption{Terminal-state frames for two LIBERO-Pro cells. The first triplet compares $\piRLinf$ on the standard \textsc{Object} task, $\piRLinf$ on the task-perturbed \textsc{Object-Pro} variant, and Harness VLA on the same perturbed task; when the task description redirects the target while the visual scene remains similar, $\piRLinf$ repeats the standard behavior instead of following the new instruction. The second triplet shows the analogous comparison for a swap-perturbed \textsc{Goal-Pro} task; $\piRLinf$ blindly moves the object toward the training-time region after the layout changes, whereas Harness VLA re-grounds the target through the agentic planner $\Pi$, uses analytic primitives for staging, and invokes \textsc{vla\_act} for the local contact-rich operation.}
\label{fig:generalization}
\end{figure}
We analyze the mechanisms behind these results through three distinct findings. Key Finding~1 focuses on semantic and scene re-grounding by the planner; Key Finding~2 studies planner-staged invocation and retry of \textsc{vla\_act}; and Key Finding~3 studies how analytic primitives isolate non-contact execution from contact-rich control. Unless otherwise noted, the following analyses use Harness VLA (CC) as a representative instantiation, without loss of generality: Codex and CC share the same harness, memory interface, primitive library, frozen VLA interface, and evaluation protocols, differing only in the planner backbone. LIBERO-family analyses use the same $\piRLinf$ checkpoint as the main evaluation, while non-LIBERO analyses use their benchmark-specific frozen VLA primitive.

\paragraph{Key Finding 1: Planner-level semantic re-grounding restores task-conditioned behavior.}
The massive gap between Harness VLA and end-to-end VLAs in Table~\ref{tab:libero_pro_perturbation} is achieved without altering the visuomotor backbone. $\piRLinf$ already solves the standard variants of these tasks, but Figure~\ref{fig:generalization} shows that its behavior is weakly conditioned on the task description and current scene binding. In the task-perturbed \textsc{Object-Pro} case, the visual scene remains similar while the instruction redirects the target, yet $\piRLinf$ repeats the standard behavior instead of following the new task description. In the swap-perturbed \textsc{Goal-Pro} case, the object layout changes, yet $\piRLinf$ still moves the object toward the training-time region. Harness VLA makes semantic grounding explicit at the planner level: the planner $\Pi$ parses the task description, resolves the current contact target from the live RGB-D observation, uses analytic primitives for staging and repositioning, and invokes or re-invokes \textsc{vla\_act} only for the local contact-rich phase (Key Finding~2). Thus, semantic and scene-level reasoning is handled by the planner, while the frozen VLA remains responsible only for executing the contact-rich operation under the planner-provided binding.

\paragraph{Key Finding 2: Planner-staged VLA invocation improves frozen-policy reliability.} In Harness VLA, the planner does not call the VLA as a one-shot black box. Although the VLA is invoked sparsely rather than as a continuous controller, each call is a planner-chosen local contact-rich attempt whose staging can determine whether the frozen policy succeeds. The planner $\Pi$ therefore treats \textsc{vla\_act} as a local contact-rich primitive whose invocation can be re-staged and retried. Given a desired contact target (the object, fixture, or local interaction region to be acted on), the planner uses analytic primitives to place the robot in a feasible pre-contact configuration, invokes the VLA, observes the resulting contact state, and decides whether the next step should continue the task or reframe the local attempt.
\begin{figure}[htbp]
    \centering

    \begin{subfigure}{0.33\linewidth}
        \centering
        \includegraphics[width=\linewidth]{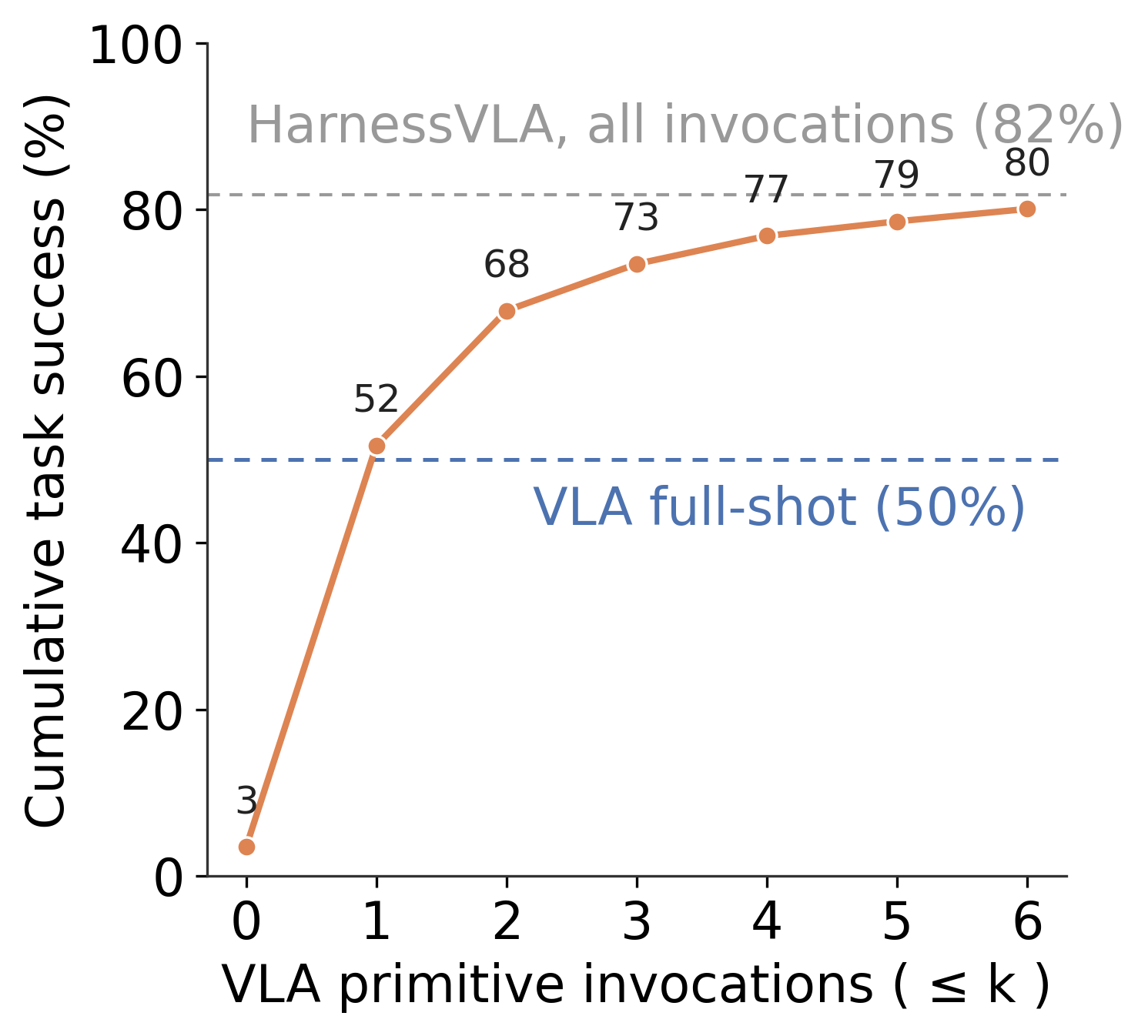}
        \caption{\textsc{LIBERO-Pro}}
    \end{subfigure}
    \hfill
    \begin{subfigure}{0.32\linewidth}
        \centering
        \includegraphics[width=\linewidth]{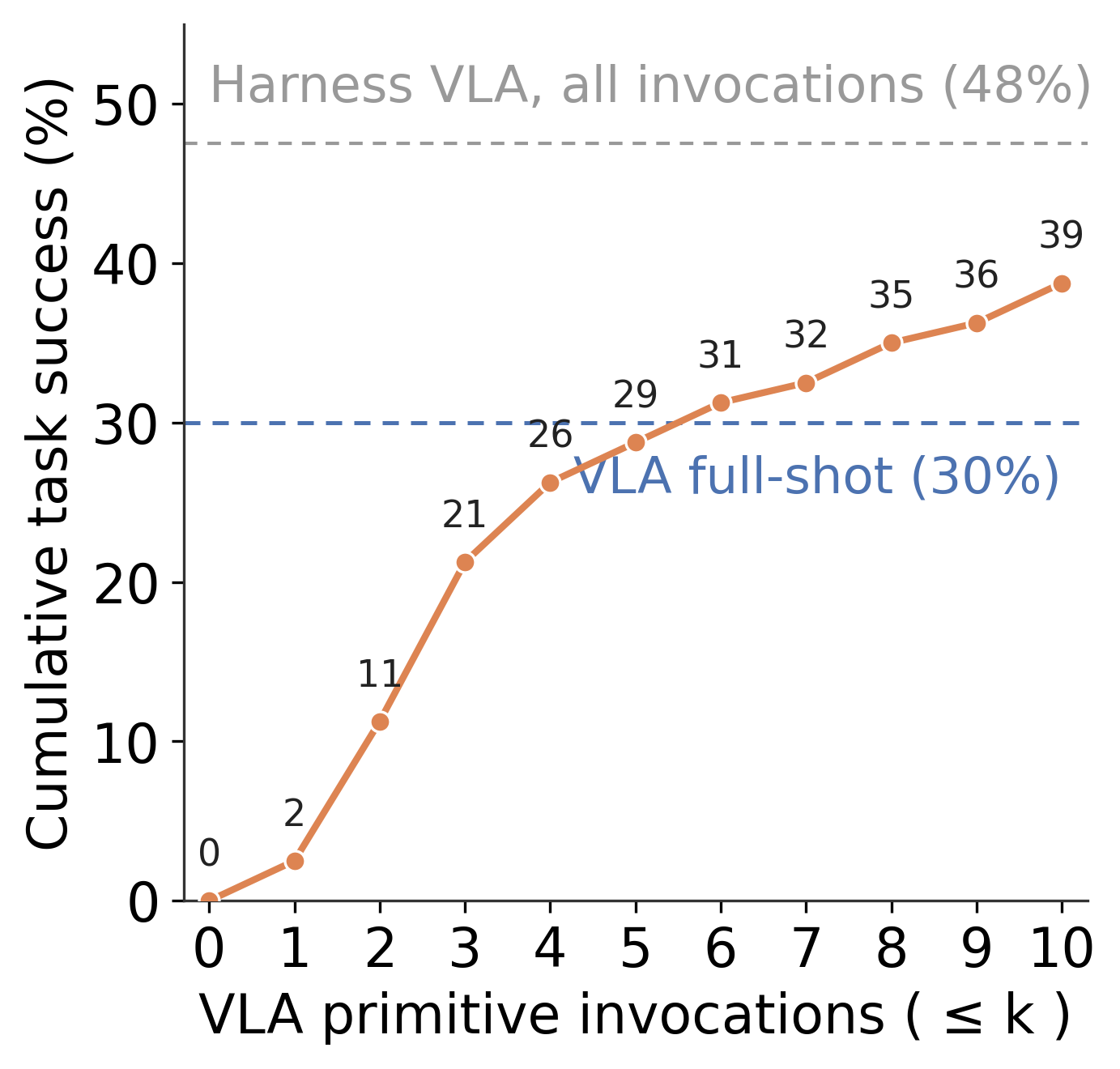}
        \caption{\textsc{RoboCasa365}}
    \end{subfigure}
    \hfill
    \begin{subfigure}{0.32\linewidth}
        \centering
        \includegraphics[width=\linewidth]{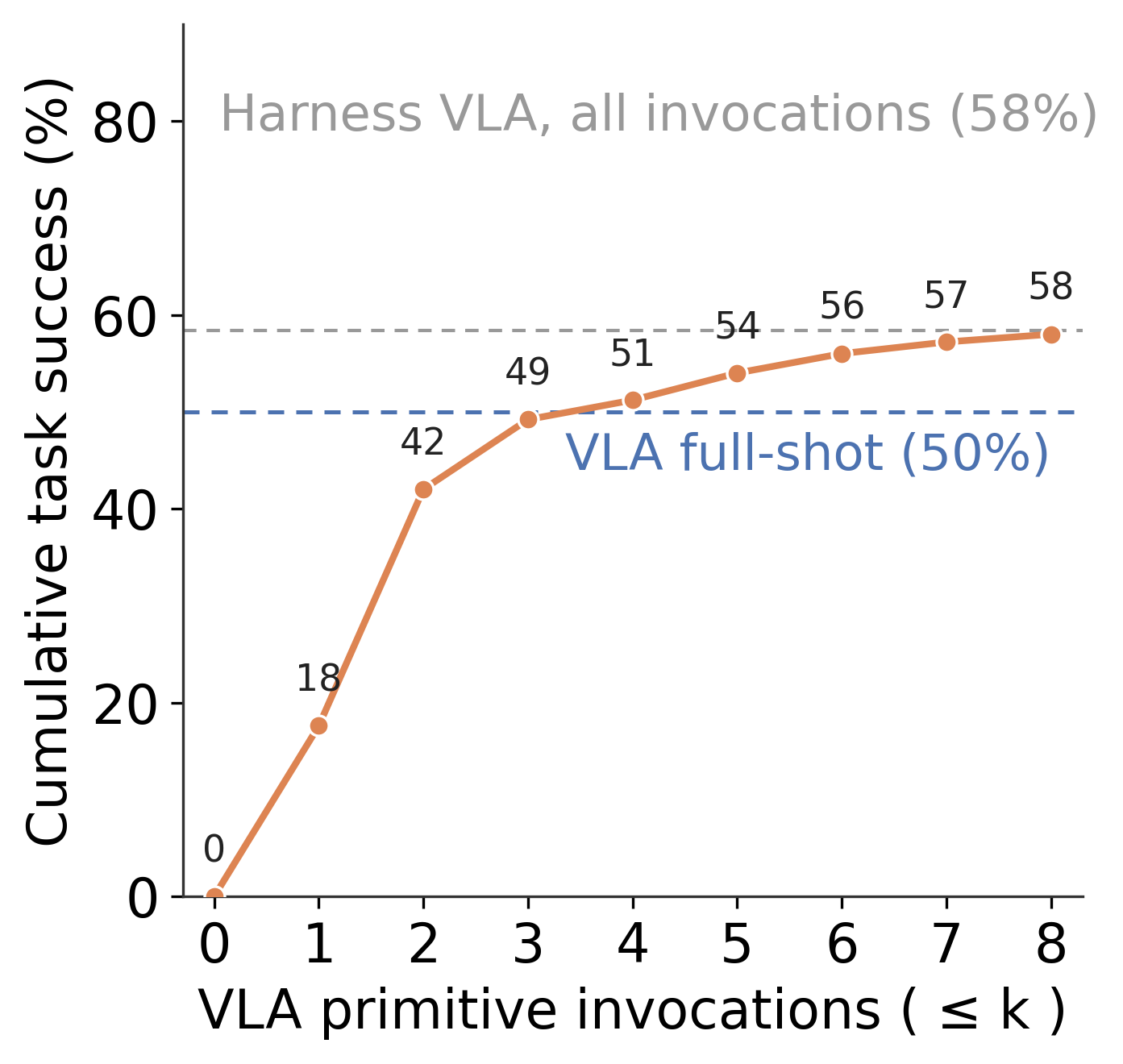}
        \caption{\textsc{RoboTwin C2R}}
    \end{subfigure}

    \caption{\textbf{Adaptive VLA invocation improves success across benchmarks.} Each panel plots cumulative task success as a function of the maximum number of VLA primitive invocations allowed per episode. The blue dashed line marks the corresponding frozen-policy baseline, while the gray dashed line marks full Harness VLA performance with all planner-selected invocations. Across LIBERO-Pro, RoboCasa365, and RoboTwin C2R, success rises rapidly after the first few VLA calls and then saturates toward the full harness result, indicating that repeated planner-staged invocation is useful but remains sparse.}
    \label{fig:vla_invocation_analysis}
\end{figure}
\paragraph{First, staging restores a VLA-compatible local state.}
Under deployment perturbations such as semantic retargeting and spatial-layout shifts, the original VLA viewpoint or pre-contact pose may no longer expose the correct contact target in a familiar configuration. By re-staging the robot around the current scene, the agentic planner $\Pi$ brings the target back into a VLA-compatible local observation while preserving the correct semantic binding. This mechanism explains why the harness can improve a frozen VLA without changing its parameters: it learns where the VLA should begin acting, rather than asking the policy to absorb the full distribution shift by itself.

\paragraph{Second, retry localizes contact failures.}
Because VLA execution is stochastic and short-horizon contact is physically brittle, a single failed attempt need not terminate the entire rollout. Harness VLA localizes such errors to the current contact-rich subtask: the planner can observe an incomplete or unstable outcome, re-stage the robot, and re-invoke \textsc{vla\_act}, rather than letting a transient failure propagate through a monolithic long-horizon policy. Thus, repeated VLA calls are not continuous low-level control; they are sparse, planner-selected attempts that make contact-rich execution recoverable.

\begin{figure}[htbp]
    \centering

    \includegraphics[width=0.155\linewidth]{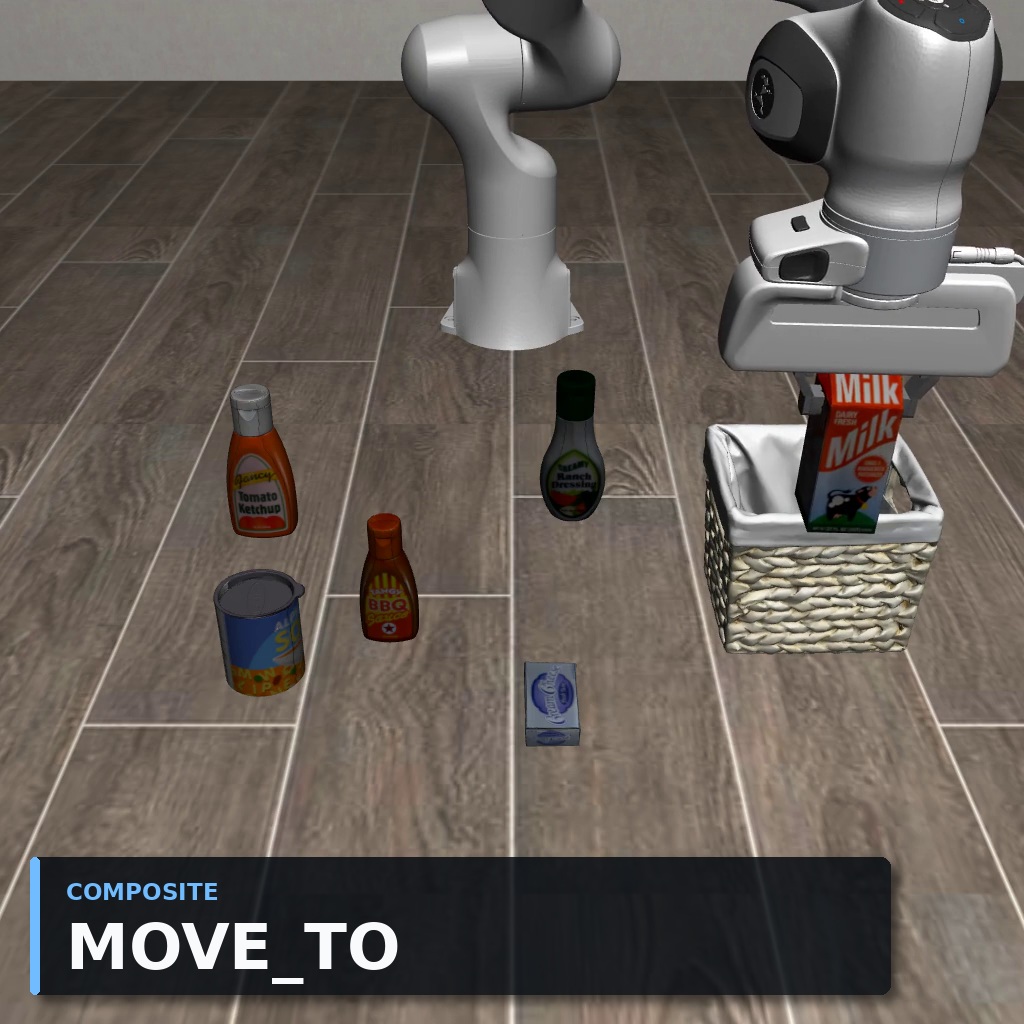}
    \hfill
    \includegraphics[width=0.155\linewidth]{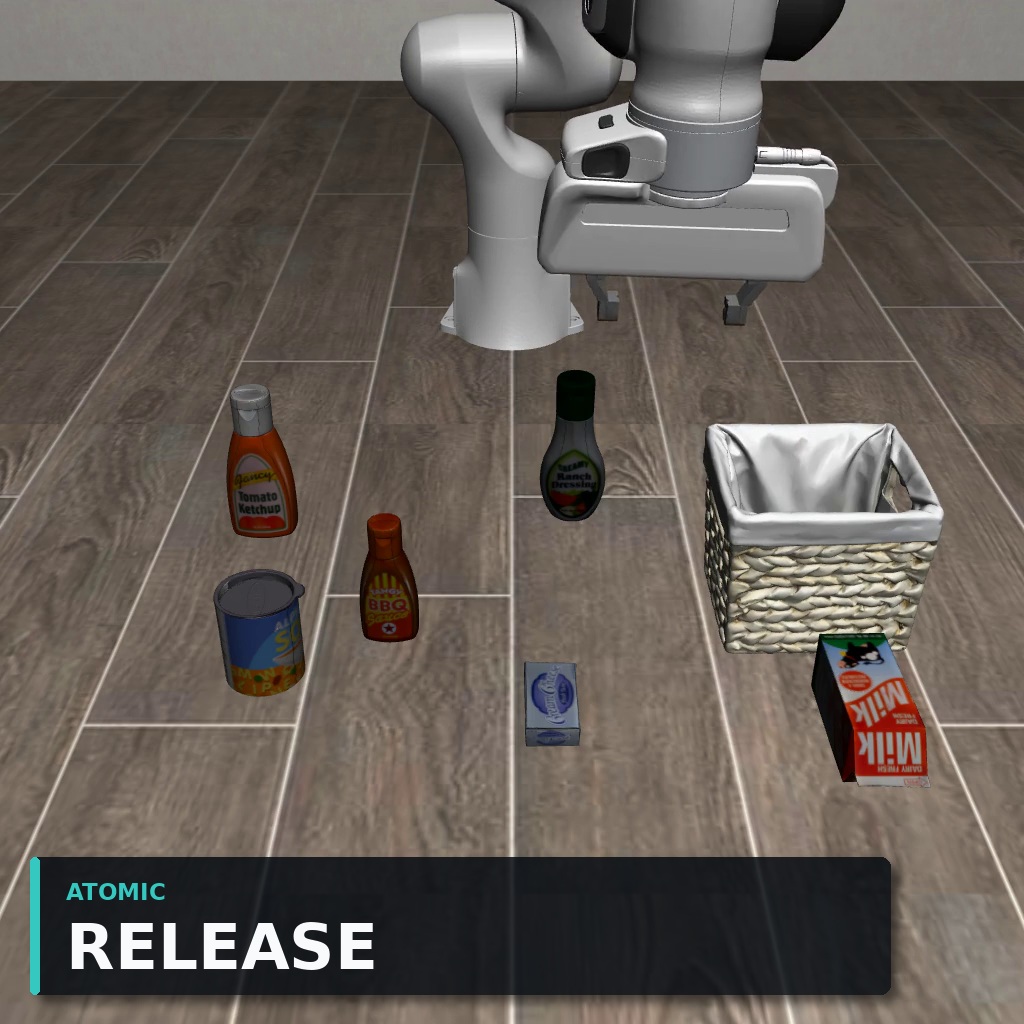}
    \hfill
    \includegraphics[width=0.155\linewidth]{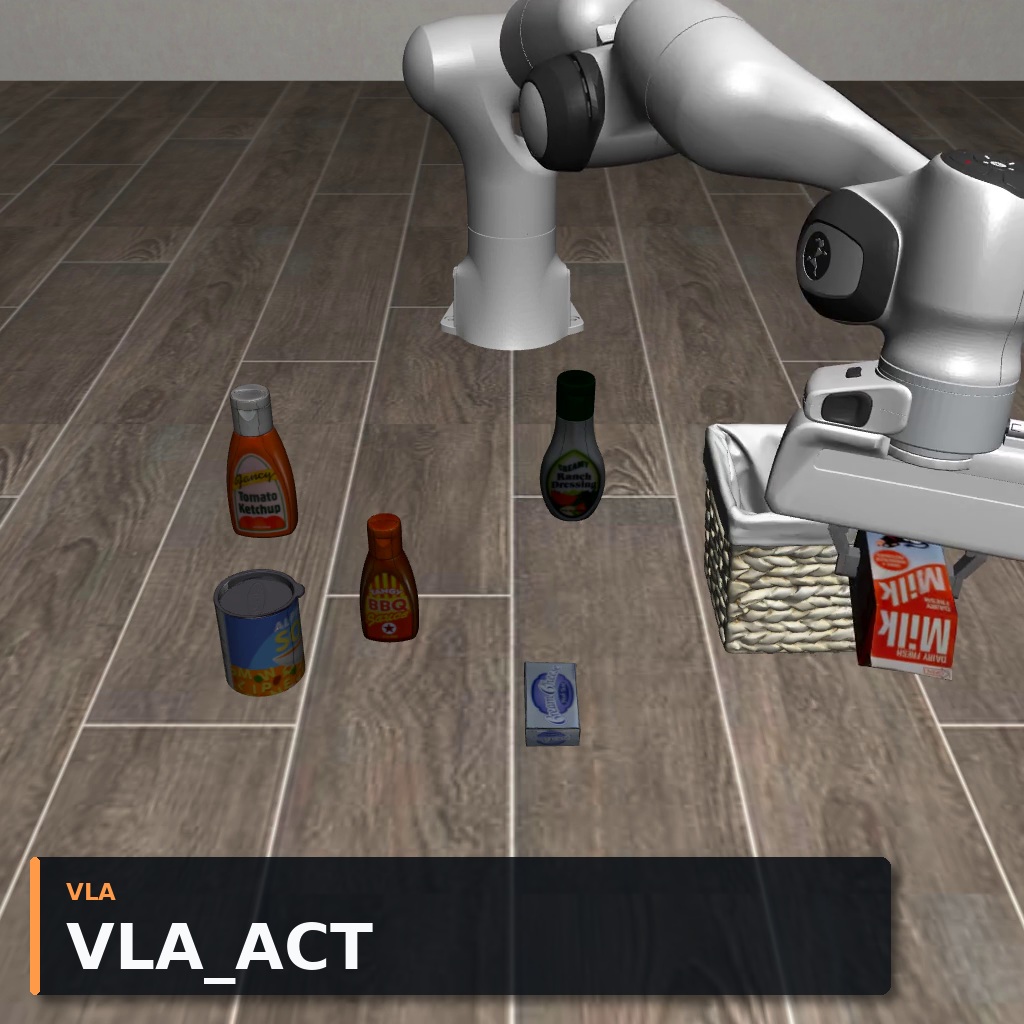}
    \hfill
    \includegraphics[width=0.155\linewidth]{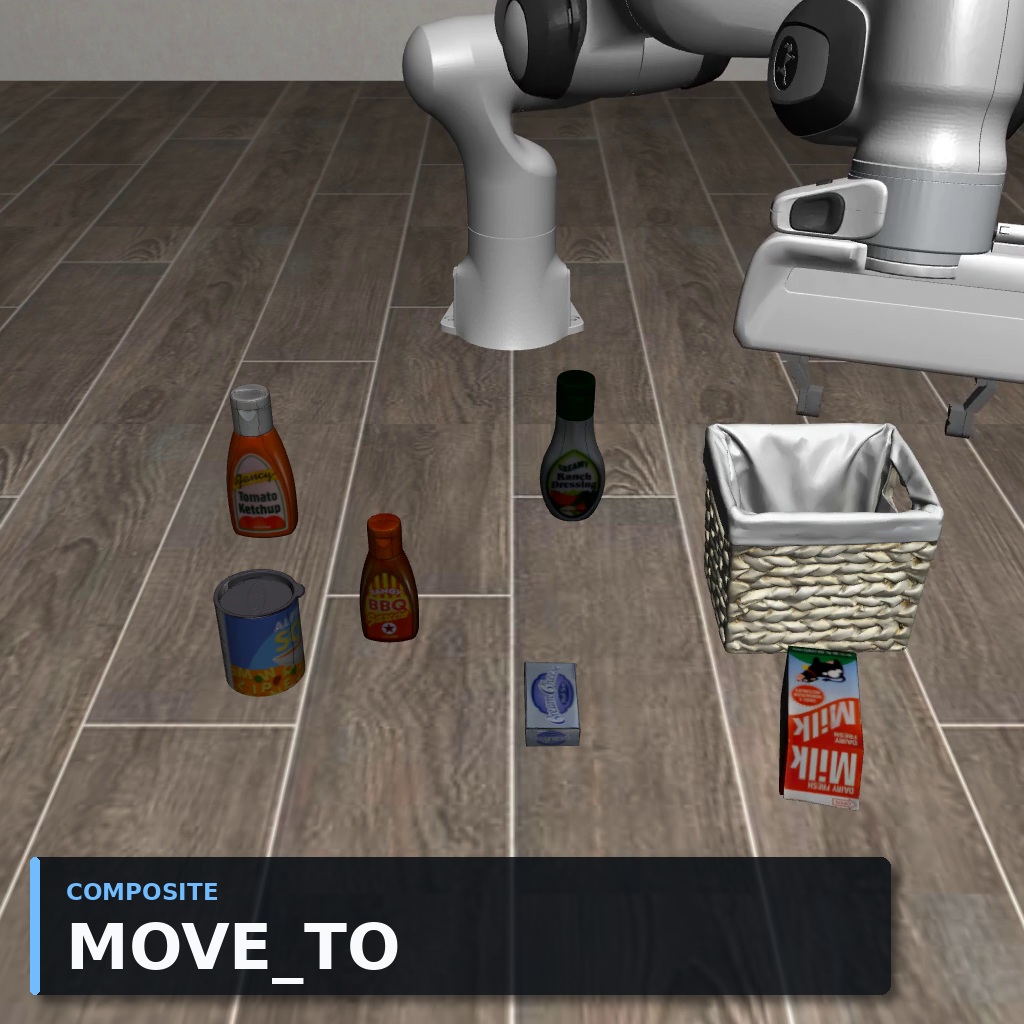}
    \hfill
    \includegraphics[width=0.155\linewidth]{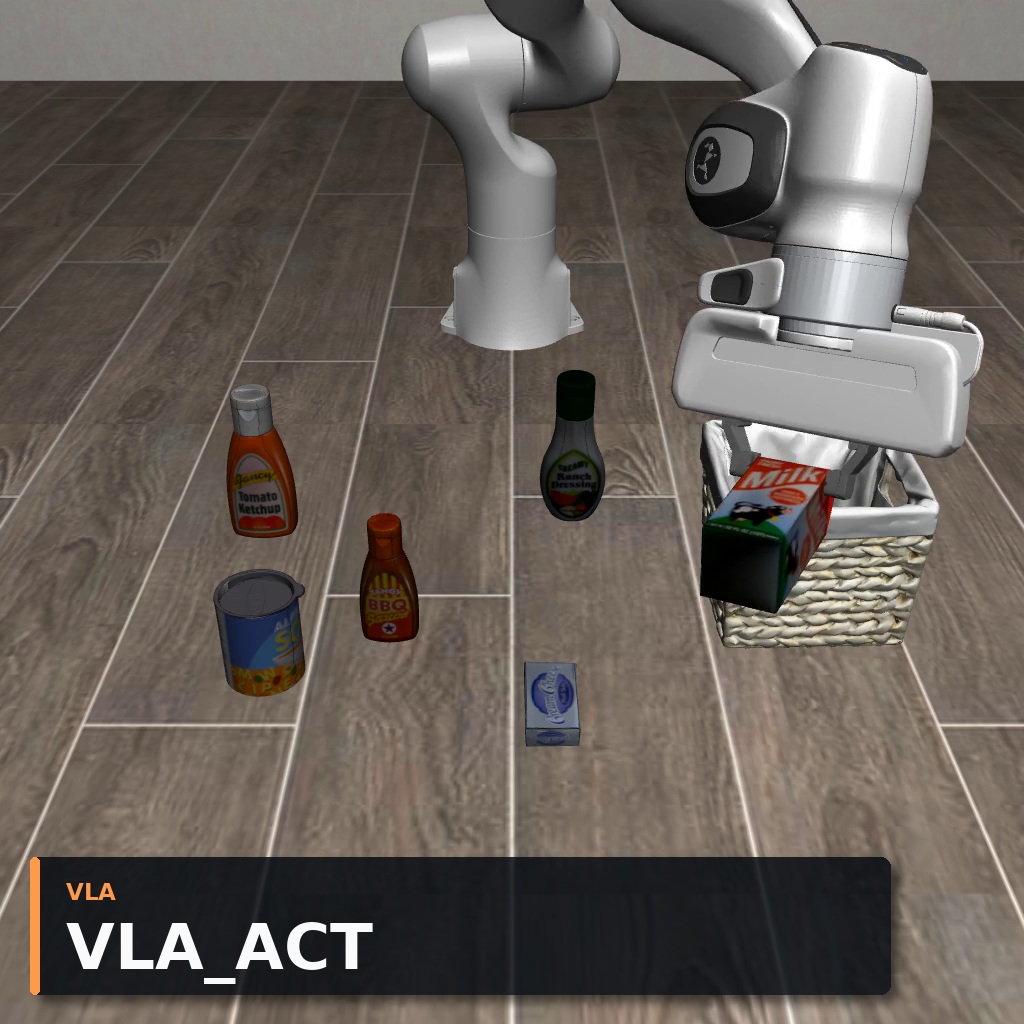}
    \hfill
    \includegraphics[width=0.155\linewidth]{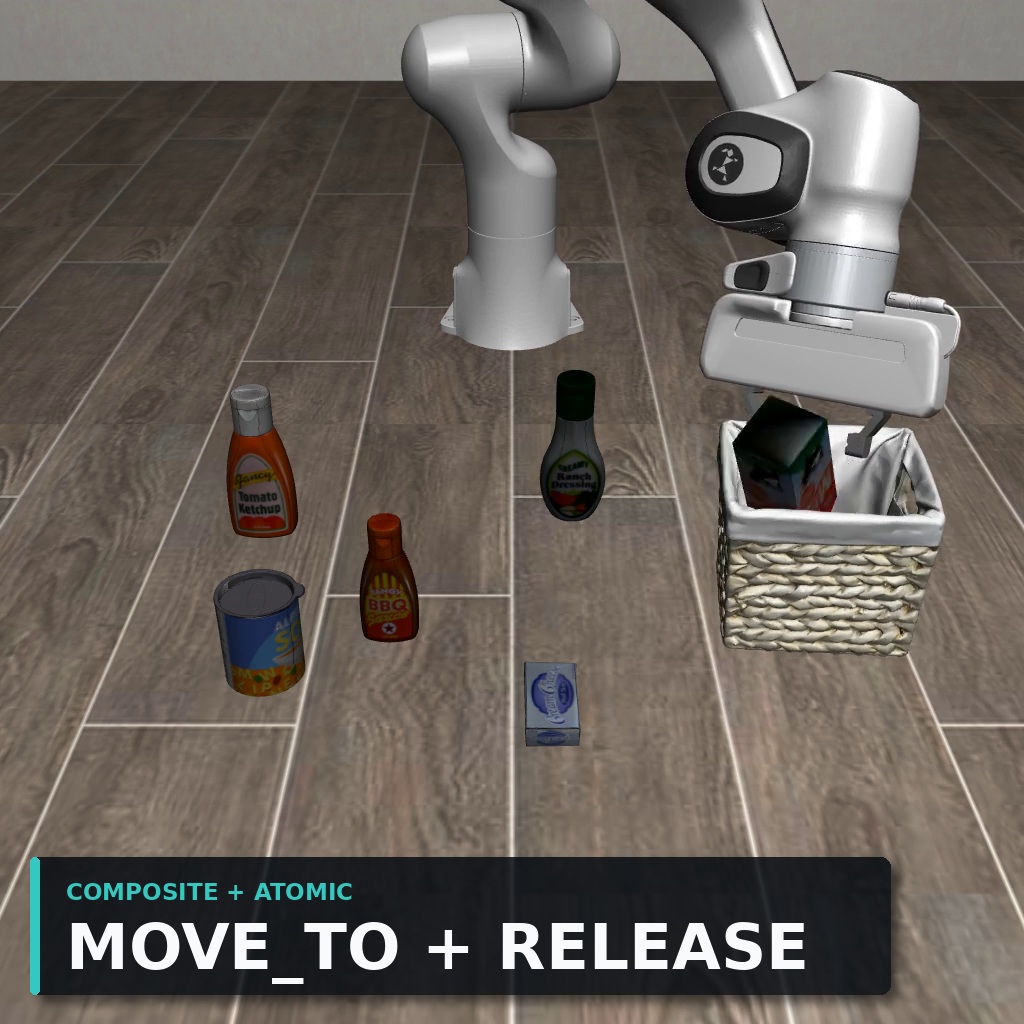}

    \vspace{0.15cm}

    \includegraphics[width=0.155\linewidth]{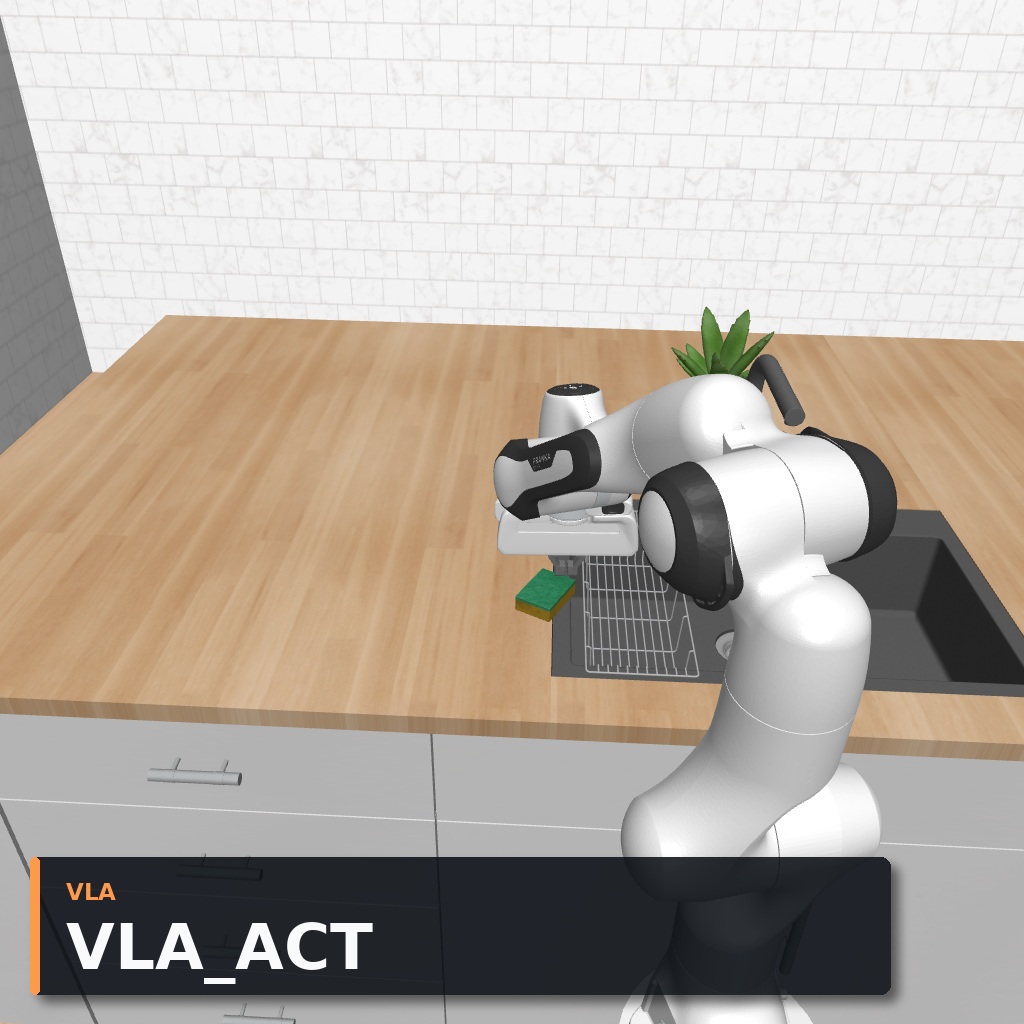}
    \hfill
    \includegraphics[width=0.155\linewidth]{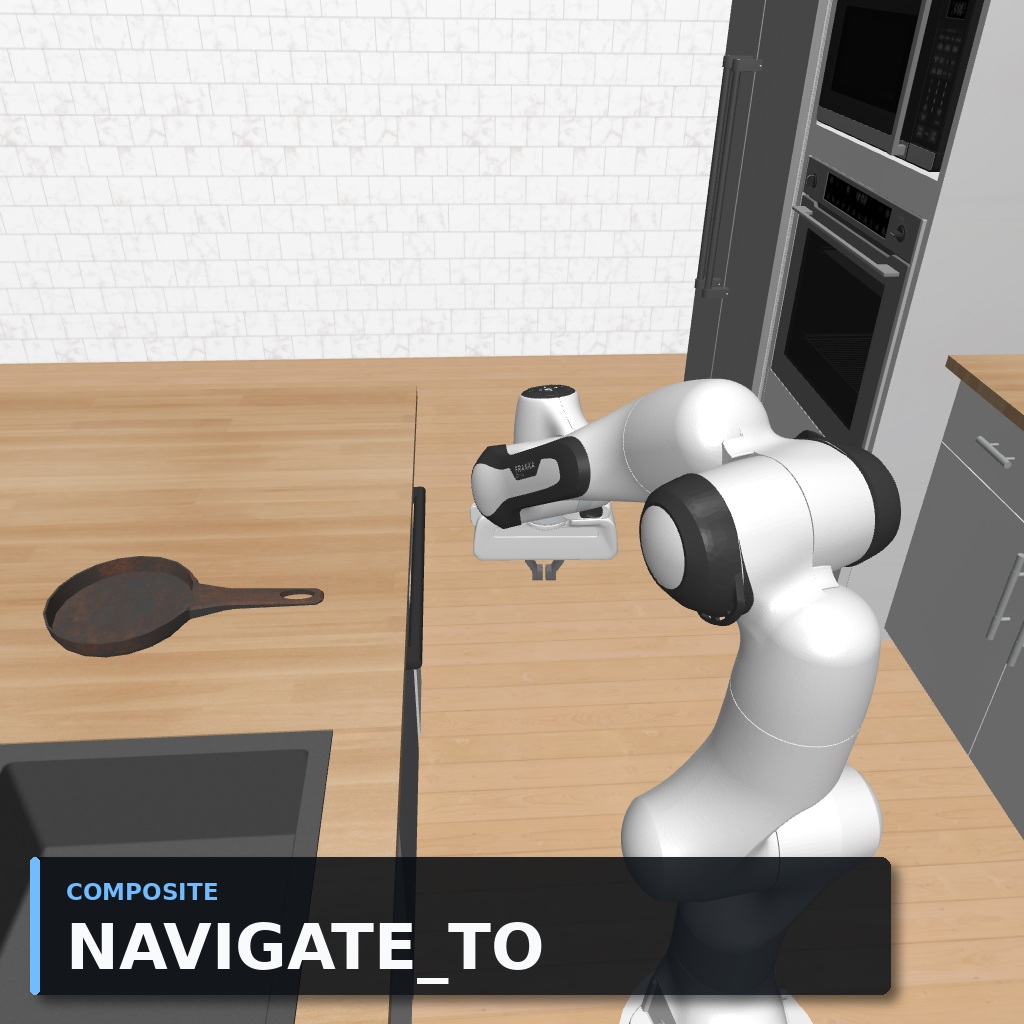}
    \hfill
    \includegraphics[width=0.155\linewidth]{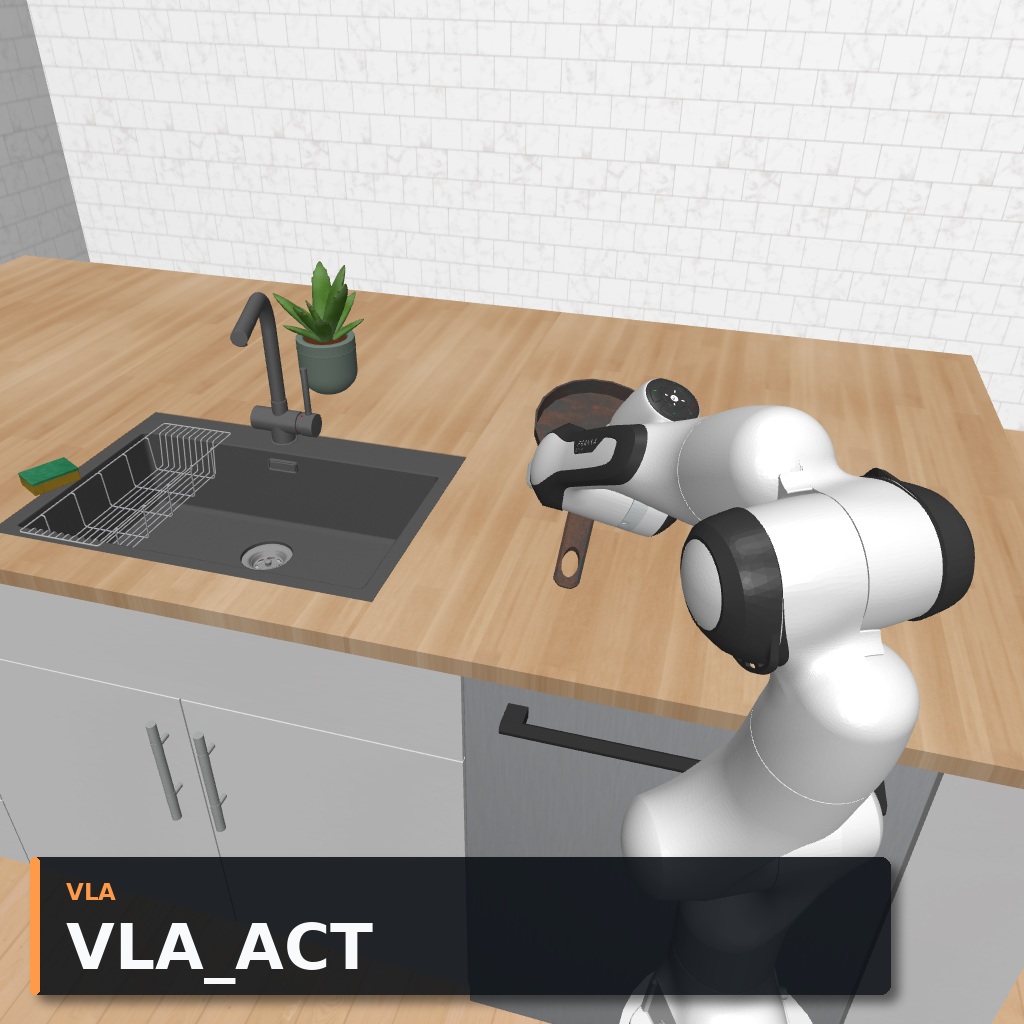}
    \hfill
    \includegraphics[width=0.155\linewidth]{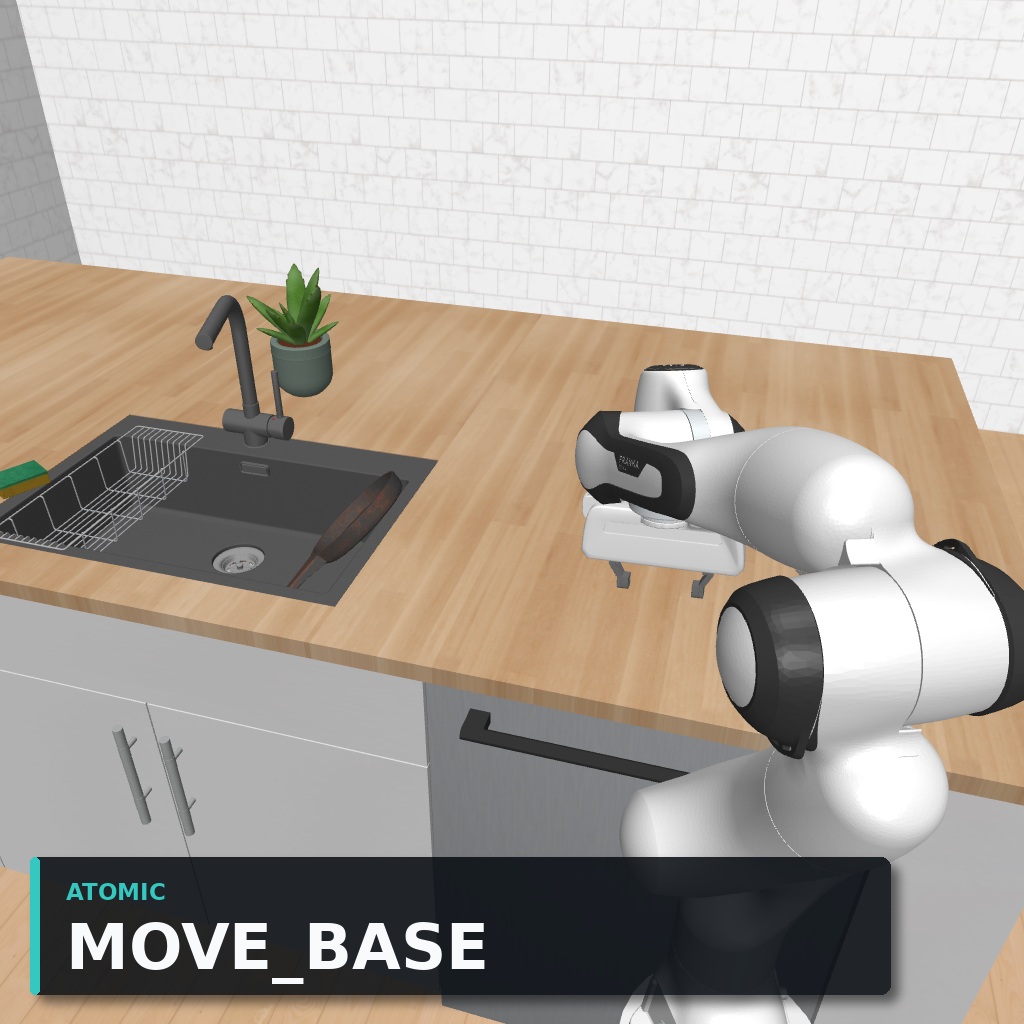}
    \hfill
    \includegraphics[width=0.155\linewidth]{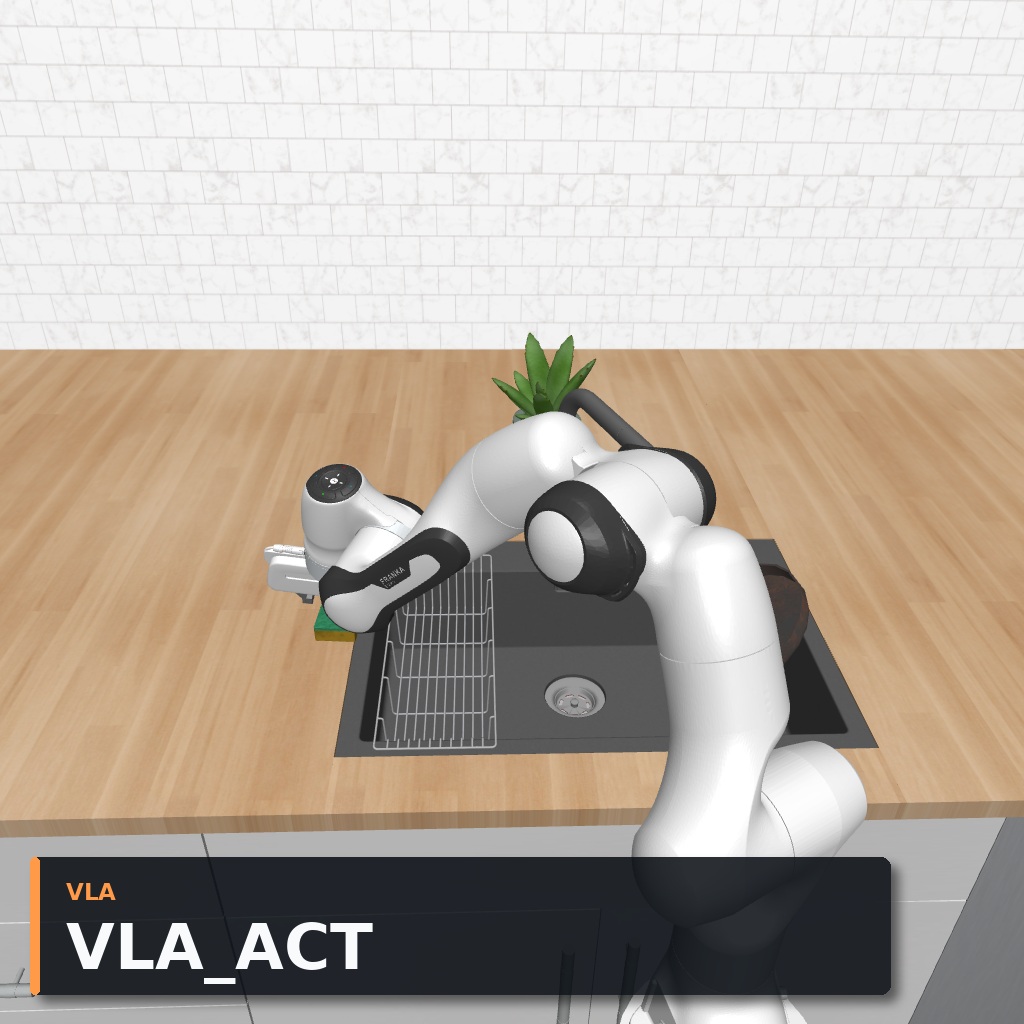}
    \hfill
    \includegraphics[width=0.155\linewidth]{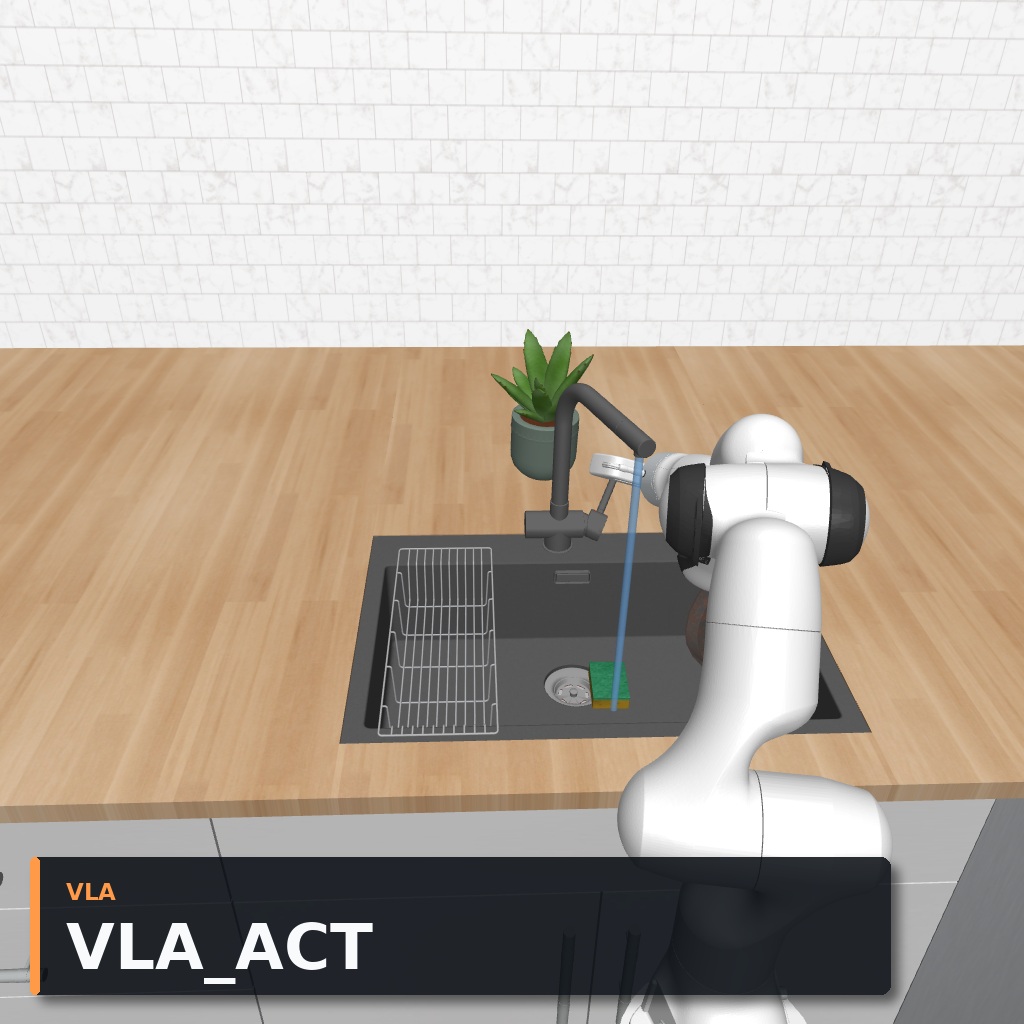}

    \caption{\textbf{Representative rollout frames for adaptive VLA invocation.} Top row: a Harness VLA rollout on \textsc{LIBERO-Pro Object} task 4. The planner repeatedly invokes \textsc{vla\_act} around the milk carton after intermediate grasping or placement attempts leave the object outside or only partially inside the basket; after re-staging the end-effector and retrying the local contact-rich operation, the milk carton is finally placed stably inside the basket. Bottom row: a RoboCasa365 \textsc{PreSoakPan} rollout. The planner adjusts the mobile base and arm pose around the pan, retries \textsc{vla\_act} until a stable grasp is obtained, places the pan into the sink, and later invokes \textsc{vla\_act} again to actuate the faucet. These examples show that repeated VLA calls are not continuous control, but planner-selected contact attempts embedded inside analytic navigation, staging, and verification.}
    \label{fig:vla_invocation_rollout_frames}
\end{figure}

Figure~\ref{fig:vla_invocation_analysis} provides the aggregate evidence for this effect by capping the maximum number of VLA primitive invocations allowed in an episode. A small number of planner-selected invocations already exceeds the corresponding frozen-policy baseline, while additional invocations further improve success on longer or more contact-heavy tasks. Figure~\ref{fig:vla_invocation_rollout_frames} gives representative case studies behind this curve: the planner observes an incomplete or unstable contact outcome, re-stages the robot or base, and calls \textsc{vla\_act} again for the next local contact attempt. Together, these results show that the VLA is used sparsely, but the ability to re-stage and invoke it again is central to the robustness of the harness.

\paragraph{Key Finding 3: Analytic primitives isolate non-contact execution from contact-rich control.} Analytic primitives do not replace the VLA on contact-rich operations. Instead, they handle the surrounding non-contact structure of the task: free-space transport, pre-contact staging, wrist or base reorientation, retreat, and post-contact repositioning. This lets the planner reserve \textsc{vla\_act} for the local contact-rich phases where learned visuomotor control is needed, including grasping, constrained placement, button pressing, faucet turning, drawer manipulation, and coffee-machine operation.

Once the robot has established stable contact with the target, the planner $\Pi$ can use analytic primitives to move, rotate, or navigate the robot toward the next relevant region, while invoking the VLA again when the next contact-rich phase begins. Appendix~\ref{app:primitive_stats} reports the primitive usage statistics. Thus, the analytic vocabulary does not solve contact-rich manipulation by itself; it expands the conditions under which the same frozen VLA can be reused. By handling the non-contact context around each local interaction, the planner prevents the VLA from being responsible for long-horizon composition, scene-level grounding, and every intermediate motion in the rollout.

\begin{figure}[htbp]
    \centering
    \includegraphics[width=\linewidth,height=0.24\textheight,keepaspectratio]{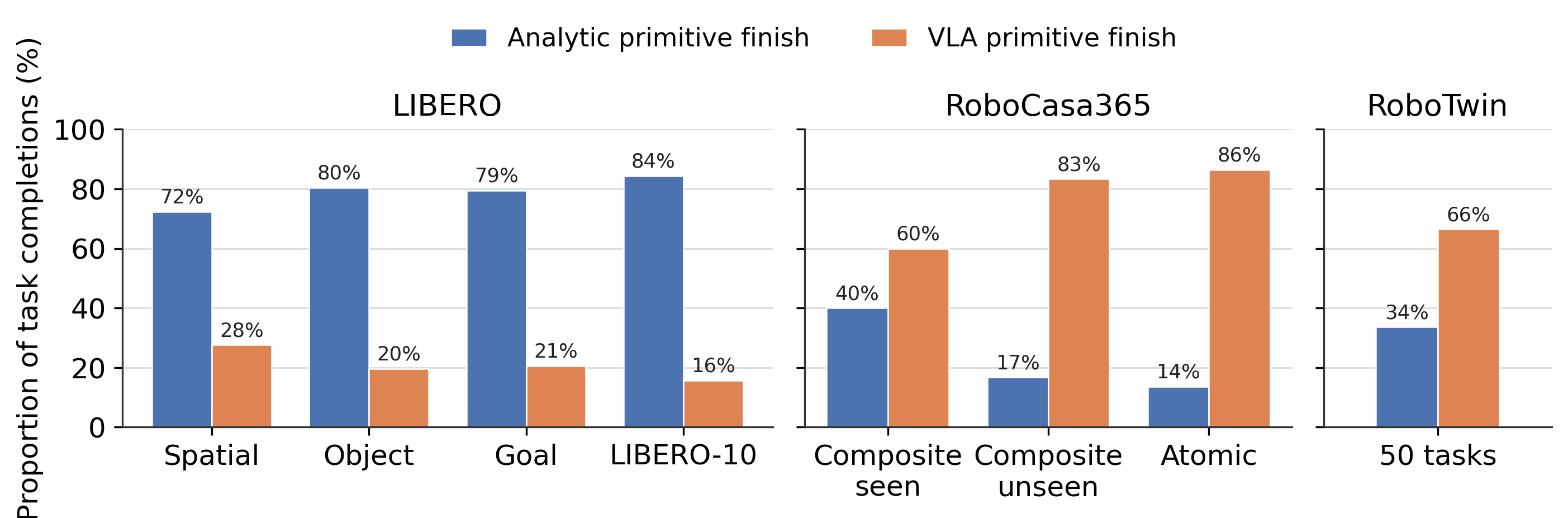}
    \caption{\textbf{Task completion attribution across benchmarks.} Bars show the fraction of successful rollouts whose final benchmark completion predicate fires after an analytic primitive (blue) or after a VLA primitive (orange). LIBERO Pro-family tasks are mostly finished by analytic primitives after the VLA has established stable contact, whereas RoboCasa365 and RoboTwin C2R contain more terminal contact-rich operations such as fixture actuation, constrained placement, or bimanual object interaction.}
    \label{fig:finisher_attribution}
\end{figure}

\begin{figure}[htbp]
    \centering

    \includegraphics[width=0.155\linewidth]{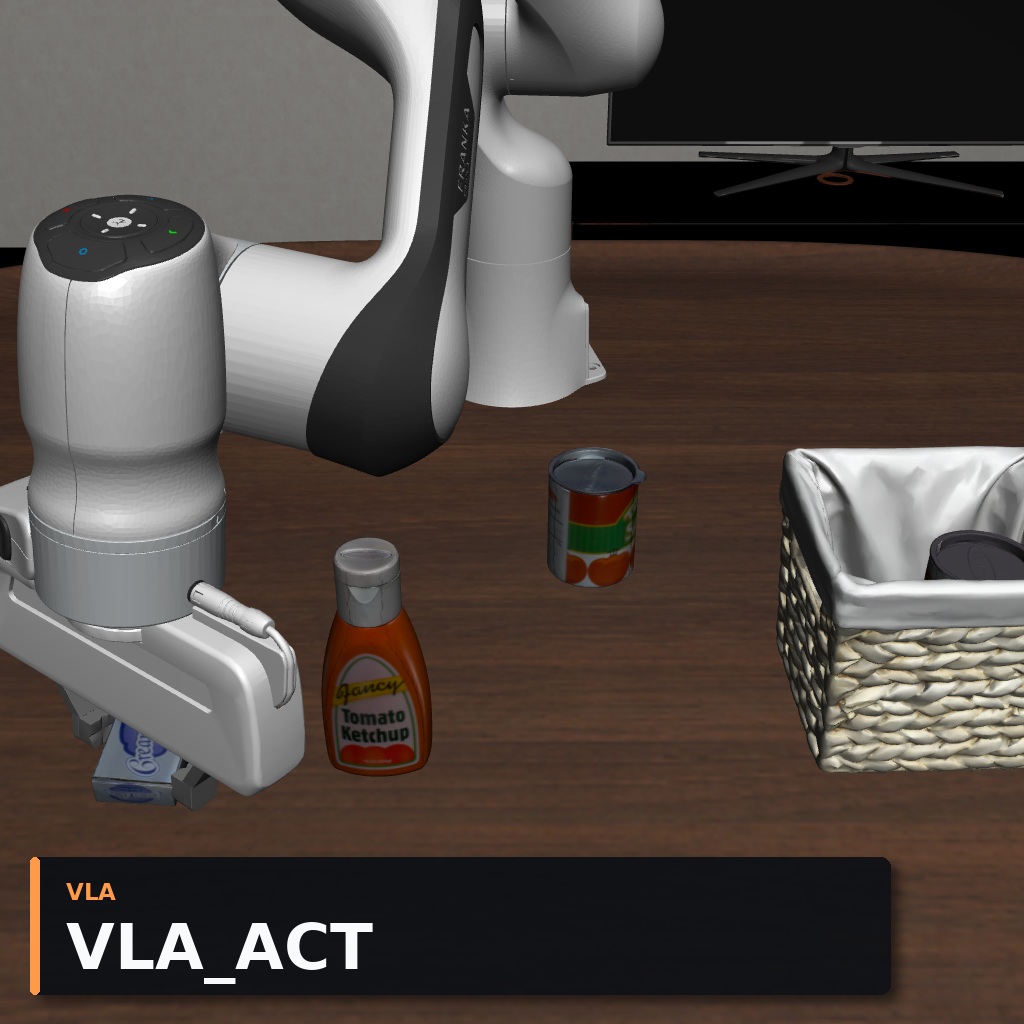}
    \hfill
    \includegraphics[width=0.155\linewidth]{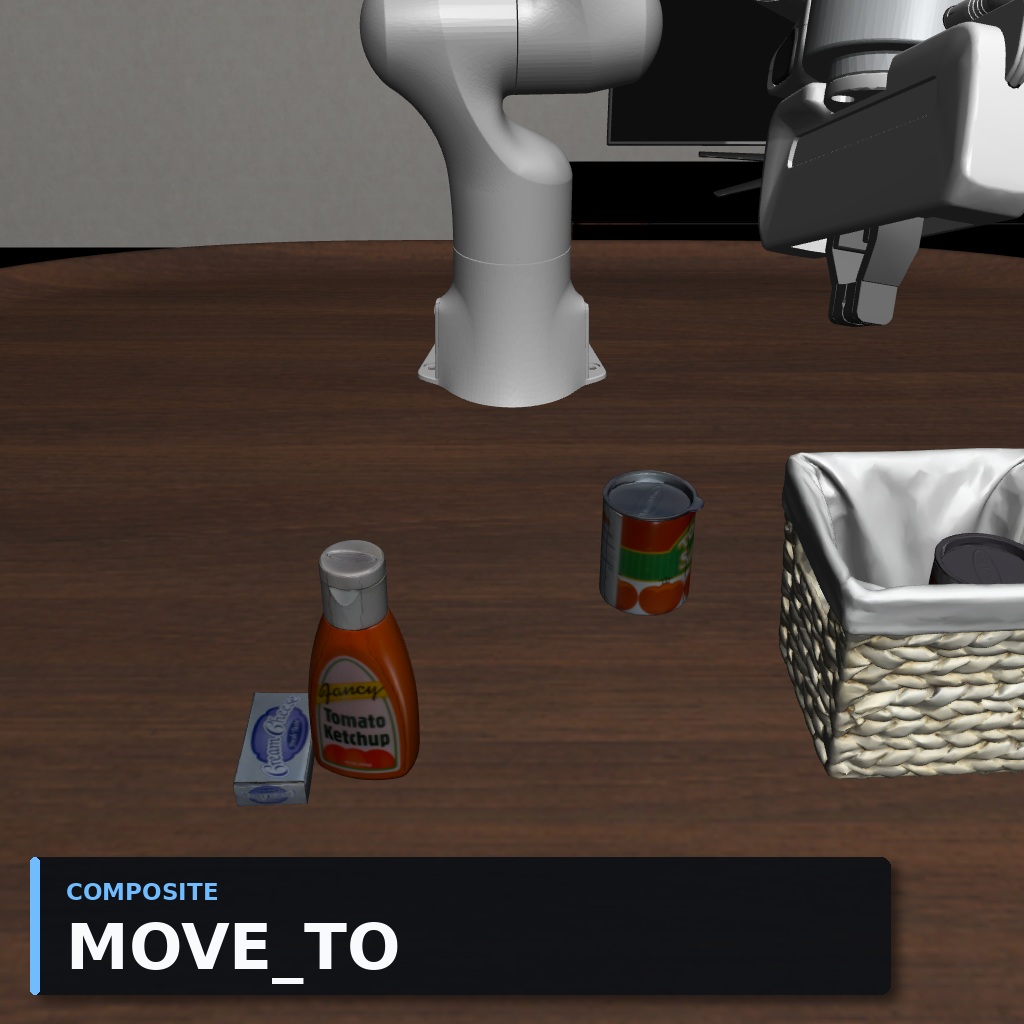}
    \hfill
    \includegraphics[width=0.155\linewidth]{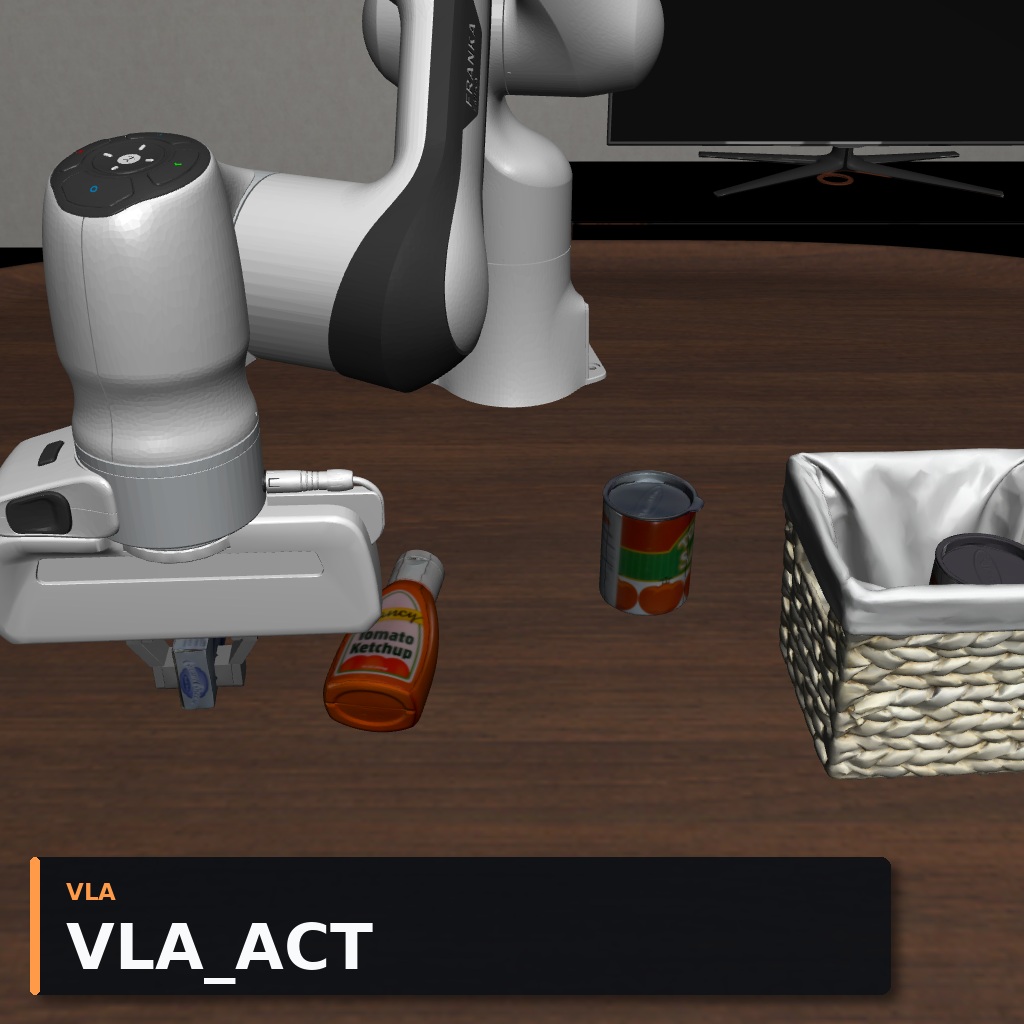}
    \hfill
    \includegraphics[width=0.155\linewidth]{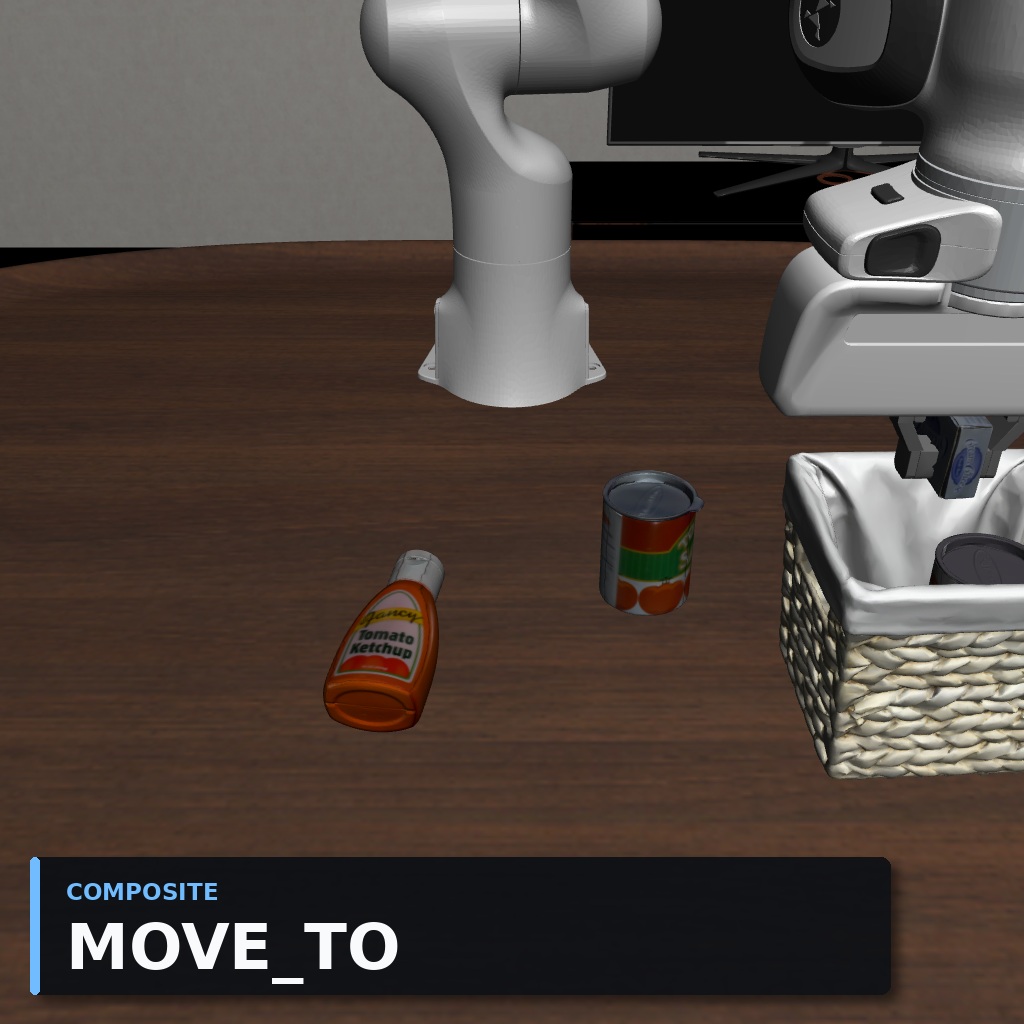}
    \hfill
    \includegraphics[width=0.155\linewidth]{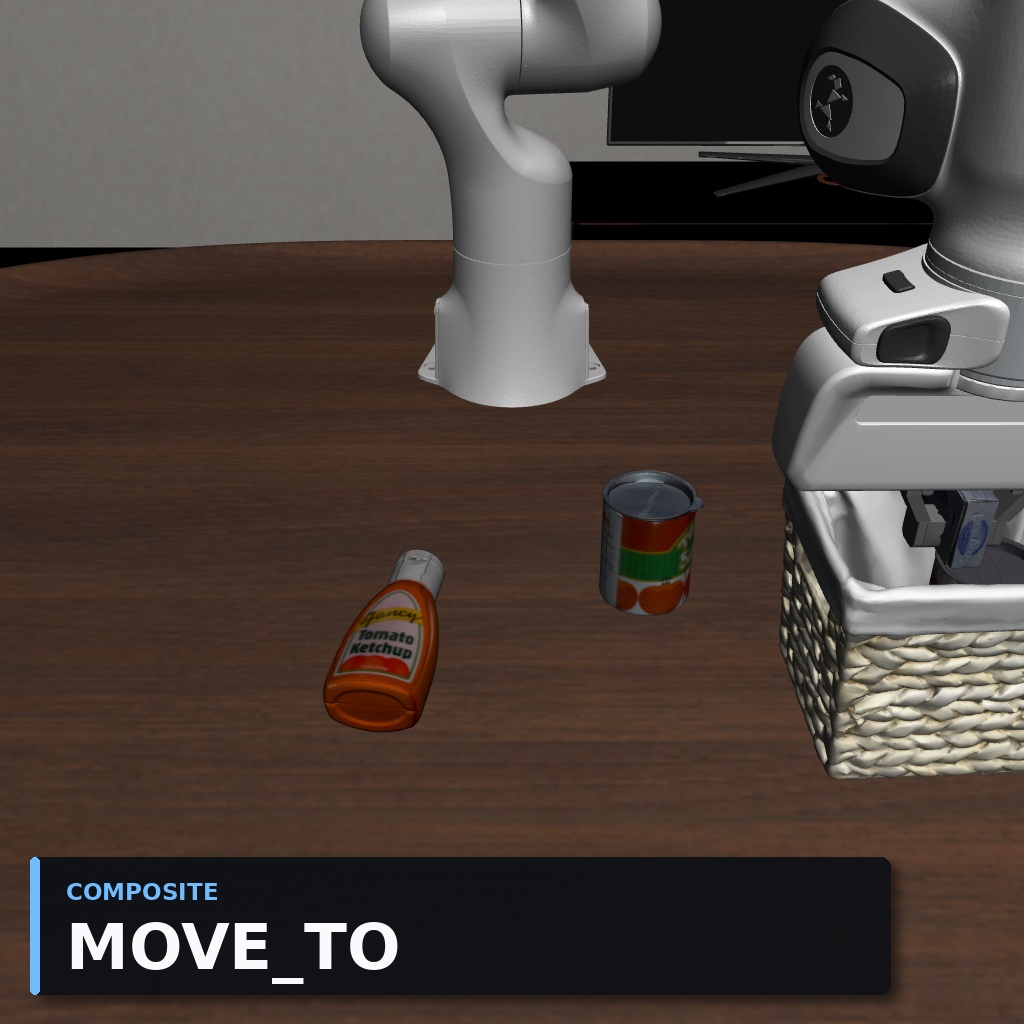}
    \hfill
    \includegraphics[width=0.155\linewidth]{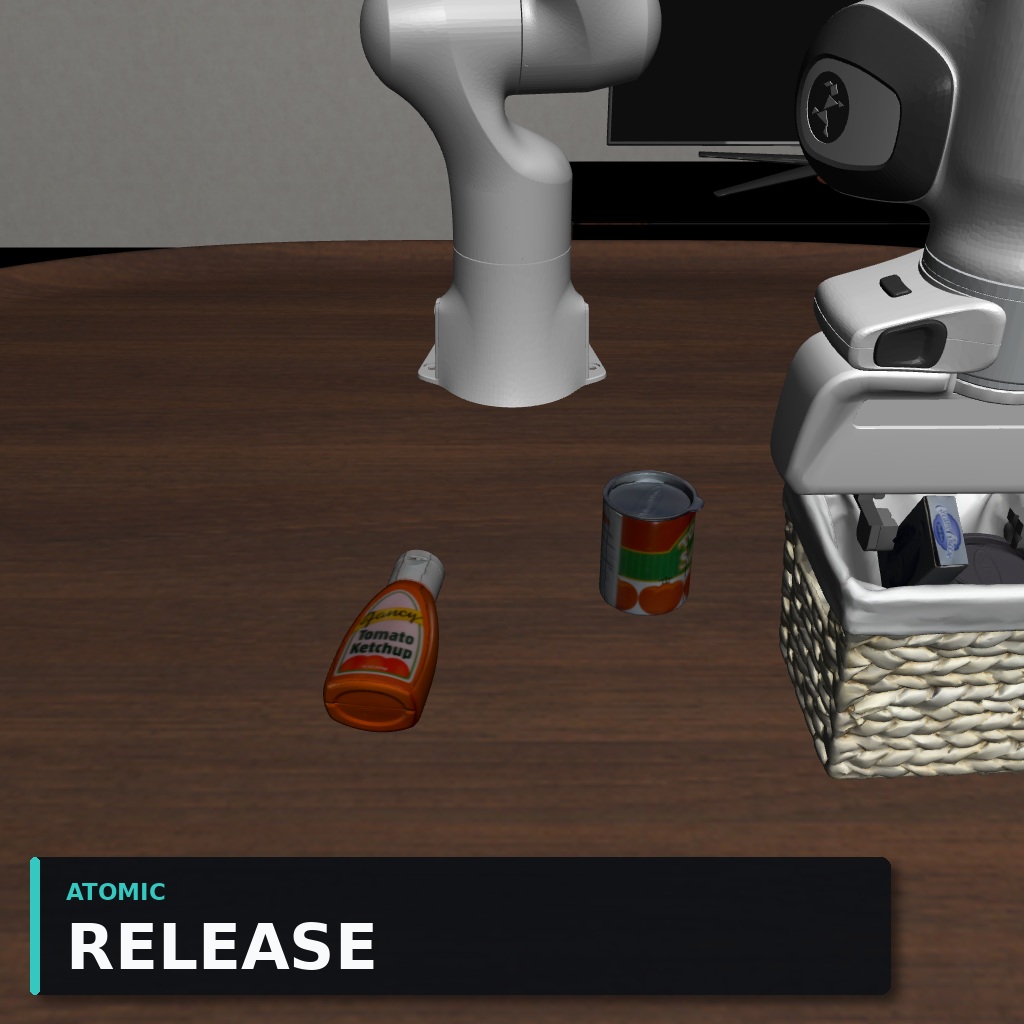}

    \vspace{0.15cm}

    \includegraphics[width=0.155\linewidth]{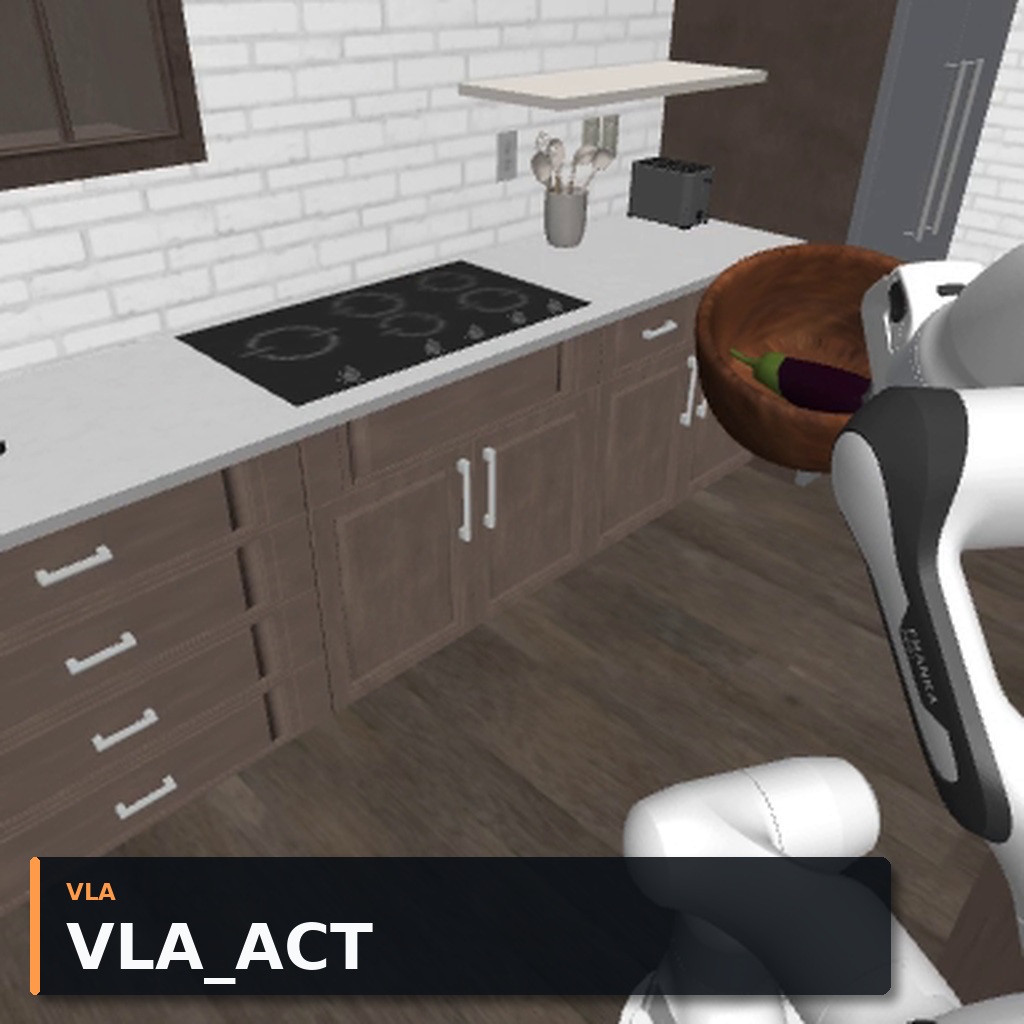}
    \hfill
    \includegraphics[width=0.155\linewidth]{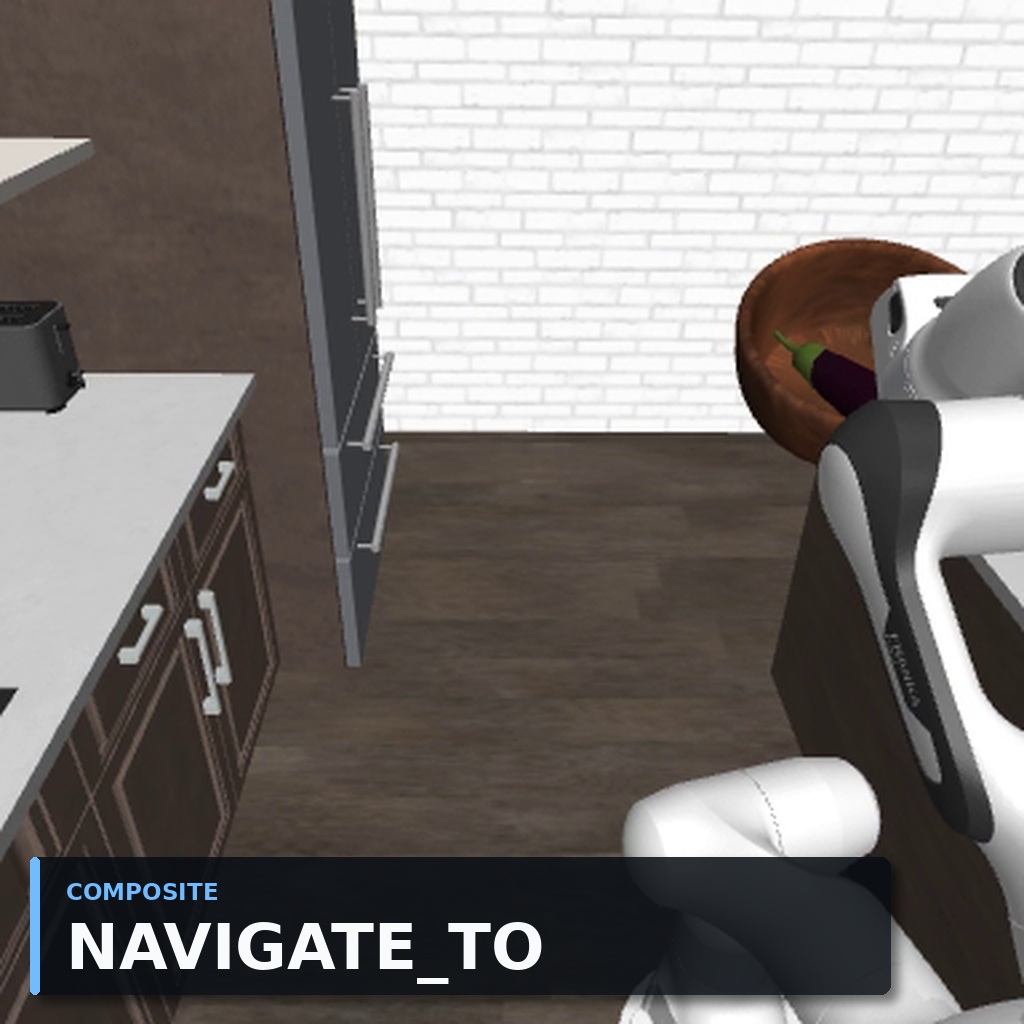}
    \hfill
    \includegraphics[width=0.155\linewidth]{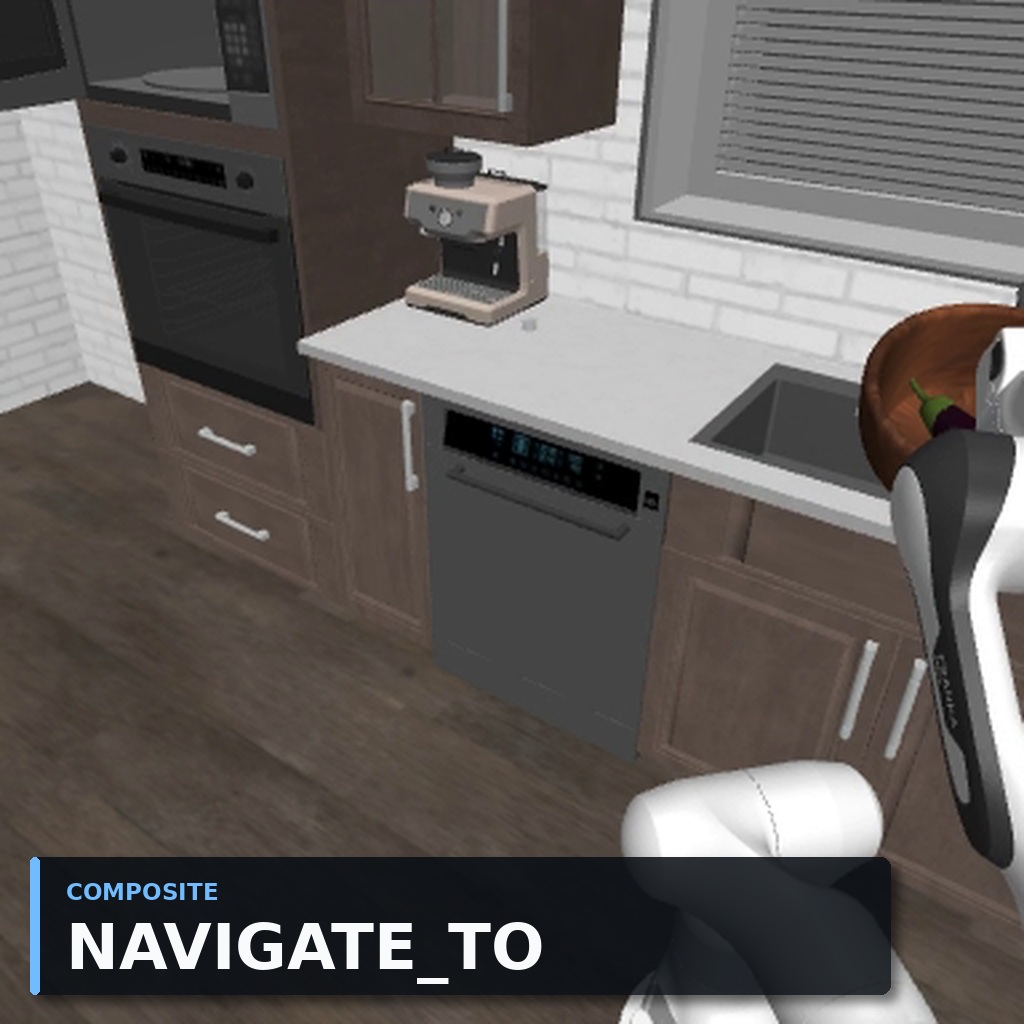}
    \hfill
    \includegraphics[width=0.155\linewidth]{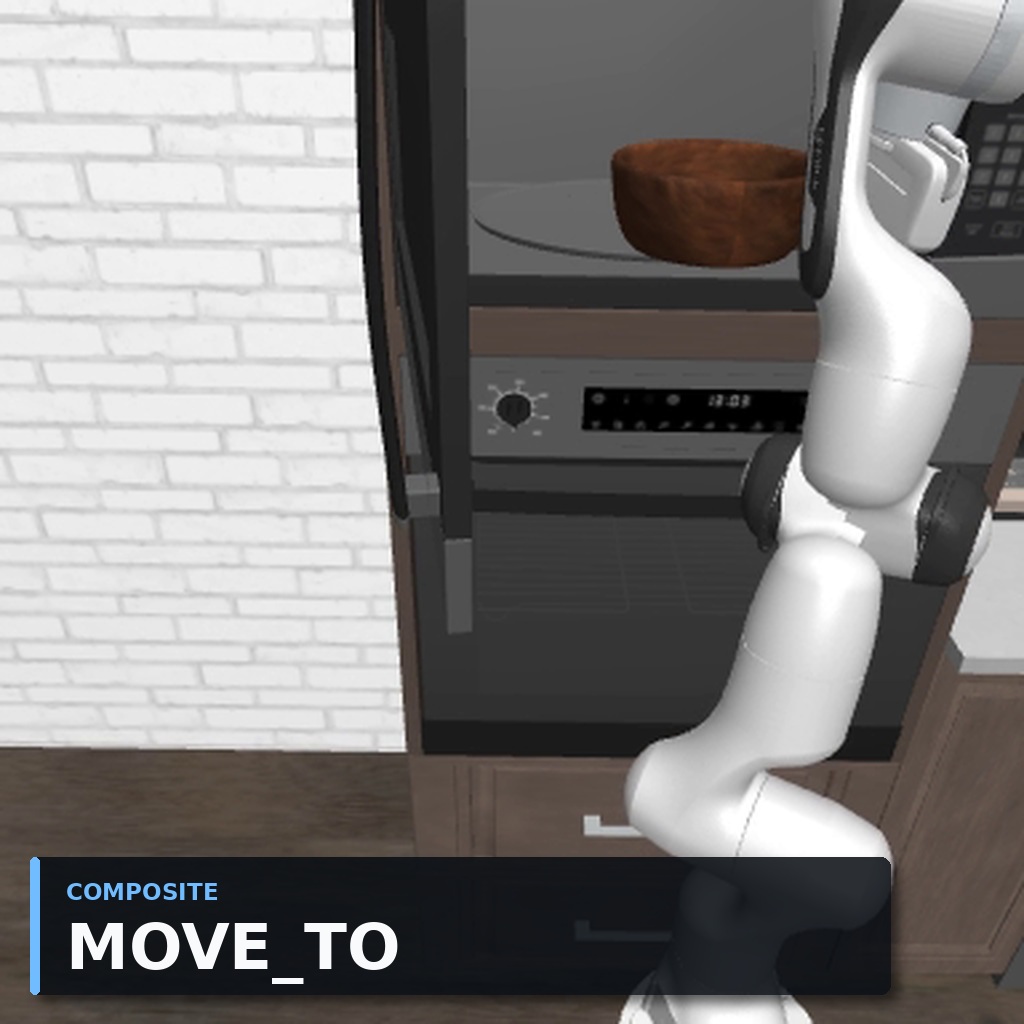}
    \hfill
    \includegraphics[width=0.155\linewidth]{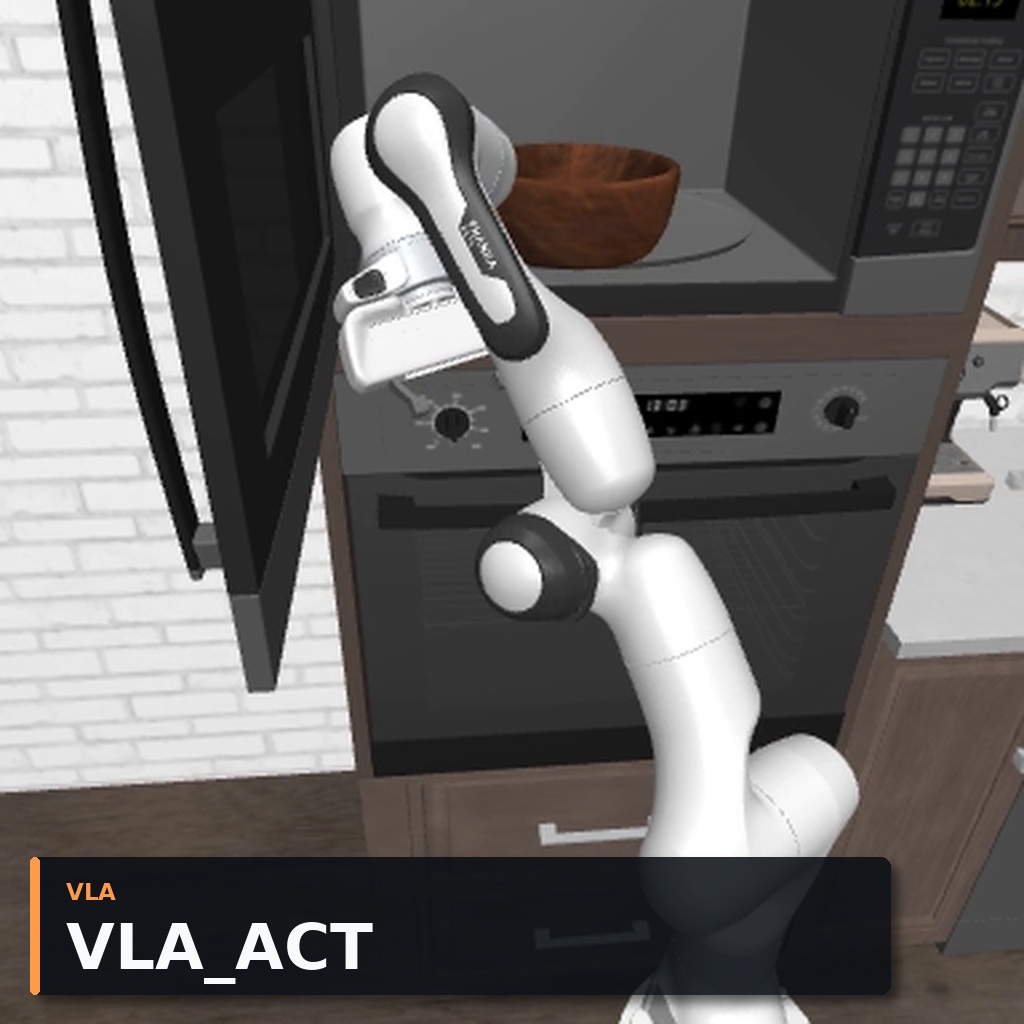}
    \hfill
    \includegraphics[width=0.155\linewidth]{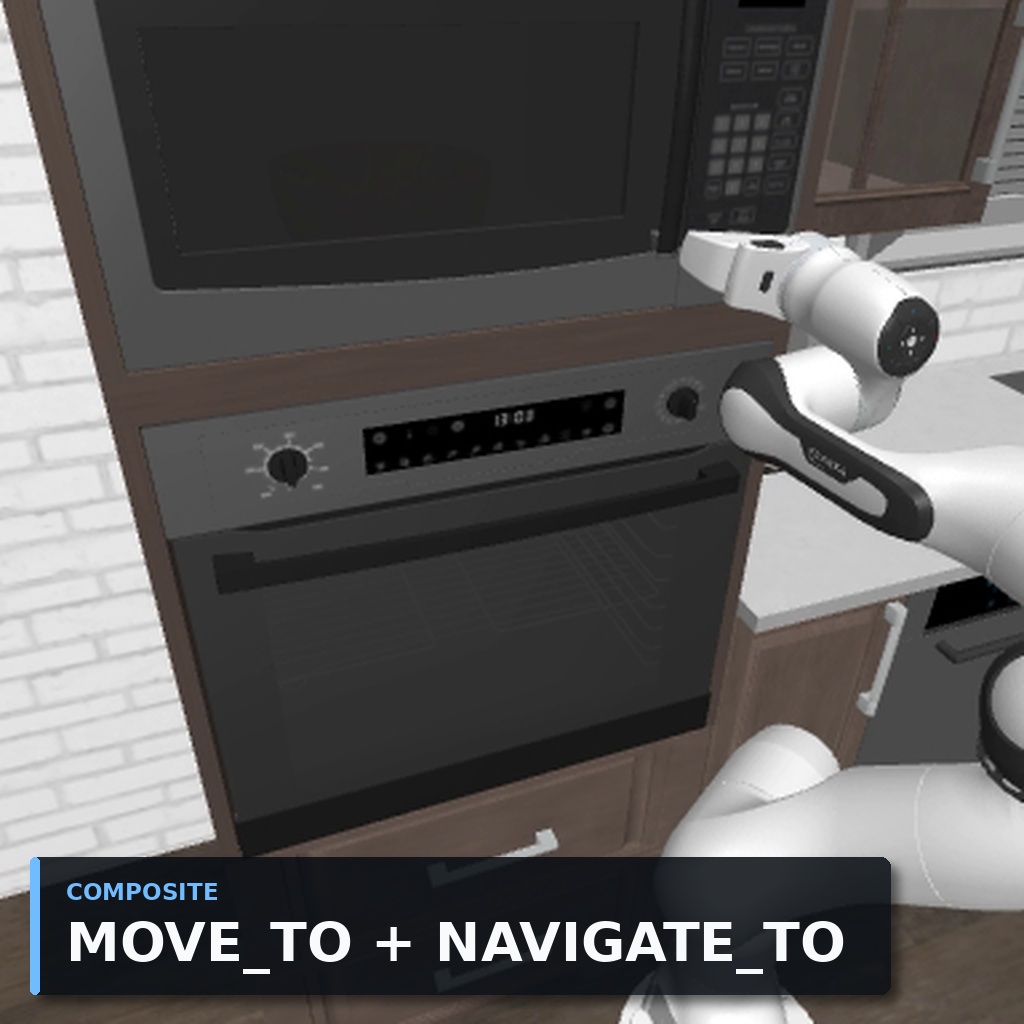}

    \caption{\textbf{Representative rollout frames for analytic decomposition around contact-rich phases.} Top row: on a \textsc{LIBERO-10-Pro} swap task, the agent first invokes \textsc{vla\_act} and starts moving toward the basket, then detects during \textsc{move\_to} that the VLA has not actually grasped the cream-cheese box. The planner moves back, retries \textsc{vla\_act}, and, after a successful grasp, completes the subtask with \textsc{move\_to} and \textsc{release}. Bottom row: on the RoboCasa \textsc{SteamInMicrowave} composite-seen task, the agent successfully grasps the bowl with \textsc{vla\_act}, searches and repositions until the microwave is localized, invokes \textsc{vla\_act} to place the bowl inside, pushes it in with \textsc{move\_to}, closes the door, and finally uses \textsc{move\_to} and \textsc{navigate\_to} to press the switch.}
    \label{fig:analytic_recovery_rollouts}
\end{figure}

Figure~\ref{fig:finisher_attribution} provides the aggregate attribution for this division of labor by separating successful rollouts according to the primitive class that triggers the final benchmark completion predicate. LIBERO Pro-family tasks are usually completed after analytic transport, release, or repositioning once contact has been established. In RoboCasa365 and RoboTwin C2R, the final predicate often depends directly on a contact-rich operation, so successful rollouts more frequently finish inside the VLA primitive. Figure~\ref{fig:analytic_recovery_rollouts} gives representative examples: analytic primitives localize execution, expose failed or incomplete contacts, and move the robot back into a configuration where \textsc{vla\_act} can be invoked again. The combined evidence supports the same division of labor: analytic primitives organize the task around contact-rich phases, while the VLA remains responsible for the phases where learned visuomotor control is needed.

\section{Flash Mode and Real-World Deployment}
\label{sec:flash_real_world}

Harness VLA extends the memory-guided primitive interface to two deployment settings: Flash Mode executes prepared procedures without online agent planning, while physical deployment uses the same separation between frozen contact skills and task-level coordination.

\subsection{Flash Mode}
\label{sec:flash_cards}
\label{sec:flash_execution}

Flash Mode reuses a successful primitive sequence through a task card that records the action order, VLA prompts, and non-spatial arguments. Spatial waypoints are stored relative to task-relevant objects or destinations. At deployment, a VLM identifies these targets in RGB images; the system uses the corresponding depth and camera calibration to back-project their image locations into 3D coordinates. It then updates the waypoints and invokes the recorded primitives in order, without online agent planning. Grasp checks and a limited number of retries are predefined in the procedure.

For each task, we compile all available successful agent runs into candidate cards by extracting their primitive sequences and identifying the objects and destinations used by each action. We replay the candidates and retain the card with the highest observed success rate, denoted Flash Mode (max), using one card across all seeds of the task. Flash Mode currently targets simple tasks with a fixed subgoal order and directly localizable objects. It can replay a longer fixed routine, but new subgoals, reordered actions, or unforeseen recovery require online planning. Appendix~\ref{app:flash_mode} details card generation and selection on the same seeds used for reporting.
\begin{figure}[!ht]
    \centering
    \makebox[\linewidth][l]{\small\textbf{VLA fullshot}}\par
    \vspace{0.05cm}
    \includegraphics[width=0.32\linewidth]{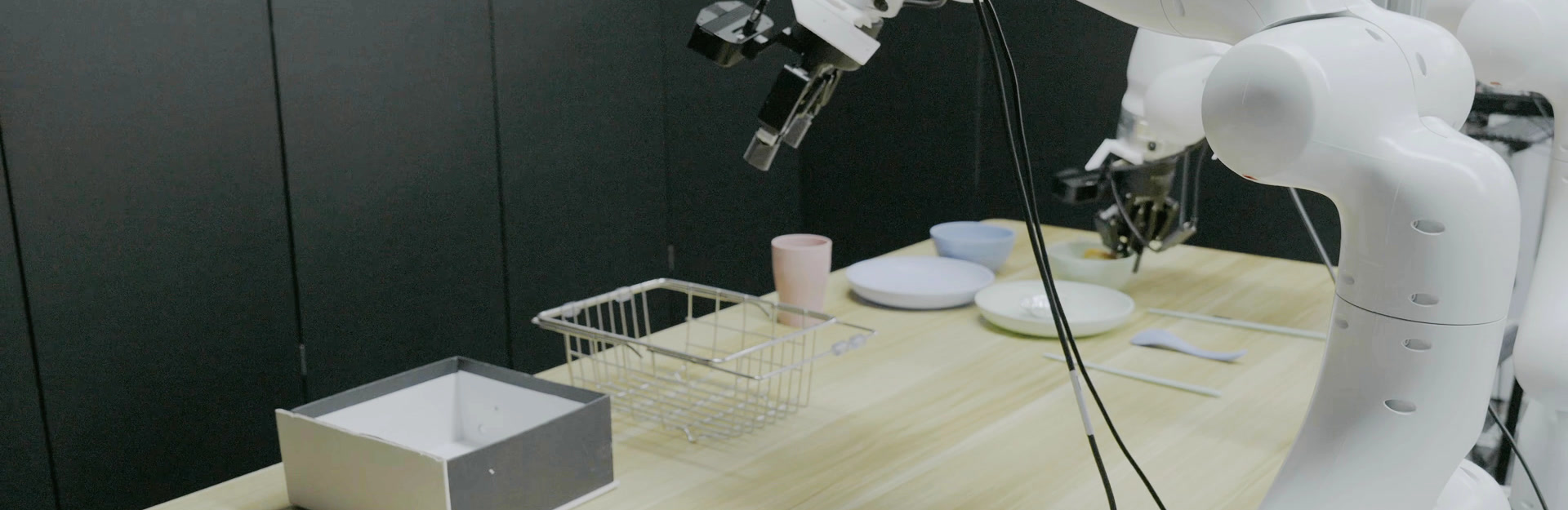}
    \hfill
    \includegraphics[width=0.32\linewidth]{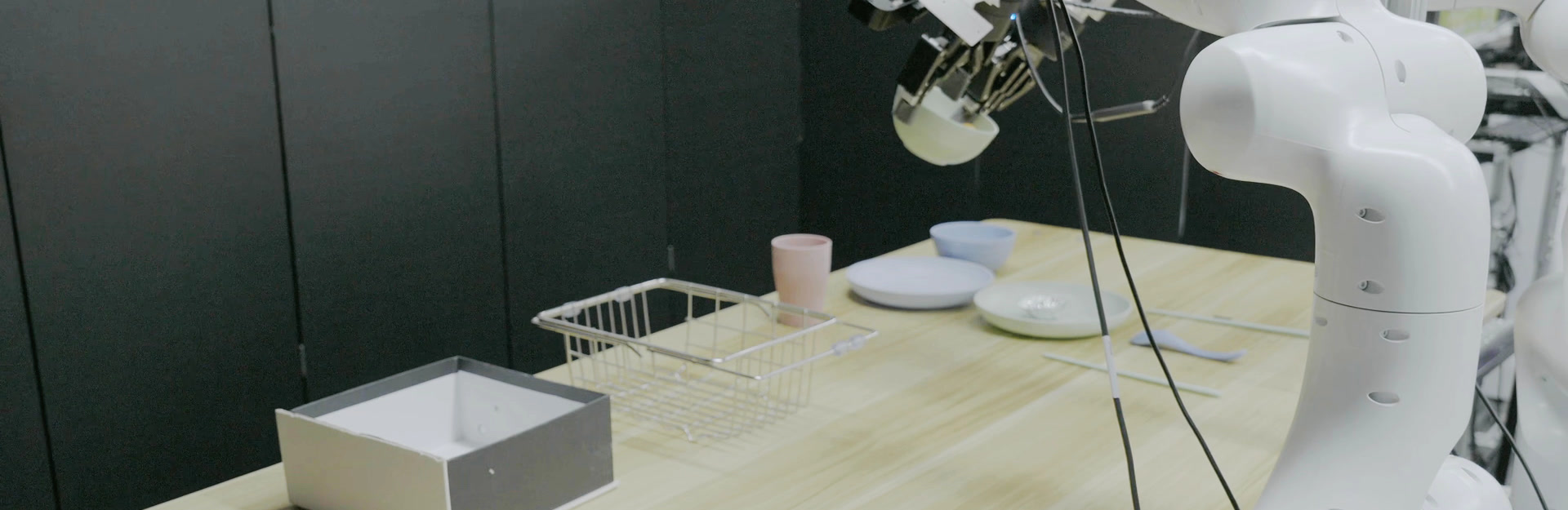}
    \hfill
    \includegraphics[width=0.32\linewidth]{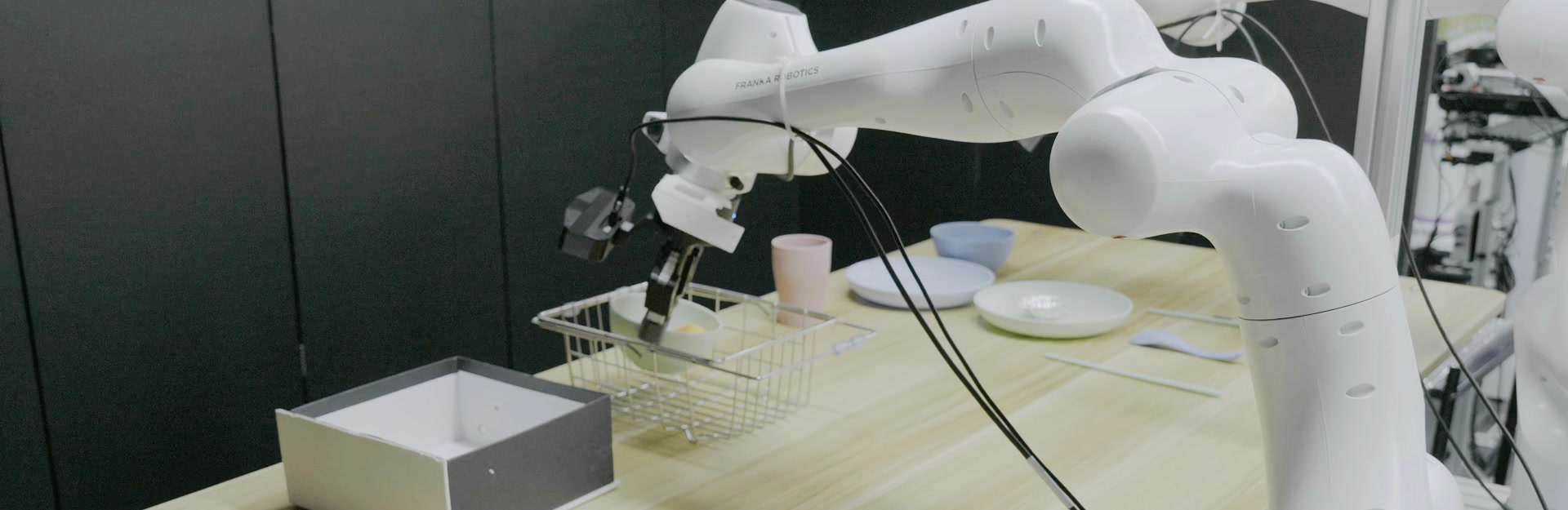}

    \vspace{0.15cm}
    \makebox[\linewidth][l]{\small\textbf{Harness VLA (online planning)}}\par
    \vspace{0.05cm}
    \includegraphics[width=0.32\linewidth]{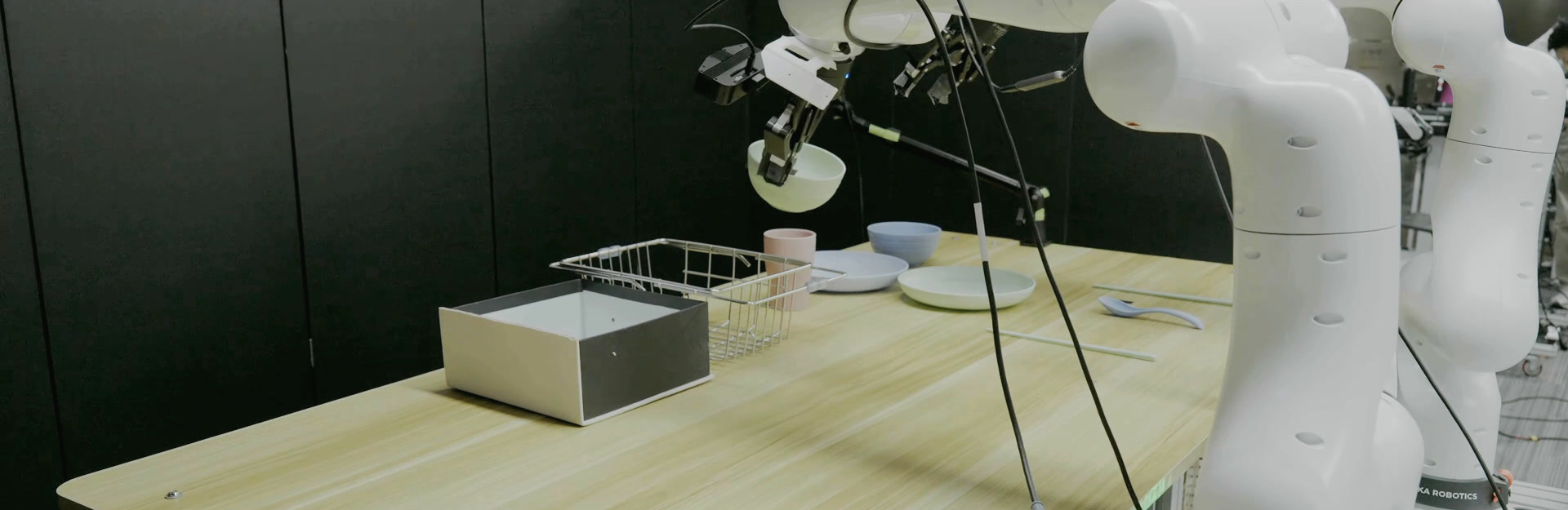}
    \hfill
    \includegraphics[width=0.32\linewidth]{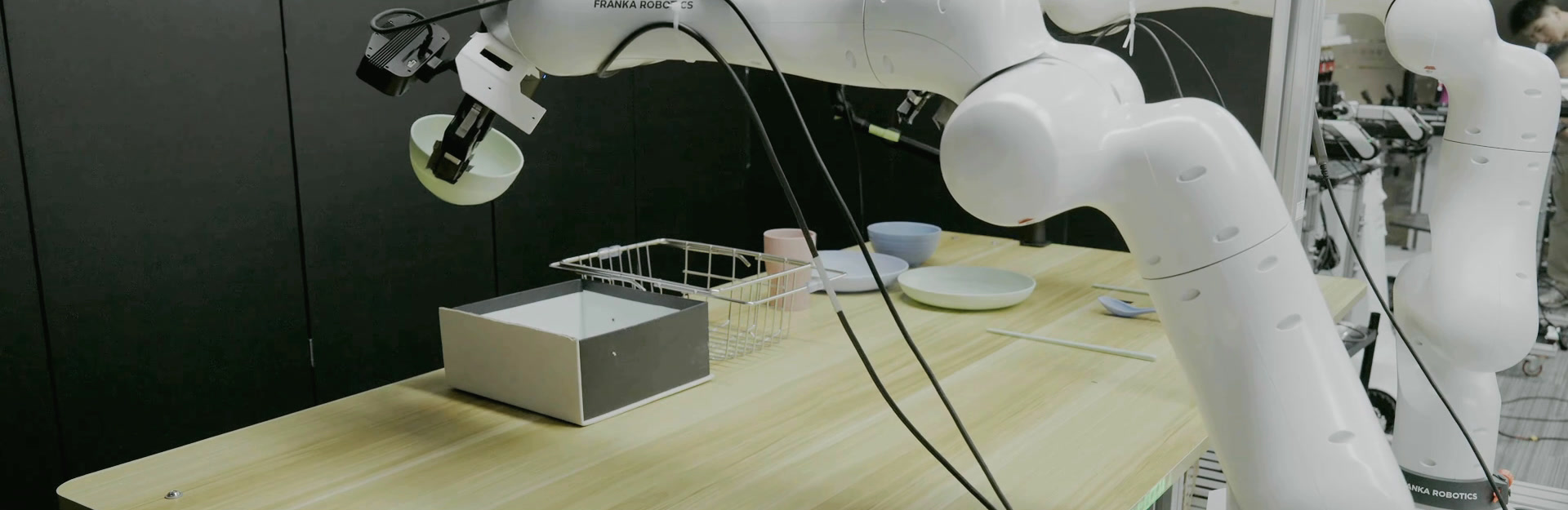}
    \hfill
    \includegraphics[width=0.32\linewidth]{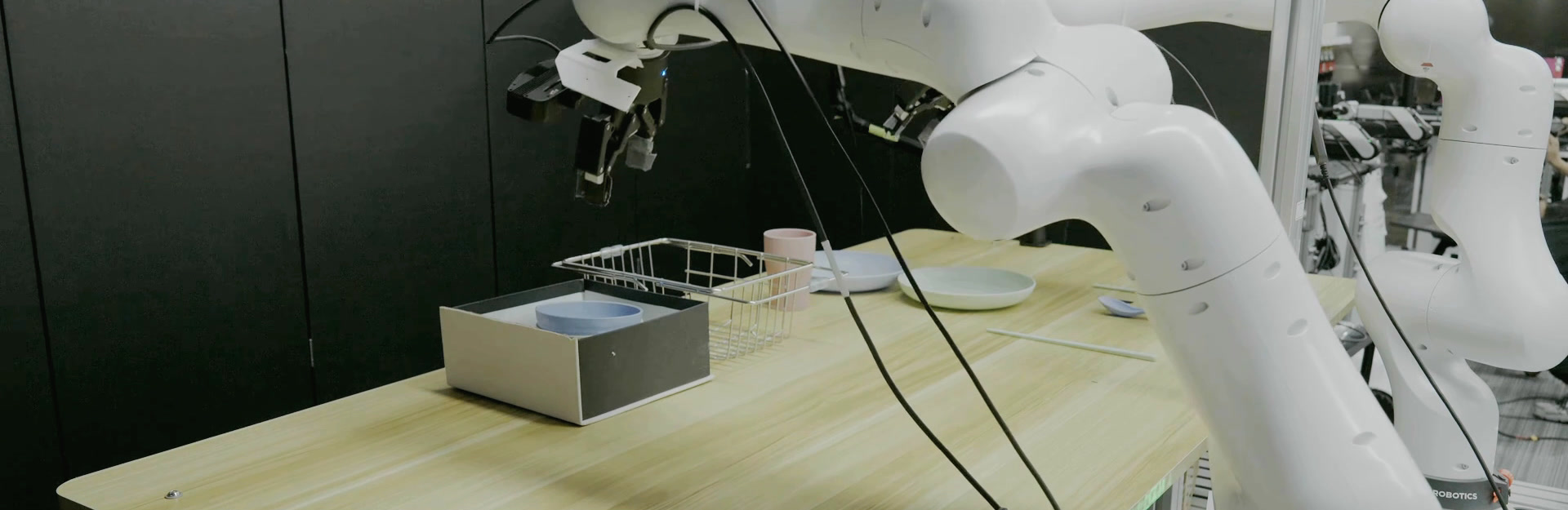}

    \vspace{0.15cm}

    \includegraphics[width=0.32\linewidth]{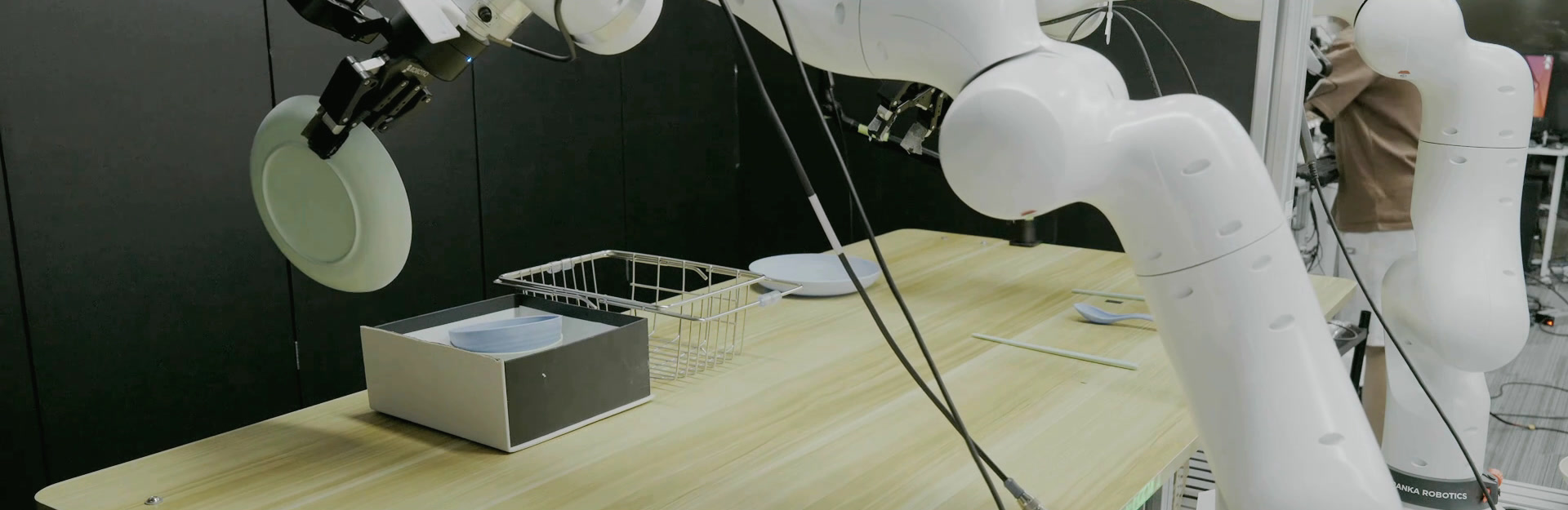}
    \hfill
    \includegraphics[width=0.32\linewidth]{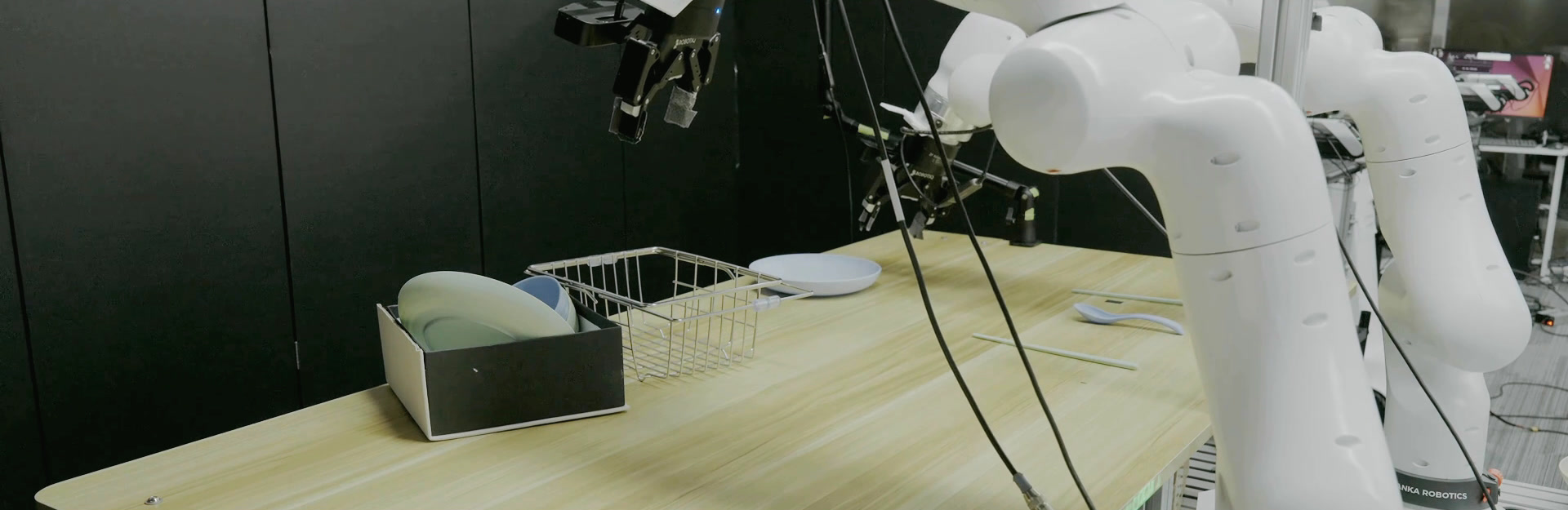}
    \hfill
    \includegraphics[width=0.32\linewidth]{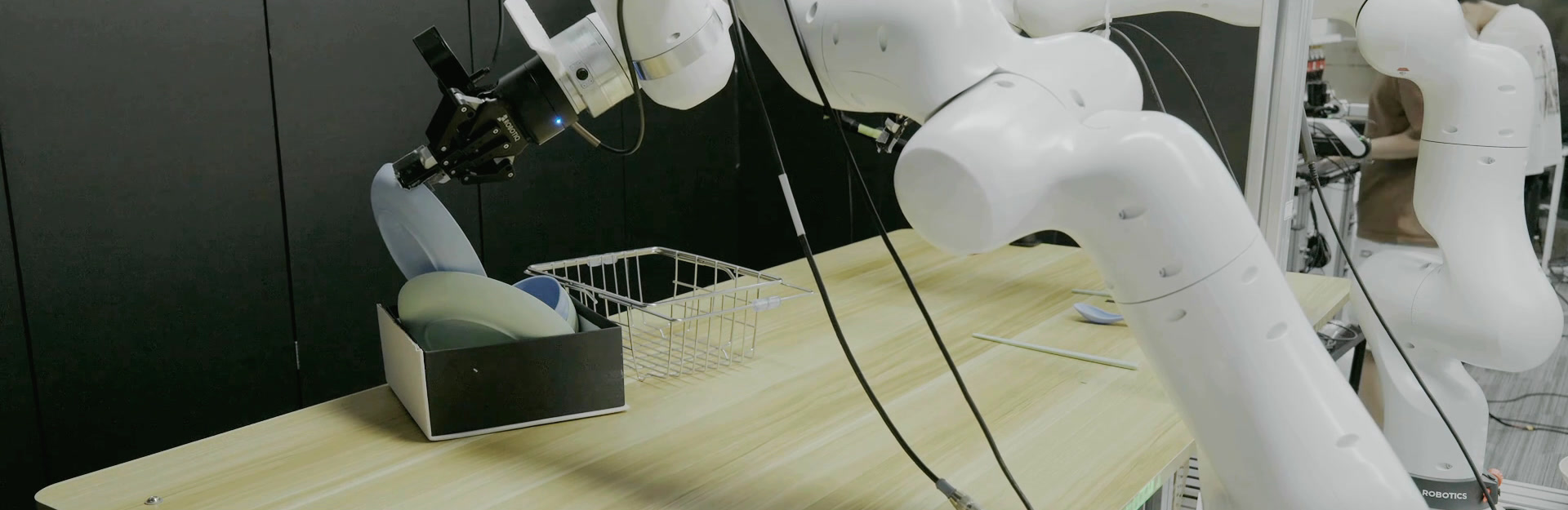}

    \vspace{0.15cm}

    \includegraphics[width=0.32\linewidth]{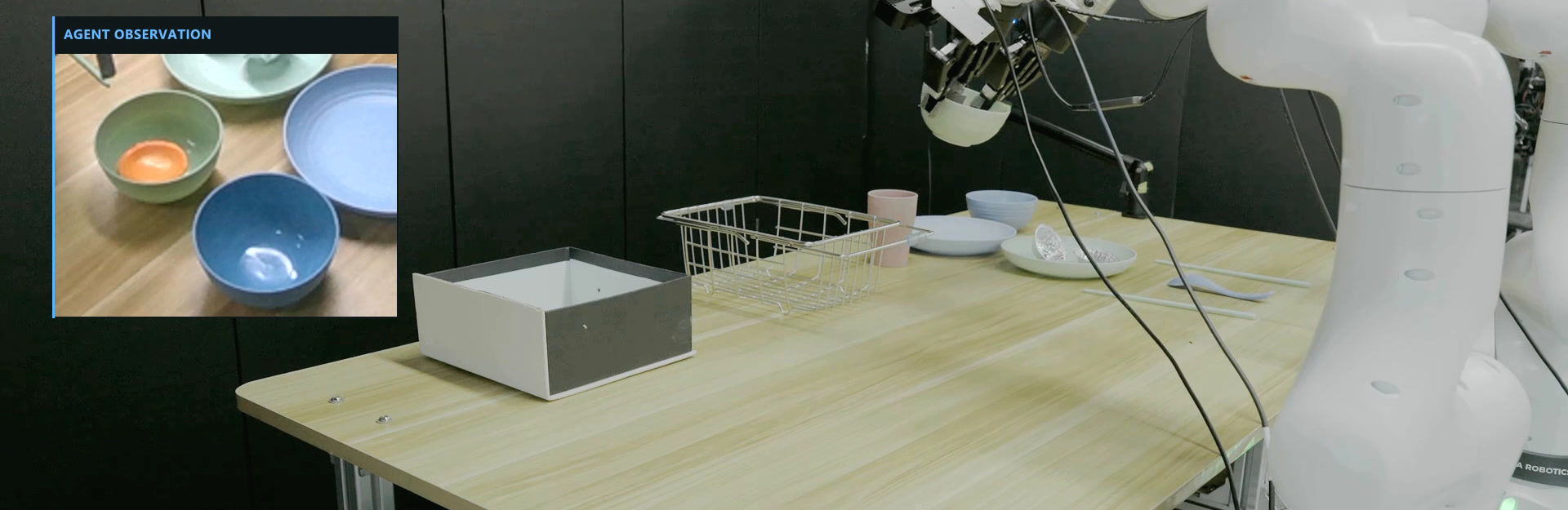}
    \hfill
    \includegraphics[width=0.32\linewidth]{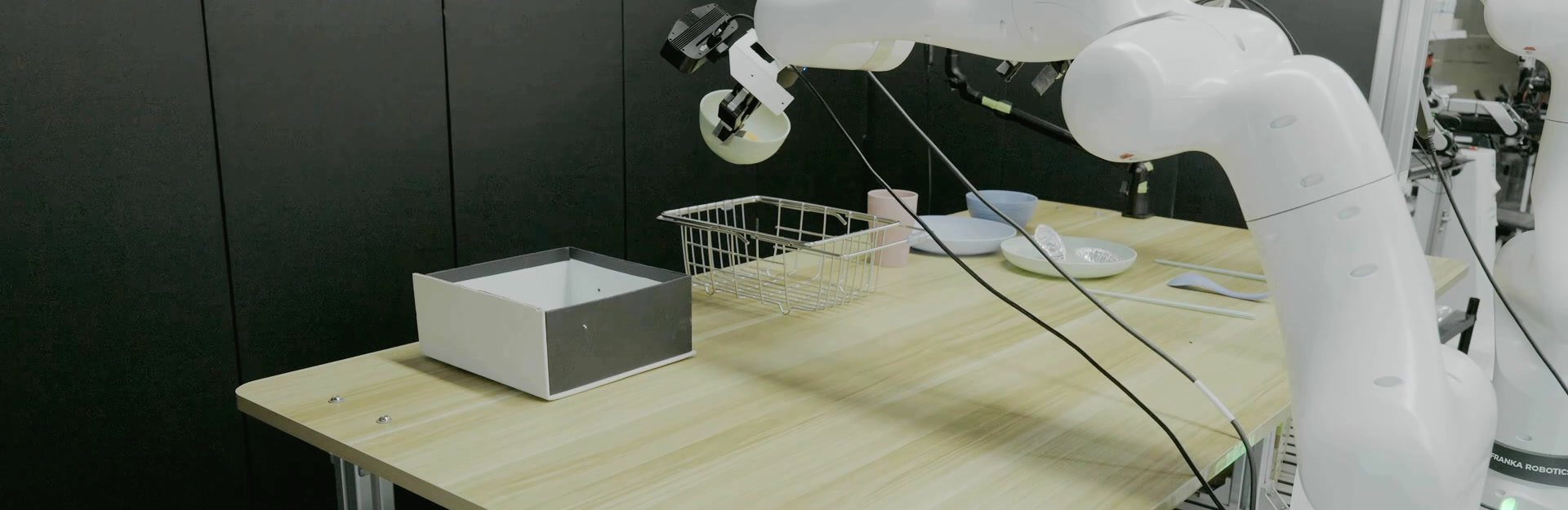}
    \hfill
    \includegraphics[width=0.32\linewidth]{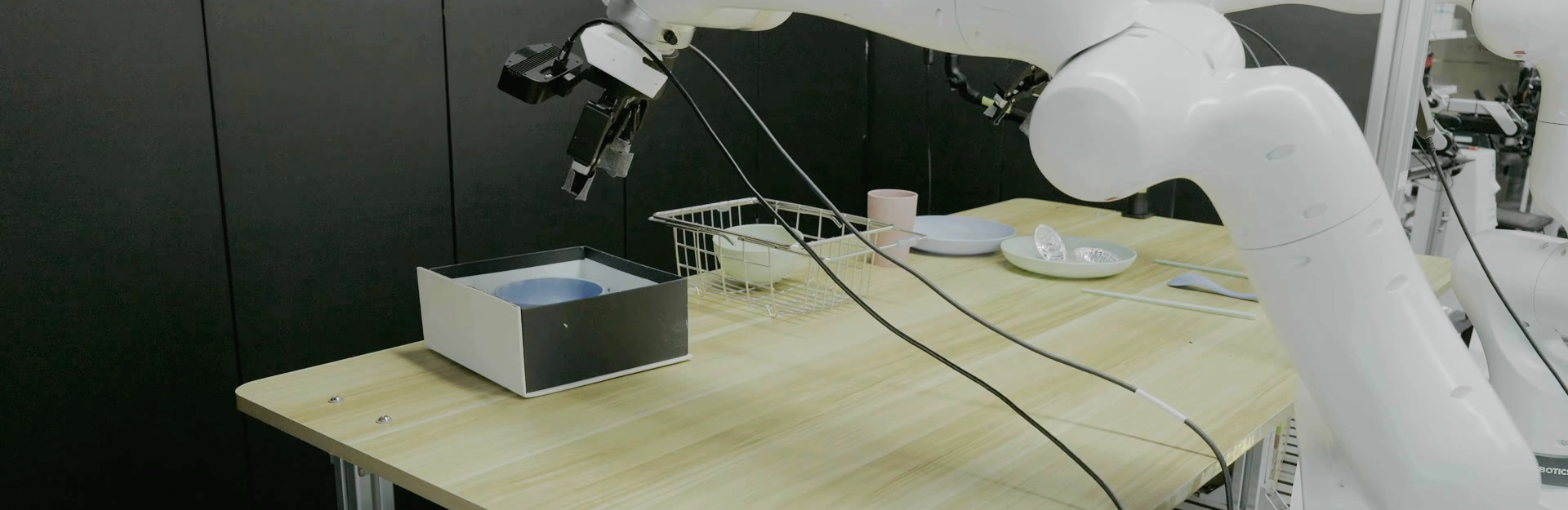}

    \vspace{0.15cm}

    \includegraphics[width=0.32\linewidth]{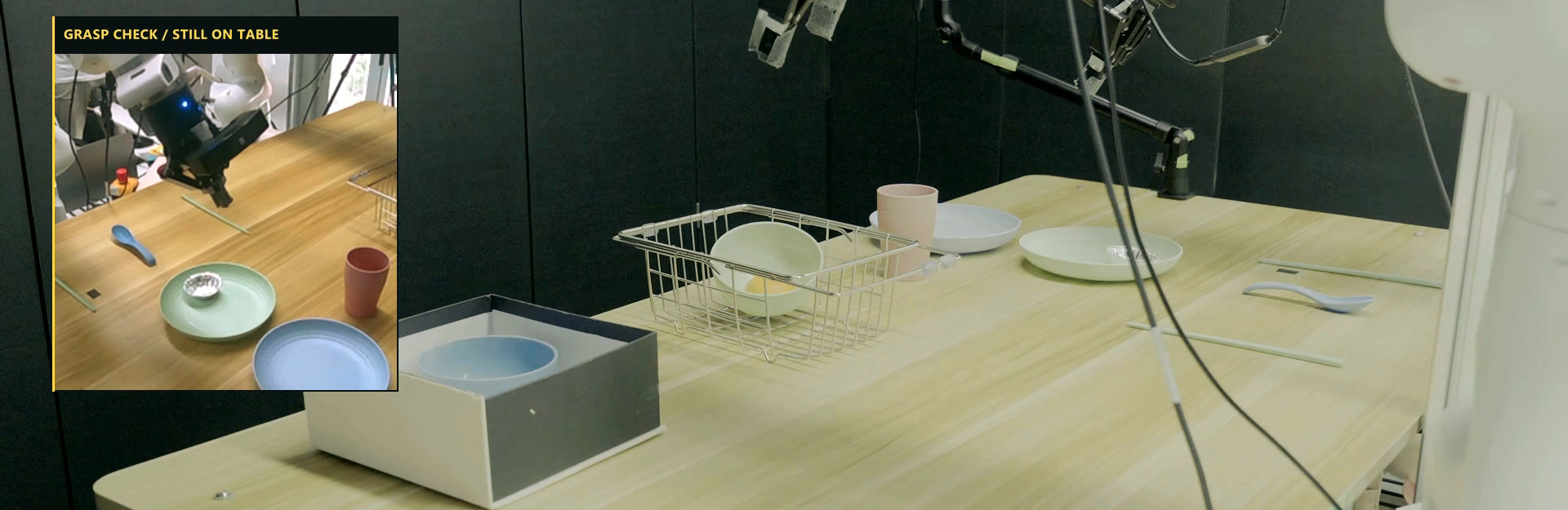}
    \hfill
    \includegraphics[width=0.32\linewidth]{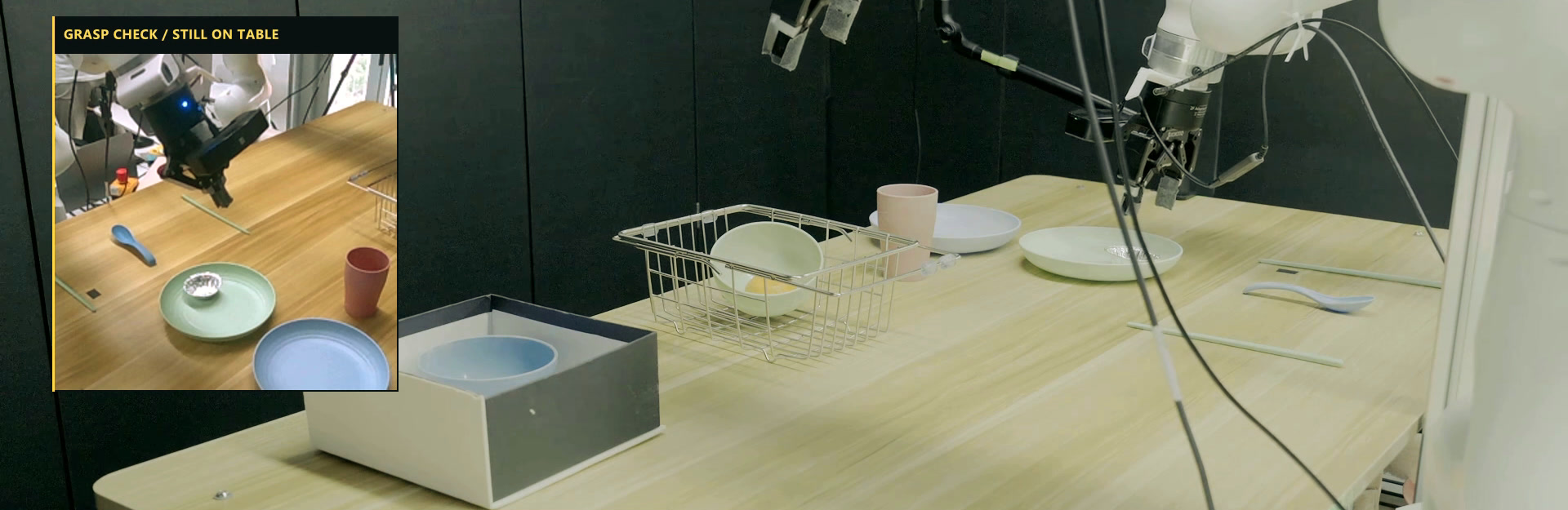}
    \hfill
    \includegraphics[width=0.32\linewidth]{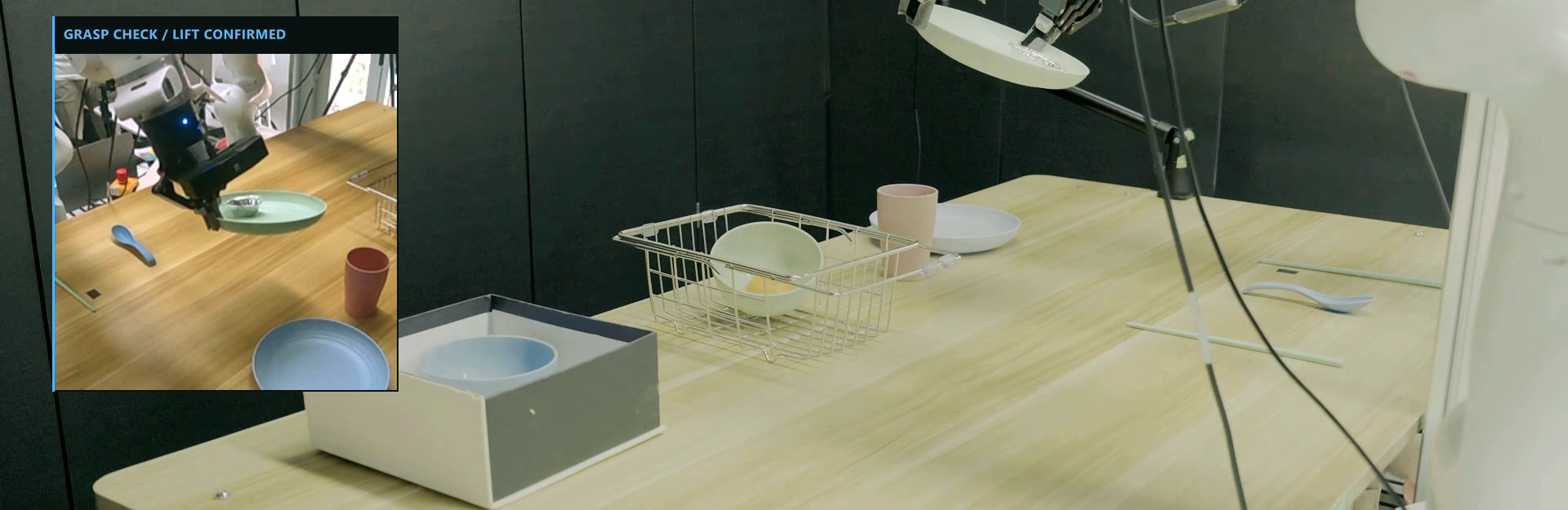}

    \caption{\textbf{From the learned routine to online adaptation.} Each row progresses from left to right. Top row (VLA fullshot): grasp a bowl, hand it off, and place it in the basket. The four Harness VLA rows show: stage a bowl above the box, invoke placement, and place the second bowl; stage the green plate, continue with the blue plate, and place it after re-grounding; inspect the marked bowl, stage it above the basket, and route the unmarked bowl to the box; reopen after a failed grasp, retry, and confirm the lift before handoff.}
    \label{fig:real_world_cases}
\end{figure}
\paragraph{Success and execution time.}
\label{sec:flash_results}
Table~\ref{tab:flash_libero_pro} compares Flash Mode with Codex under high reasoning and no reasoning on LIBERO-Pro. Flash reaches $72.63\%$ success ($581/800$), between the two Codex settings. Its timing statistic measures tool execution in the selected source trajectories, while Codex timing measures planner duration; these values do not establish an end-to-end speedup.

\begin{table}[htbp]
\centering
\caption{Success rates (\%) and execution time on LIBERO-Pro from the RPent Flash comparison. \textbf{T}/\textbf{S} denote instruction-redirection/position-swap perturbations; each suite contains $100$ episodes. $\dagger$ Flash time averages source-trajectory tool durations; Codex time averages planner durations. Appendix~\ref{app:flash_mode} gives the timing protocol and same-seed max-card selection details. Missing cards count as failures.}
\label{tab:flash_libero_pro}
\small
\setlength{\tabcolsep}{2.5pt}
\begin{tabular}{l cccccccc c r}
\toprule
Method & Spat-T & Spat-S & Obj-T & Obj-S & Goal-T & Goal-S & L10-T & L10-S & Overall & Time (s) \\
\midrule
Codex (reasoning high) & 88.0 & 88.0 & 95.0 & 98.0 & 77.0 & 56.0 & 68.0 & 58.0 & 78.50 & 529.23 \\
Codex (no reasoning) & 86.0 & 67.0 & 94.0 & 92.0 & 15.0 & 64.0 & 43.0 & 39.0 & 62.50 & 344.17 \\
Flash Mode (max) & 79.0 & 70.0 & 86.0 & 93.0 & 74.0 & 65.0 & 60.0 & 54.0 & 72.63 & $56.17^{\dagger}$ \\
\bottomrule
\end{tabular}
\end{table}

\paragraph{Real-world deployment.}
\label{sec:dual_franka_deployment}

We demonstrate online and Flash execution on two dual-Franka installations. In the online setup, base and wrist cameras provide policy observations, and a calibrated RealSense D455 supplies metric localization. Codex with GPT-5.5 invokes a frozen $\pi_{0.5}$ desk-clearing policy for grasping, handoff, and placement, using phase-specific stopping conditions. Following Section~\ref{sec:method:primitives}, analytic primitives handle pre-contact staging, free-space transport, and reorientation, while \textsc{vla\_act} executes local contact-rich phases. This division reuses the same policy without fine-tuning.

\paragraph{Reset and reward.}
Real-world resets and success judgments require a human in the loop. To reset the scene, an operator restores the initial object arrangement and confirms that a new attempt can begin. Objects are not reset within an evaluation attempt. Task success is recorded only after an operator inspects the current scene and confirms that the task goal has been achieved. The remaining execution and memory workflow follows Section~\ref{sec:method}.

\paragraph{Qualitative comparisons.}
\label{sec:real_world_cases}
\label{sec:physical_flash}

Figure~\ref{fig:real_world_cases} compares online Harness VLA with fullshot execution of the frozen desk-clearing policy. The top row shows the learned grasp--handoff--basket-placement routine. The remaining rows show Harness VLA redirecting bowls to a box, transferring plates in a specified order, assigning bowls to destinations based on a visible dirty marker, and recovering from a missed grasp.

\begin{figure}[!ht]
    \centering
    \makebox[\linewidth][l]{\small\textbf{VLA fullshot}}\par
    \vspace{0.05cm}
    \includegraphics[width=0.32\linewidth]{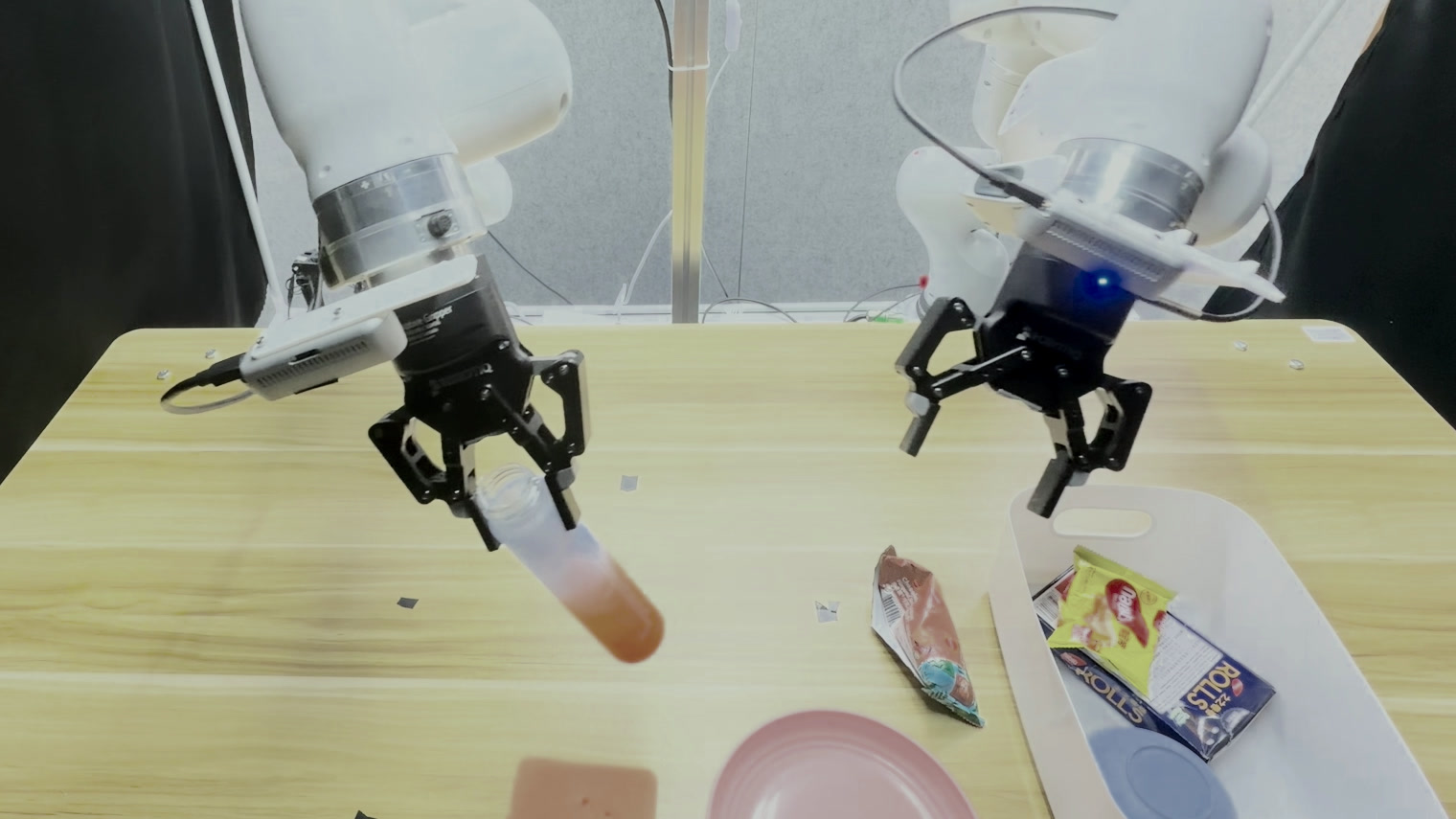}
    \hfill
    \includegraphics[width=0.32\linewidth]{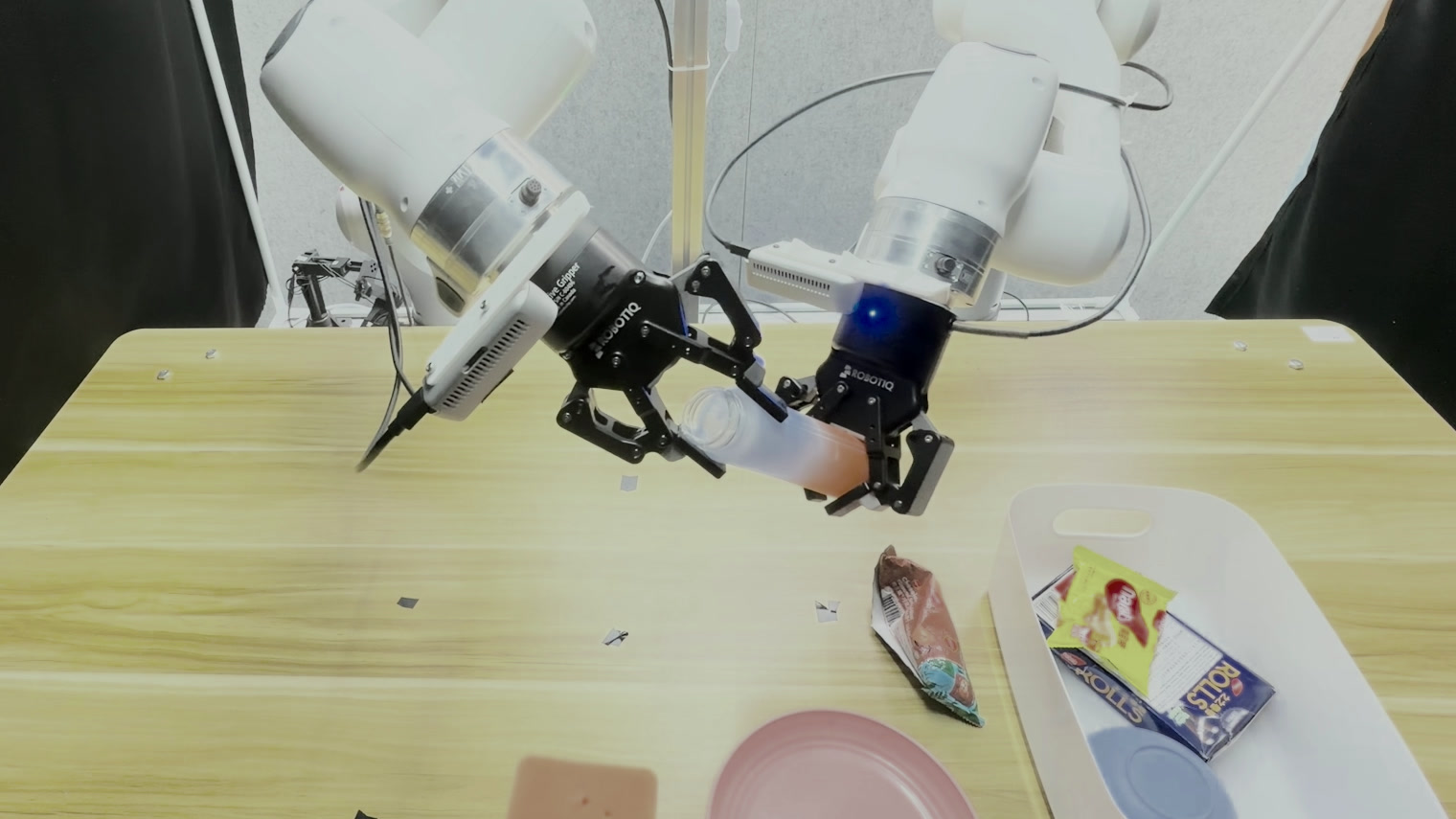}
    \hfill
    \includegraphics[width=0.32\linewidth]{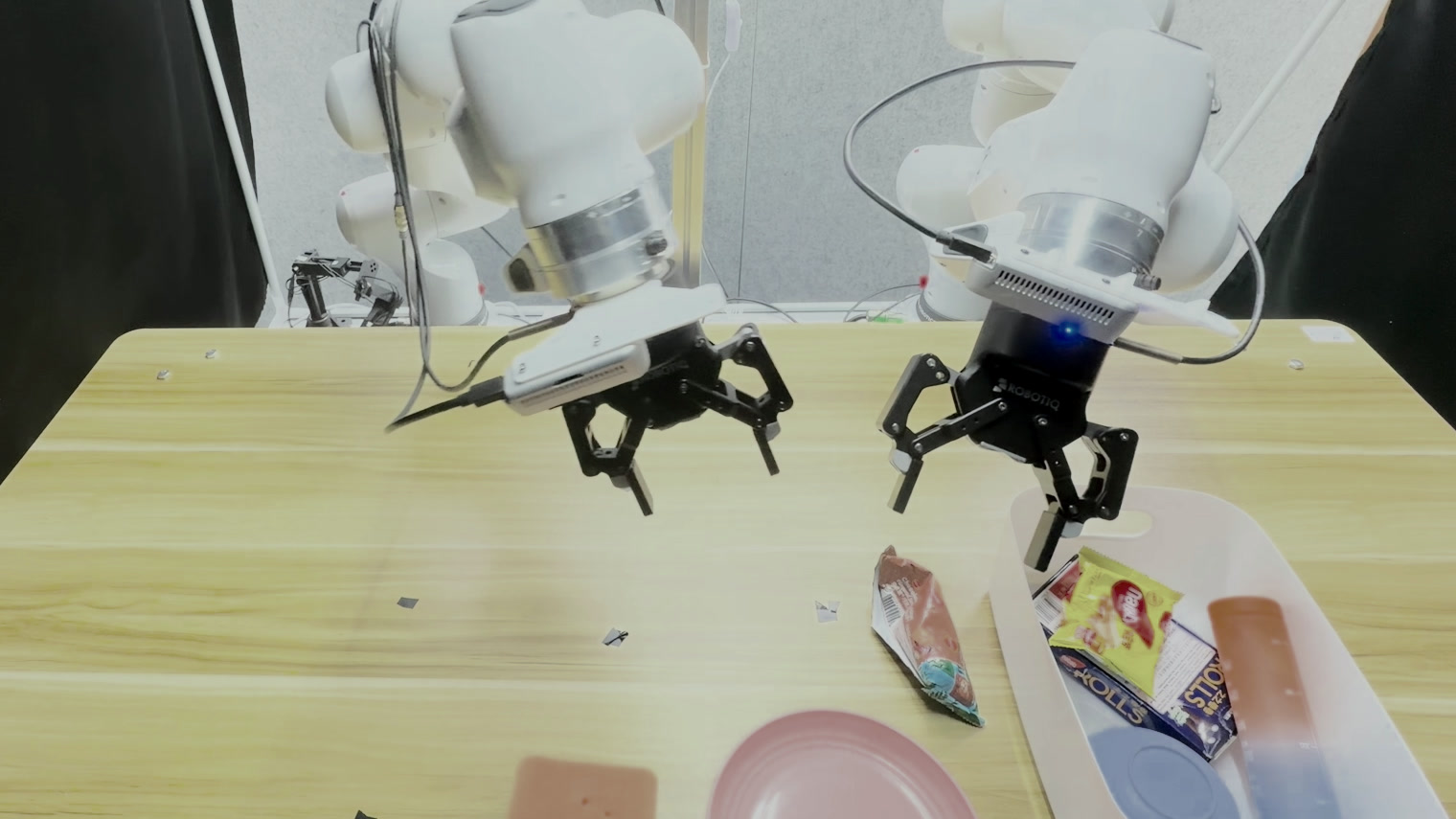}

    \vspace{0.15cm}
    \makebox[\linewidth][l]{\small\textbf{Harness VLA (Flash Mode)}}\par
    \vspace{0.05cm}
    \includegraphics[width=0.32\linewidth]{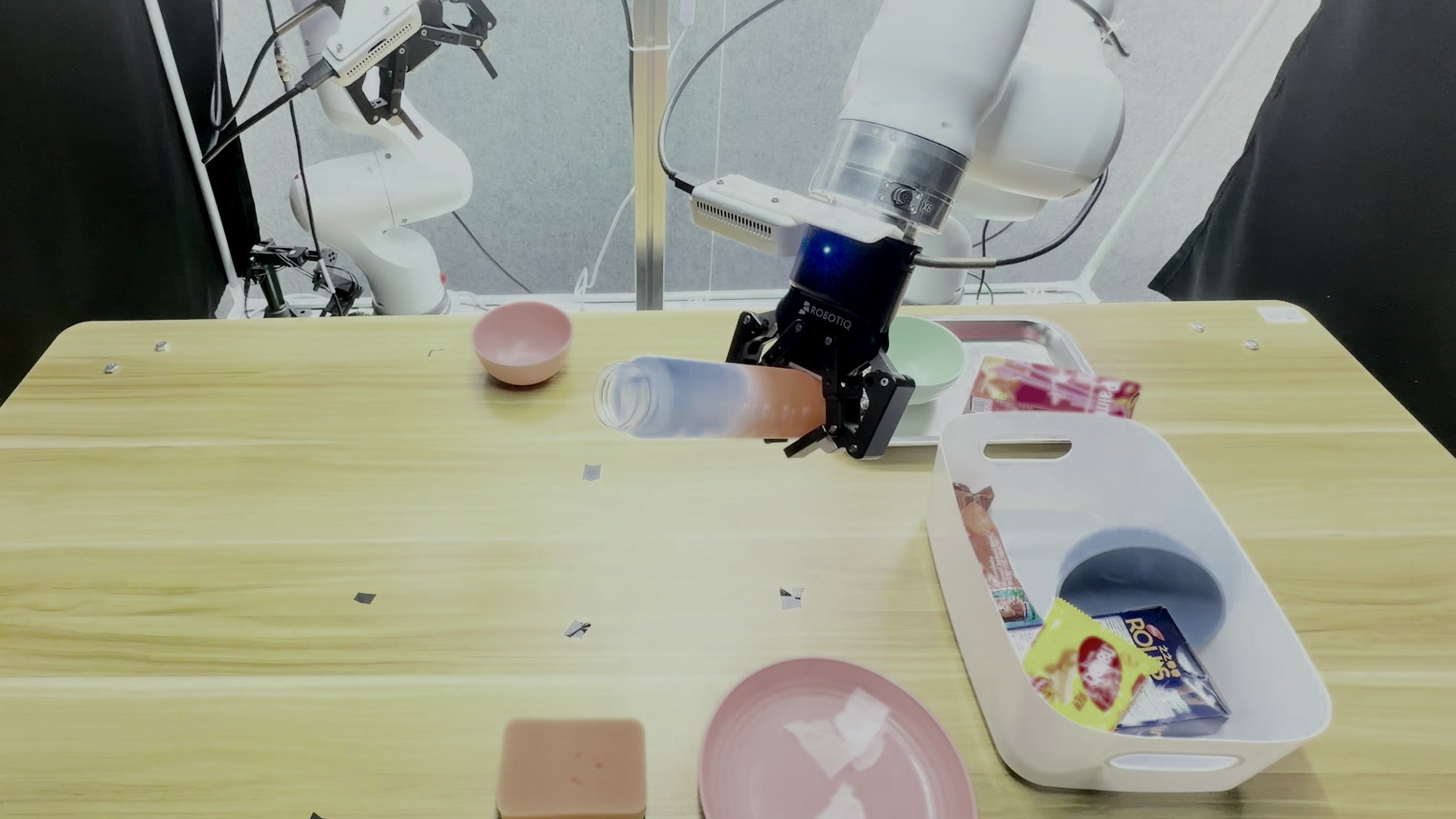}
    \hfill
    \includegraphics[width=0.32\linewidth]{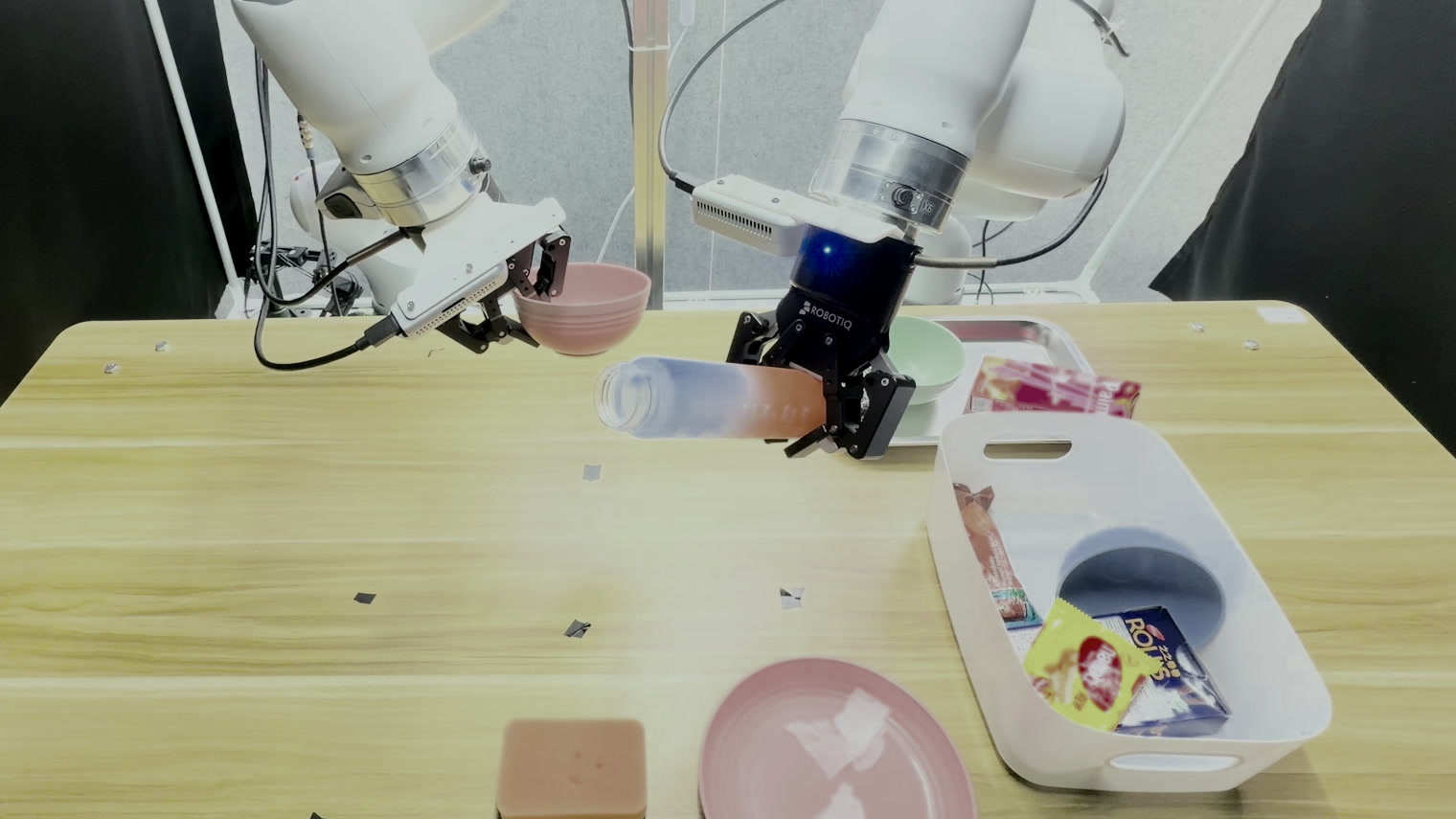}
    \hfill
    \includegraphics[width=0.32\linewidth]{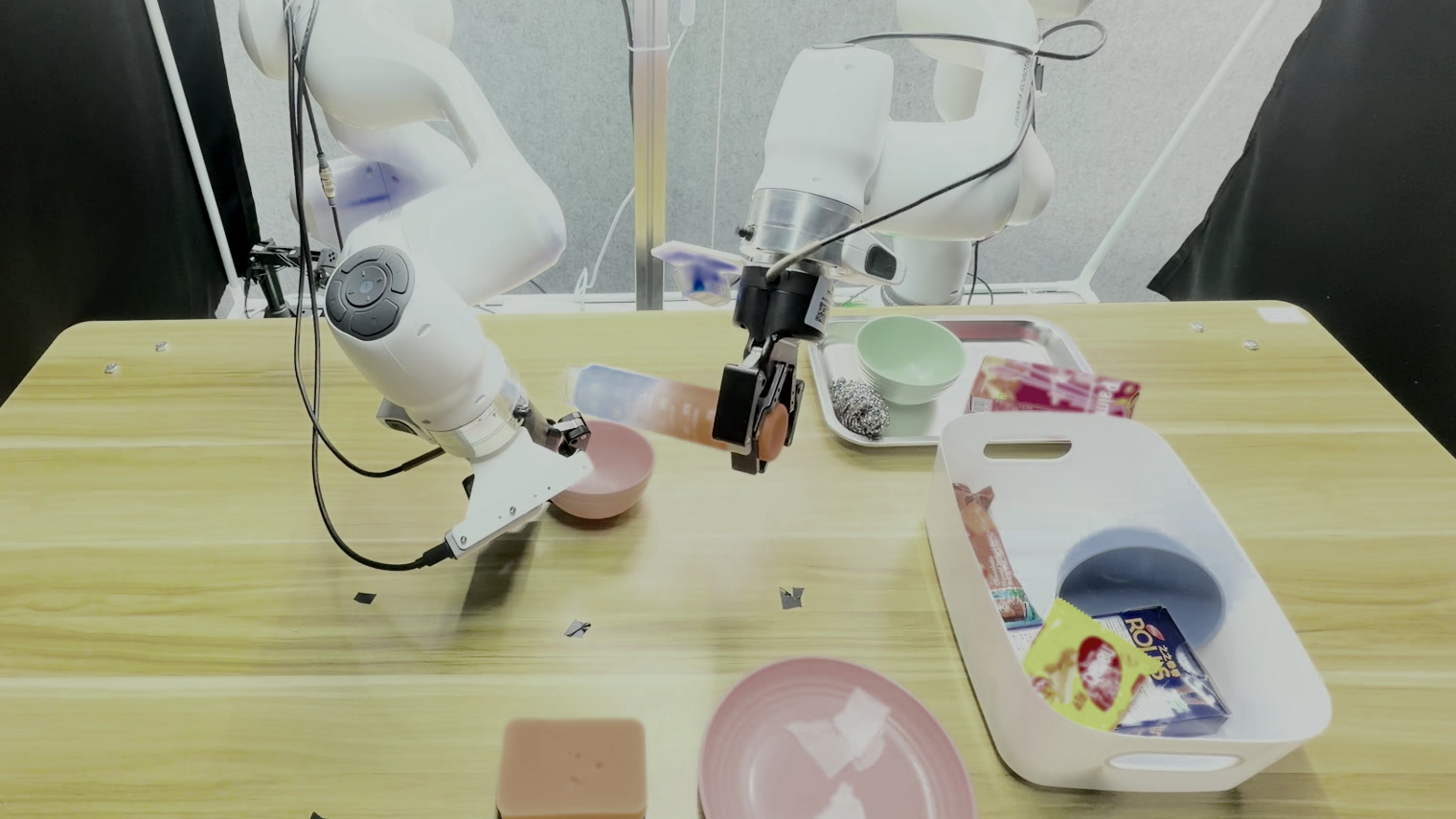}

    \vspace{0.15cm}

    \includegraphics[width=0.32\linewidth]{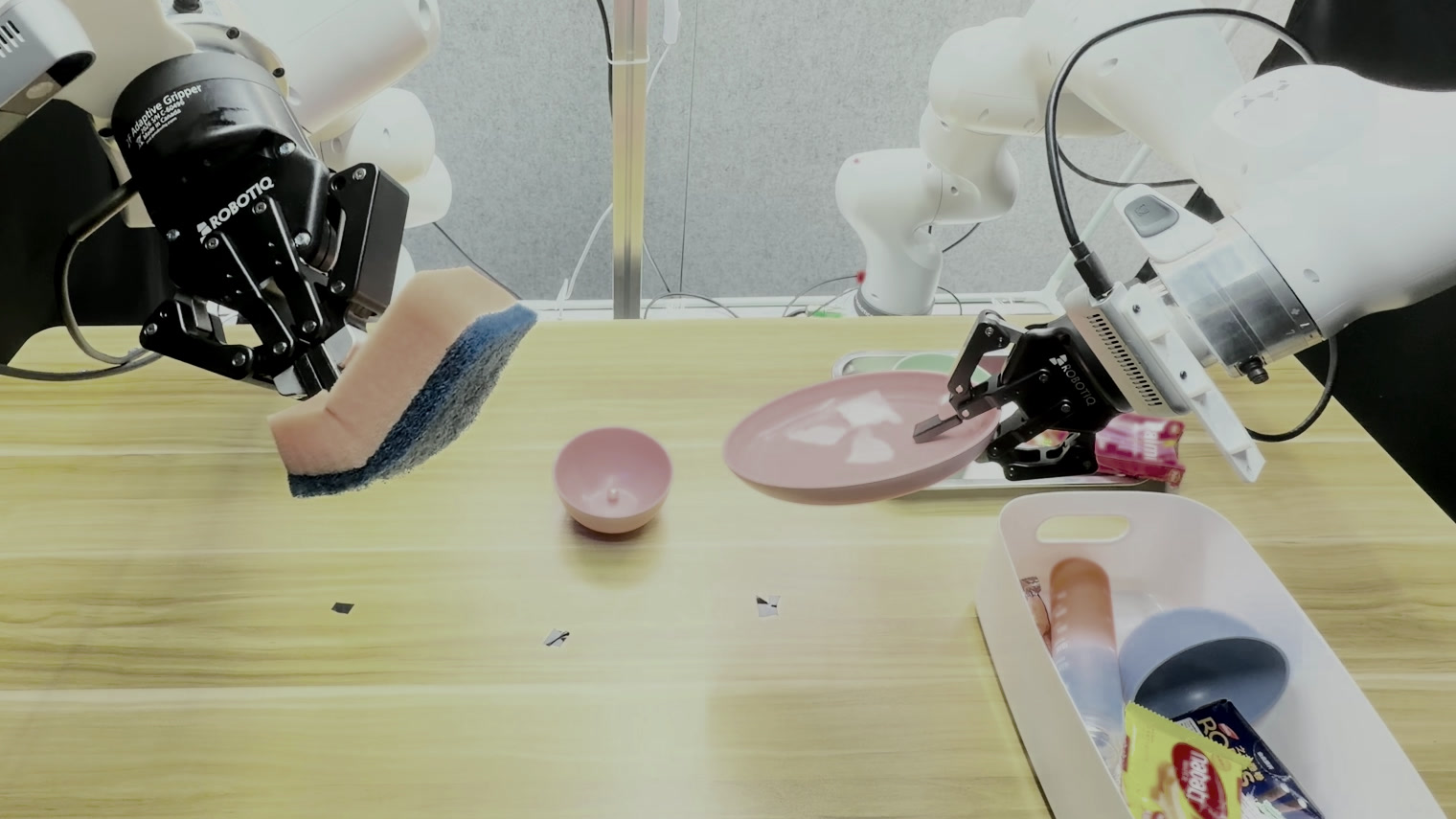}
    \hfill
    \includegraphics[width=0.32\linewidth]{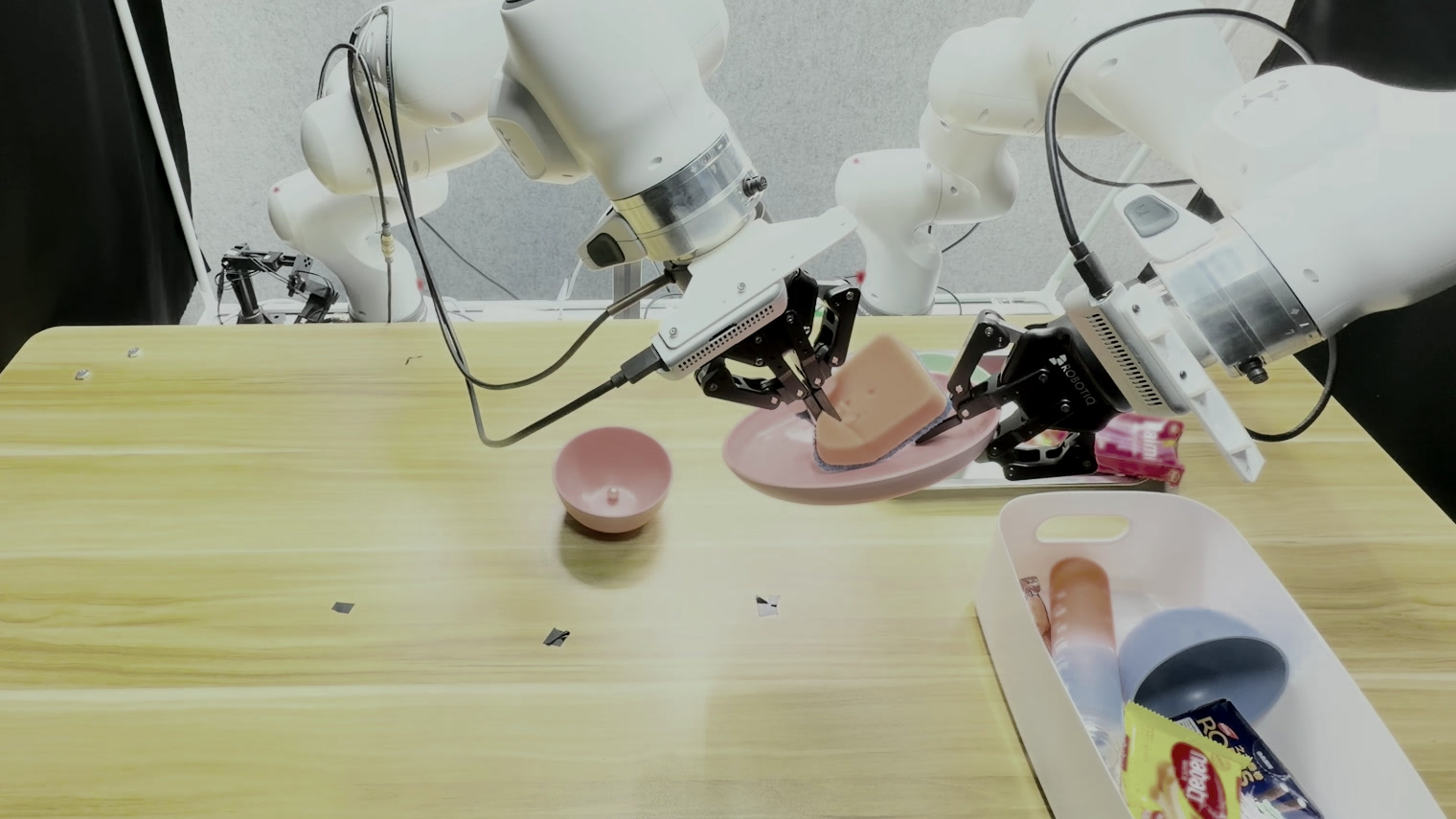}
    \hfill
    \includegraphics[width=0.32\linewidth]{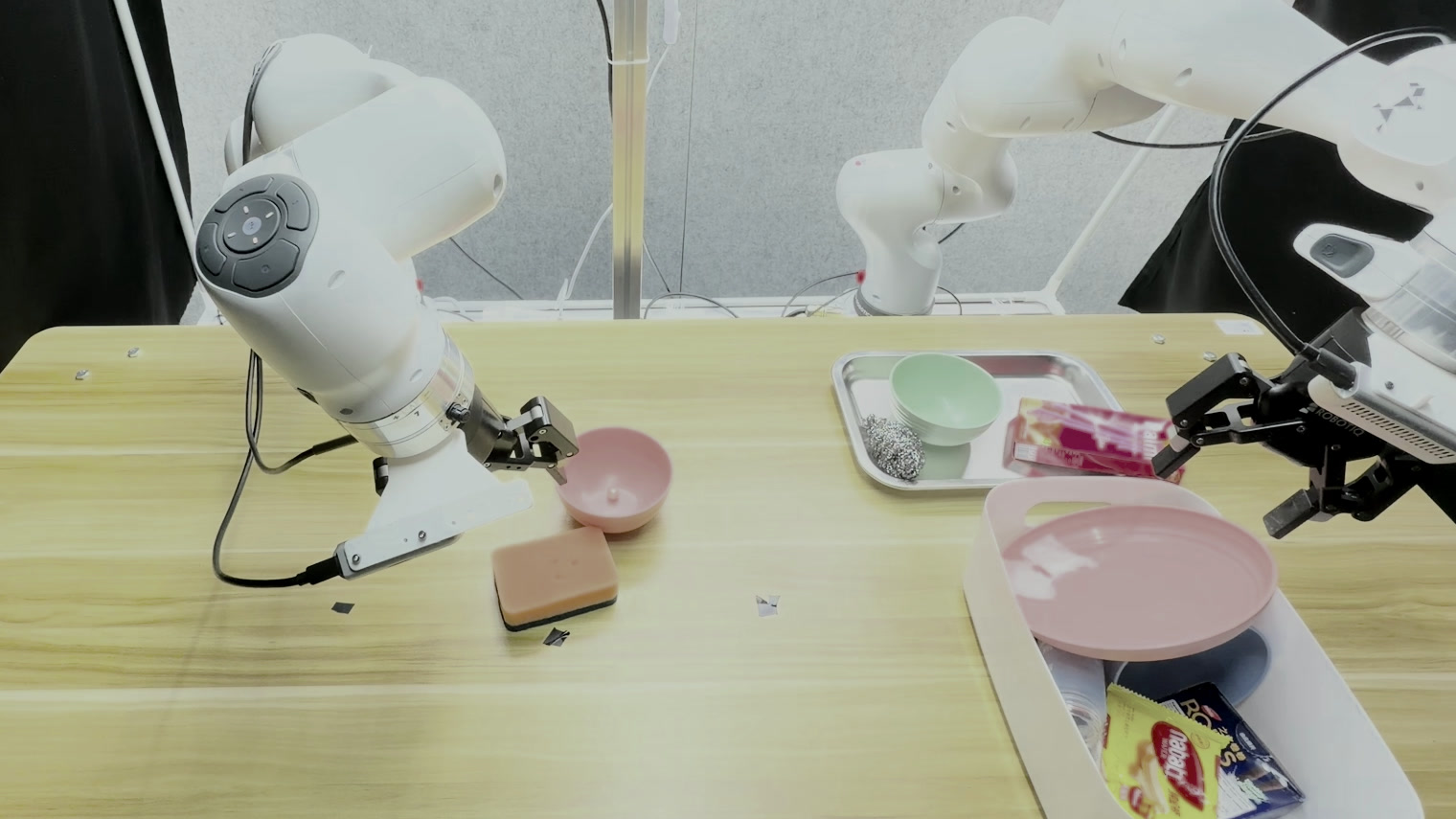}

    \caption{\textbf{From the learned routine to Flash composition.} Each row progresses from left to right. Top row (VLA fullshot): grasp the container, hand it off, and place it in the bin. Middle row (Flash): hold the container, acquire the receiving bowl, and tilt the container to pour steel beads. Bottom row (Flash): acquire the sponge while holding the plate, wipe the plate, and stow it after setting down the sponge. Both Flash procedures execute without online agent planning.}
    \label{fig:physical_flash}
\end{figure}

Figure~\ref{fig:physical_flash} compares fullshot VLA execution with two Flash procedures. The fullshot baseline grasps a container, hands it off, and places it in the bin. The Flash demonstrations instead pour steel beads into a receiving bowl and wipe a held plate with a sponge before stowing the objects. Both procedures execute from task cards without online planning.

Together, these cases illustrate the mechanisms analyzed in Section~\ref{sec:exp:analysis}. Online planning separates destination selection and action order from the learned routine, while local retry enables recovery from a missed grasp (Key Findings~1 and~2). Flash Mode illustrates how analytic transport and reorientation compose the same contact skills into new bimanual procedures (Key Finding~3). The capability extension comes from semantic re-grounding and primitive composition, with the VLA weights unchanged. These recordings provide qualitative demonstrations rather than comparative success-rate estimates.

\section{Related Work}
\label{sec:related}

Harness VLA sits at the intersection of three lines of work: end-to-end robot foundation policies, multimodal LLM agents, and programmatic robot-control systems. We review these areas through the role each assigns to learned policies and explicit control. This perspective clarifies our position: rather than fine-tuning a stronger VLA or expanding the primitive library, we study how a memory-guided agent can turn a frozen VLA into a controllable contact-rich primitive and compose it with fixed analytic controllers.

	\paragraph{VLA Models.} End-to-end Vision-Language-Action (VLA) models map natural-language instructions and visual observations directly to low-level robot actions by extending pretrained vision-language backbones with an action head. The generalist-policy line opened by RT-1~\citep{brohan2022rt} and Octo~\citep{octo_2023}, scaled by RT-2~\citep{brohan2023rt2} and the Open X-Embodiment release~\citep{open_x_embodiment_rt_x_2023}, established the training pattern of co-training a large VLM with cross-embodiment robot demonstrations. OpenVLA~\citep{kim2024openvla} brought this paradigm into the open by combining a Prismatic-style VLM~\citep{karamcheti2024prismatic} with a Llama-2 action tokenizer, while the flow-matching $\pi_0$~\citep{black2025pi0} and $\pi_{0.5}$~\citep{black2025pi05} models reported substantial gains from co-training with heterogeneous data and out-of-distribution language. Recent large-scale systems such as GR00T~\citep{bjorck2025gr00t} and Gemini Robotics~\citep{team2025gemini} continue to scale this pattern to humanoid and general-purpose embodiments, alongside related lines on 3D-aware VLAs~\citep{zhen20243dvla, driess2023palme}, VLM-based imitation~\citep{li2024roboflamingo, huang2023embodied}, and CLIP-conditioned controllers~\citep{shridhar2022cliport, chi2023diffusion, zhao2023act}. Empirically, however, these models exhibit a sharp asymmetry: they are strongest at contact-rich visuomotor phases---in particular for irregular grasping and fixture actuation that defeat analytic controllers---but degrade dramatically on instruction following, long-horizon composition, and out-of-distribution scenes~\citep{kim2024openvla, black2025pi05, pi0-experiment-wild}. This asymmetry motivates a factorization in which the VLA is delegated planner-selected contact-rich operations while a higher-level controller assumes responsibility for language interpretation, target grounding, transport, posture adjustment, navigation, and release.

	\paragraph{LLM-driven Multimodal Agent.} Frontier multimodal large language models have rapidly closed the gap on dense perception, spatial reasoning, and long-horizon tool use. Recently, releases including GPT-5.2~\citep{openai2025gpt52}, Gemini~3~\citep{pichai2025gemini3}, Qwen3-VL~\citep{bai2025qwen3vltechnicalreport}, the Claude~4 family (Sonnet~4.5 and Opus~4.7~\citep{anthropic2025claudesonnet45}), and Llama~4~\citep{llama4} have demonstrated qualitatively stronger physical-scene grounding than the GPT-4o~\citep{openai2024gpt4ocard} / Gemini~1.5~\citep{team2023gemini} generation, while open-weight models such as Molmo~\citep{deitke2025molmo} and Qwen3-VL~\citep{bai2025qwen3vltechnicalreport} make these capabilities broadly available. Targeted spatial benchmarks~\citep{chen2024spatialvlm} and tool-augmented browsing agents~\citep{oaidr, geminidr, wu2025mmsearchr1, geng2025webwatcher, huang2026vision, jin2025search, li2025websailor} further show that, given a closed-loop interface, these backbones can sustain multi-hop perception and decision making over rich, partially observed environments. These advances make it increasingly viable to delegate semantic grounding and deterministic manipulation phases---language parsing, target localization, transport planning, posture adjustment, navigation, and release timing---to a frontier VLM at the top of the agent stack~\citep{team2025gemini, geminirobotics1d52025, wang2023voyager}. We build on this premise but, rather than driving the robot end-to-end with the VLM, place it inside an agentic harness that emits structured primitive calls, observes execution feedback, and iterates---reserving direct action prediction for planner-selected contact-rich phases.

	\paragraph{Programmatic and Tool-Using Robot Agents.} Code-as-policies systems recast robot control as program synthesis: the model writes an executable program that coordinates perception and motion APIs, leveraging the LLM's compositional generalization while keeping low-level control deterministic. Beginning with Code-as-Policies~\citep{liang2023code}, ProgPrompt~\citep{singh2023progprompt}, Instruct2Act~\citep{huang2023instruct2act}, and ChatGPT-for-Robotics~\citep{vemprala2023chatgpt}, this paradigm has been extended with multimodal program synthesis (RoboCodeX~\citep{mu2024robocodex}, ViperGPT~\citep{suris2023vipergpt}, VisProg~\citep{gupta2023visprog}), 3D value-map generation~\citep{huang2023voxposer}, VLM-supervised assembly~\citep{goldberg2024bloxnet}, and long-horizon agentic frameworks~\citep{li2026roboclaw, shi2025maestro}. Harness VLA shares this literature's goal of making language-model reasoning executable through explicit perception and control interfaces, but differs in the action representation: our agentic planner does not synthesize executable code or new control programs. It emits structured JSON primitive invocations inside a closed-loop harness, observes the execution outcome after each primitive, and re-binds the next primitive arguments from current RGB-D evidence and memory. Recent work also studies how agents can grow their own reusable skill libraries: ASPIRE~\citep{lu2026aspire} uses fine-grained execution traces to diagnose failures, synthesize validated repairs, and admit the resulting localization, navigation, motion, grasping, and debugging patterns into a continually expanding skill library. This direction is complementary to ours: ASPIRE expands the agent's reusable skills, whereas Harness VLA deliberately keeps the primitive vocabulary fixed and studies how memory-guided composition can extend a frozen VLA without deployment-time primitive expansion. A second strand draws on software-engineering agents: executable code is empirically a strong action representation for LLM agents~\citep{wang2024codeact}, and systems such as SWE-agent~\citep{yang2024sweagent} and OpenHands~\citep{wang2025openhands} formalize harnesses for iterative editing, execution, and feedback. We borrow the harness principle---structured interfaces, persistent state, execution feedback, and memory---rather than the requirement that actions be represented as code. Reliability is further improved by self-correction~\citep{shinn2023reflexion, madaan2023selfrefine, chen2023selfdebug} and persistent symbolic state across steps~\citep{yoneda2024statler}, while LLM-driven planners~\citep{ahn2022saycan, huang2022inner, mu2023embodiedgpt, zeng2023socratic, liu2023llmplanning, chen2024autotamp, rana2023sayplan, mandi2024roco} demonstrate that LLMs can sequence pretrained primitives or modules, ground 3D scenes, and recover from failures. Two limitations recur across this literature, however. First, task-specific execution traces are rarely represented as reusable, parameterized memory that can be grounded again under new spatial layouts. Second, failure knowledge is seldom distilled into a Global Memory that prevents the planner from repeating known empty grasps, false successes, or unstable staging choices. Voyager~\citep{wang2023voyager} showed that persistent memory can improve embodied agents in a digital sandbox, but this memory-centric design has not been combined with a VLA-backed contact-rich primitive for physical manipulation. Our framework couples the two: a frozen VLA serves as a contact-rich specialist invoked through a single primitive interface, successful primitive sequences are stored in \textbf{Task Specific Memory}, and reusable success rules and failure models are distilled into \textbf{Global Memory}. Together, these two design choices---VLA delegation for contact-rich operations, and memory-augmented primitive composition for everything else---let a single memory-guided agentic planner cover the full task distribution exposed by a given environment, including paraphrased and re-targeted natural-language instructions that defeat monolithic VLAs whose language channel is largely vestigial~\citep{kim2024openvla, black2025pi05}.

\section{Conclusion and Limitations}
\label{sec:conclusion}

We introduced Harness VLA, an asymmetric hierarchical framework that casts a frozen VLA as a single contact-rich primitive interface within an LLM-driven agent, delegating transport, posture, navigation, and release phases to the planner. From the perspective of agent harness engineering, Harness VLA shows that reliable manipulation can come not only from training a stronger policy, but also from surrounding a frozen policy with an auditable execution loop, fixed primitive contracts, memory, feedback, and task-level verification. Evaluations across standard and heavily perturbed benchmarks confirm that Harness VLA achieves state-of-the-art robustness. These results demonstrate that pretrained VLAs are most effective when isolated to contact-rich visuomotor control; abstracting semantic and spatial bindings away from the VLA prevents the catastrophic failures frequently observed in monolithic deployments.

\textbf{Limitations and future work.} Our current framework is limited by an open feedback loop between the high-level planner and low-level VLA. Additionally, the system lacks joint fine-tuning via environmental rewards and human preferences---an issue necessitating future sample-efficient reinforcement learning (e.g., GRPO). Finally, the absence of fine-grained image captioning constrains structural reasoning in highly cluttered, long-horizon tasks. A complementary future direction is to combine our fixed-vocabulary composition strategy with automatic skill-discovery systems such as ASPIRE~\citep{lu2026aspire}: when repeated primitive compositions reveal a missing abstraction, an agent could propose, validate, and admit a new reusable skill while retaining the auditable primitive interface and VLA-backed contact specialization studied here.

\clearpage

\bibliographystyle{corlabbrvnat}
\bibliography{example}

\appendix
\newpage

\section{File-Mediated REPL Protocol}
  \label{app:harness}

  The harness of Section~\ref{sec:method:harness} implements the execution loop of
  Section~\ref{sec:method:formulation} as a synchronous file-mediated
  Read-Eval-Print Loop (REPL). A long-running environment worker owns the live simulator
  state, while the planner $\Pi$ interacts with it only through serialized primitive
  invocations and persisted observations. The planner does not access privileged
  simulator state, object poses, or controller internals.

  At turn $t$, the planner reads the current observation $o_t$, the task language
  $\ell$, and retrieved context from Task Specific Memory and Global Memory. It then
  emits one primitive invocation $c_t \in \mathcal{P}$ by writing a JSON object to
  \texttt{command.json}. The object contains the primitive name in its \texttt{action}
  field and the corresponding keyword arguments. The worker consumes this file, executes
  the selected primitive in the live environment, and writes the next indexed observation
  $o_{t+1}$ together with lightweight execution records. The planner waits for these
  files before selecting the next primitive. Thus, each physical action is followed by
  observation and diagnosis before the rollout continues. Table~\ref{tab:app_repl_files} lists the files exchanged by the REPL.

  \begin{table}[h]
  \centering
  \caption{Files used by the file-mediated REPL. The main text abstracts the observation
  as RGB-D and robot state; the appendix also lists diagnostic records used for
  synchronization and auditability.}
  \label{tab:app_repl_files}
  \small
  \setlength{\tabcolsep}{4pt}
  \begin{tabular}{p{0.27\linewidth}p{0.63\linewidth}}
  \toprule
  File or artifact & Role \\
  \midrule
  \texttt{command.json} &
  Planner-issued primitive invocation $c_t$. \\
  \texttt{state\_NN.json} &
  Step-indexed task language, robot proprioception, and benchmark success signal. \\
  RGB-D / world-map files &
  Benchmark-specific perceptual evidence for semantic identification and metric
  re-grounding. \\
  \texttt{log\_NN.json} &
  Diagnostic record containing the accepted command, primitive status, step counts, and
  failure information when available. \\
  \texttt{done\_NN.flag} or terminal file &
  Synchronization signal indicating that the worker has finished the current primitive. \\
  Task Specific Memory trace &
  Append-only JSONL procedural memory for one task. Each line is one primitive command. \\
  Task Specific Memory summary &
  JSON semantic memory summarizing the outcome, strategy, recovery decisions, and
  failure modes. \\
  Global Memory &
  Cross-task success rules and failure models for using the primitive library. \\
  \bottomrule
  \end{tabular}
  \end{table}

  The index \texttt{NN} increases monotonically. The initial observation is written at
  \texttt{NN=00}; each executed primitive produces the next indexed state, perception
  files, and diagnostic log. These records make the rollout auditable without exposing
  oracle object coordinates to the planner.

  \paragraph{Task Specific Memory.}
  Task Specific Memory stores the reusable structure of a solved task instance. It
  contains a procedural JSONL trace and a semantic JSON summary. The trace records what
  primitive invocations were issued; the summary records why the strategy worked and
  what should be avoided. A simplified summary is:

  \begin{samepage}
  \begin{lstlisting}[frame=single,framerule=0.5pt]
  {"task":"put the black bowl on the wooden tray",
   "success":true,
   "trace_file":"task_specific_memory_put_black_bowl_on_tray_s0.jsonl",
   "strategy":"use VLA for grasping, then analytic transport and release",
   "avoid":["do not reuse reference xyz values",
            "verify placement with the benchmark success signal"]}
  \end{lstlisting}
  \end{samepage}

  The paired procedural trace stores the primitive order:

  \begin{samepage}
  \begin{lstlisting}[frame=single,framerule=0.5pt]
  {"action":"vla_act","prompt":"grasp the black bowl","max_chunks":2}
  {"action":"move_to","xyz":[0.12,-0.08,0.92],"gripper":null}
  {"action":"release"}
  \end{lstlisting}
  \end{samepage}

  The trace is a task-level solution skeleton, not an open-loop trajectory. It records
  the ordering of analytic and VLA-backed primitives, the placement of VLA invocations,
  and the transition points between contact-rich execution, transport, release, and
  verification. Spatial arguments in the stored trace are treated as reference-scene
  bindings. At deployment time, the planner reuses the memory structure but re-grounds
  objects, fixtures, support surfaces, and target poses from the current observation.

  \paragraph{Global Memory.}
  Global Memory stores task-independent operating knowledge for the primitive library.
  diagnose before retrying. A compact example is:

  \begin{samepage}
  \begin{lstlisting}[frame=single,framerule=0.5pt]
  Success rule:
  Use VLA primitives for contact-rich phases such as irregular grasping
  or fixture interaction. After a stable grasp, prefer analytic motion
  for long transport and precise placement.

  Failure model:
  If the gripper closes but the object does not move with the end effector,
  treat the attempt as an empty grasp. Re-localize the object and re-stage
  before retrying.

  Failure model:
  Do not terminate from visual proximity alone. Check the benchmark success
  signal and the latest execution record.
  \end{lstlisting}
  \end{samepage}

  \paragraph{Iterative memory construction.}
  Memory is constructed during interaction rather than written only after the rollout.
  After each primitive, the planner reads the new observation and diagnostic record, then
  classifies the outcome as progress, recoverable failure, or unrecoverable failure.
  Successful rollouts are stored as Task Specific Memory. Recoverable failures remain in
  the trace and are explained in the semantic summary, so that subsequent steps document
  the correction. Failed attempts are also retained as negative evidence and may
  contribute failure models to Global Memory.

  Across attempts, the memory is refined rather than simply accumulated. A later attempt
  can replace the procedural trace if it yields a shorter or more reliable solution,
  while earlier failure observations remain useful as constraints on future planning.
  This separation lets Harness VLA transfer how a task should be solved without replaying
  where objects happened to be in the reference scene.
\section{Primitive Vocabulary and Environment-Specific Extensions}
\label{app:primitives}

This appendix expands the primitive vocabulary of Section~\ref{sec:method:primitives}. We use the same primitive names across all benchmarks. Differences across environments are expressed only through availability, arm binding, and implementation backend; they are not treated as new primitive names unless the embodiment exposes a new degree of freedom.

\begin{table}[h]
\centering
\caption{Primitive availability across the three benchmark embodiments. The exploratory \textsc{reset} utility supports bootstrapping and is not counted as a manipulation primitive.}
\label{tab:app_primitive_availability}
\small
\setlength{\tabcolsep}{5pt}
\begin{tabular}{lccc}
\toprule
Primitive & LIBERO & RoboCasa365 & RoboTwin C2R \\
\midrule
\textsc{move\_to}       & yes & yes & yes \\
\textsc{move\_pose}     & yes & via composition & -- \\
\textsc{rotate\_wrist}  & yes & -- & yes \\
\textsc{rotate\_pitch}  & yes & yes & -- \\
\textsc{set\_gripper}   & yes & yes & yes \\
\textsc{release}        & yes & yes & yes \\
\textsc{vla\_act}       & yes & yes & yes \\
\midrule
\textsc{navigate\_to}   & -- & yes & -- \\
\textsc{move\_base}     & -- & yes & -- \\
\texttt{arm} binding    & -- & -- & yes \\
\bottomrule
\end{tabular}
\end{table}

\paragraph{Universal analytic primitives.}
\textsc{move\_to} is the shared end-effector transport primitive: it takes a world-frame Cartesian target and delegates to the solver available in the current environment. The internal backend may be an operational-space servo, a Jacobian-based controller, or an IK planner, but the exposed primitive semantics are identical. \textsc{move\_pose} extends \textsc{move\_to} by co-varying position with an orientation component such as pitch; when an environment does not expose it directly, the same behavior is expressed as a short composition of \textsc{rotate\_pitch} and \textsc{move\_to}. \textsc{rotate\_wrist} and \textsc{rotate\_pitch} apply yaw and pitch set-points while holding the current spatial position. \textsc{set\_gripper} drives the gripper to an open or closed state, and \textsc{release} is the corresponding open-gripper primitive with a release post-condition. Environment-specific gripper conventions are hidden behind the primitive interface.

\paragraph{Mobile-base and bimanual details.}
RoboCasa365 adds two mobile-base primitives because kitchen-scale tasks require staging outside a fixed-arm workspace. \textsc{navigate\_to} is a composite primitive that drives the base toward a world-frame planar goal, while \textsc{move\_base} is an atomic primitive that applies a local base-velocity set-point for fine repositioning. RoboTwin C2R adds no new manipulation primitive name; instead, each primitive can be bound to the left arm, right arm, or a bimanual task pattern through the \texttt{arm} argument. Handover-style tasks are therefore represented as compositions of \textsc{vla\_act}, analytic transport, and \textsc{release} under this dual-arm binding rather than as a separate primitive.

\paragraph{\textsc{vla\_act}.}
\textsc{vla\_act} is the single learned primitive in the vocabulary. It binds to the frozen VLA used by the current benchmark and executes action chunks conditioned on a prompt and live observations. The planner configures a stop predicate $\tau$, which may correspond to a lift-and-grasp condition, a contact-state condition, a benchmark predicate, or a chunk budget. The same primitive therefore covers grasping, placement, fixture actuation, insertion, and bimanual contact while preserving the planner's responsibility for semantic grounding, spatial re-binding, navigation, and re-staging.

\paragraph{JSON invocation examples.}
All primitives share a compact JSON command format. Representative calls are:
\begin{lstlisting}
{"action": (*@\textbf{"move\_to"}@*),     "xyz": [-0.101, 0.202, 1.05],
 "arm": "auto", "gripper": "open", "tol": 0.012, "max_steps": 80}
{"action": (*@\textbf{"navigate\_to"}@*), "xy": [1.20, -0.35], "tol": 0.05}
{"action": (*@\textbf{"move\_base"}@*),   "forward": 0.10, "lateral": 0.00,
 "turn": -0.15, "steps": 12}
{"action": (*@\textbf{"vla\_act"}@*),     "prompt": "grasp the black bowl",
 "arm": "auto", "max_chunks": 30, "stop": "object_lifted"}
\end{lstlisting}
The exact numerical tolerances and stop predicates are benchmark-specific, but the planner always interacts through these primitive names.
\section{Details about the Evaluation Benchmark}
\label{app:benchmarks}

\begin{figure*}[t]
\centering

\setlength{\tabcolsep}{2pt}
\renewcommand{\arraystretch}{0}

\begin{tabular}{cc}
\includegraphics[width=0.48\textwidth,height=3.2cm,keepaspectratio]{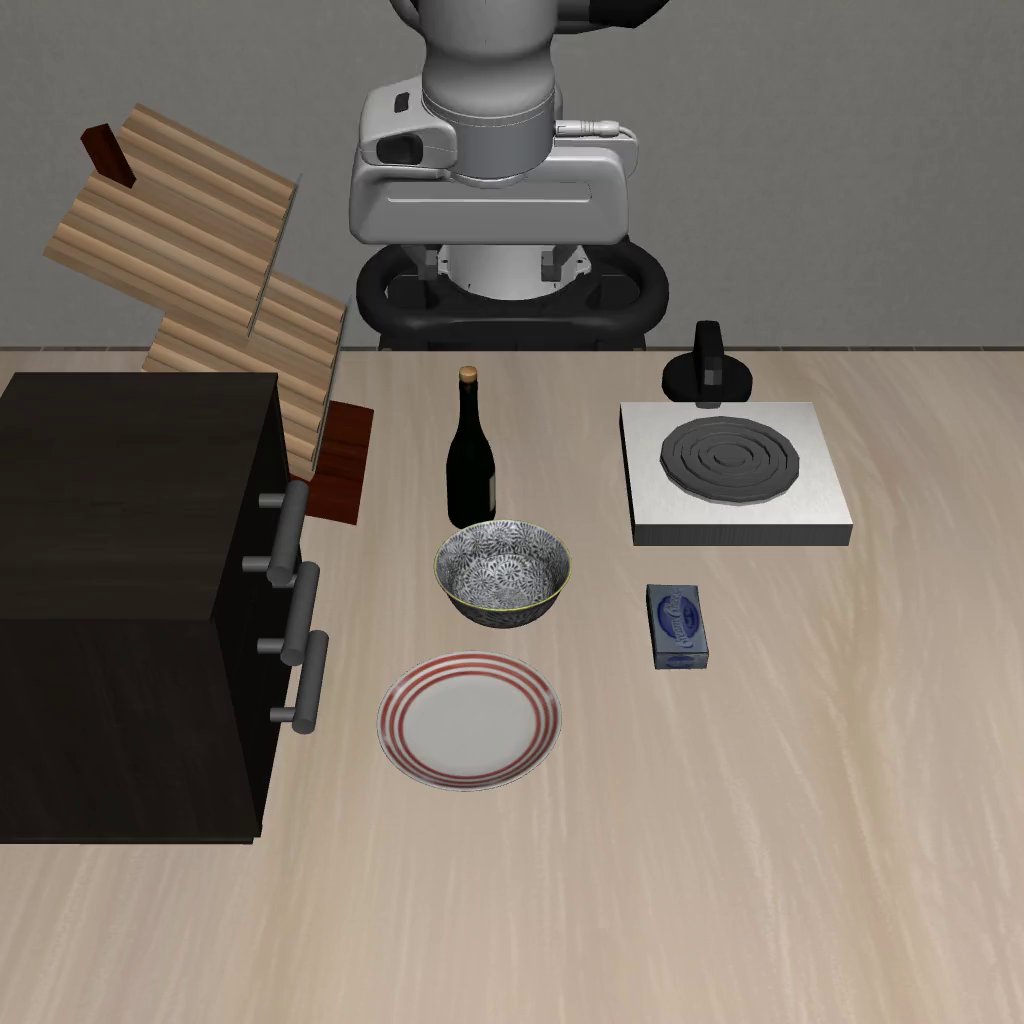} &
\includegraphics[width=0.48\textwidth,height=3.2cm,keepaspectratio]{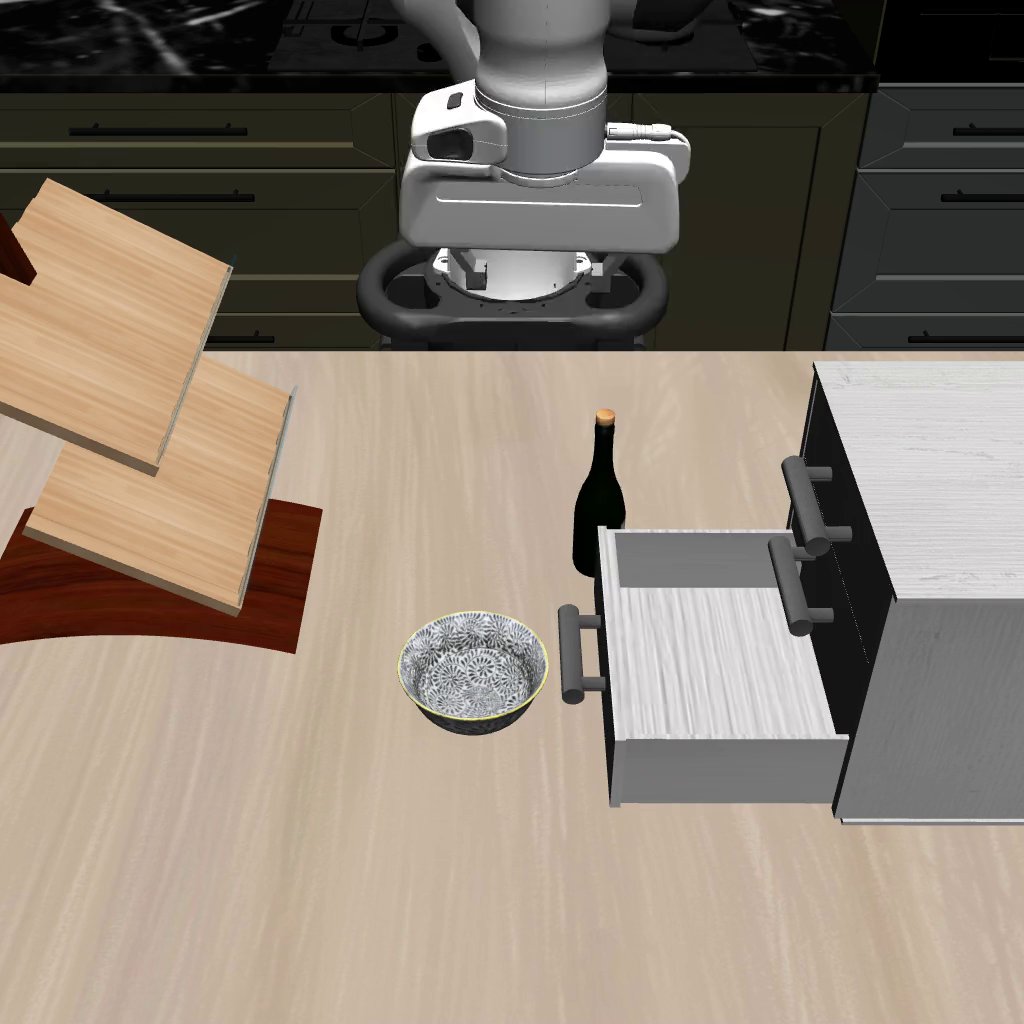} \\

\includegraphics[width=0.48\textwidth,height=3.2cm,keepaspectratio]{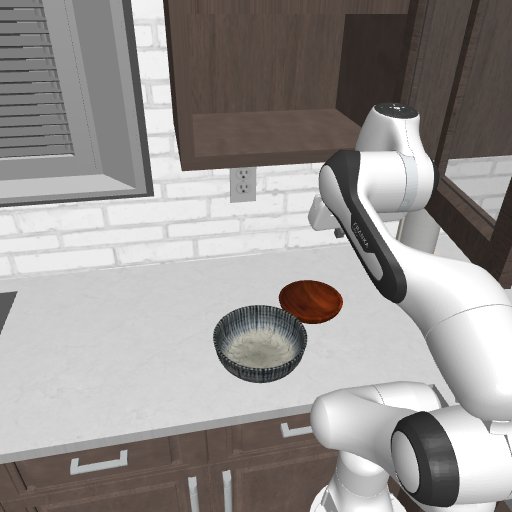} &
\includegraphics[width=0.48\textwidth,height=3.2cm,keepaspectratio]{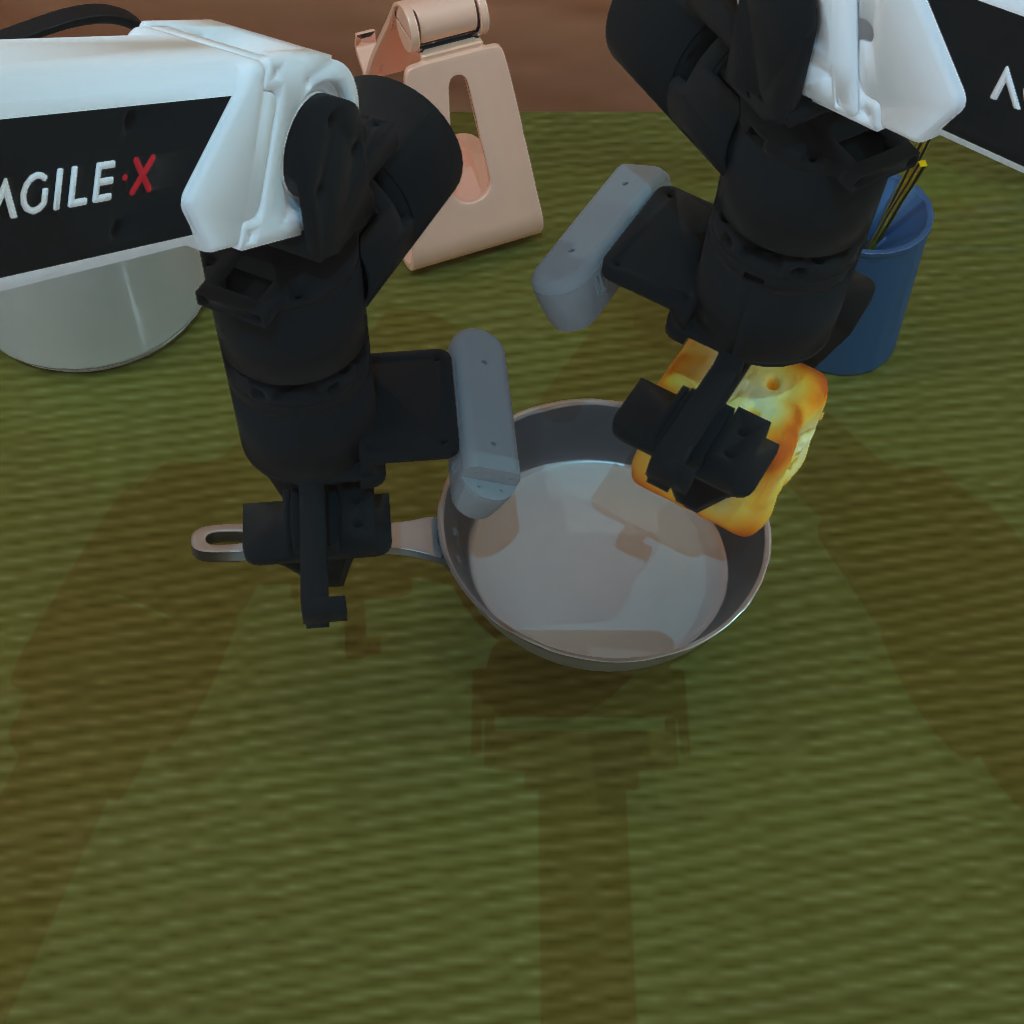} \\
\end{tabular}

\caption{
Overview of representative environments across the four benchmark families used in our evaluation.
Each benchmark captures a distinct manipulation setting: structured tabletop manipulation (LIBERO), robustness under distribution shift (LIBERO-Pro), long-horizon kitchen manipulation (RoboCasa365), and bimanual manipulation under clean-to-randomized settings (RoboTwin C2R).
}
\label{fig:benchmark_overview}

\end{figure*}

This appendix records the benchmark composition, task splits, rollout protocol, and success criteria used in our evaluation. Figure~\ref{fig:benchmark_overview} shows the four benchmark families.
We evaluate on four benchmark families: LIBERO, LIBERO-Pro, RoboCasa365, and RoboTwin C2R. LIBERO, LIBERO-Pro, and RoboCasa365 use a few-shot protocol in which seed $s_0$ (seed 0) for each task serves only as the exploratory reference seed for Task Specific Memory construction. On this seed, the agent searches for a successful primitive sequence and stores the resulting audit summary and JSONL command trace as Task Specific Memory. Seed $s_0$ is not counted in reported evaluation. Reported evaluation rollouts are run on held-out seeds that retrieve and re-ground the corresponding Task Specific Memory under new initial states. RoboTwin C2R uses a separate zero-shot clean-to-randomized protocol.

Across all benchmarks, task success is determined by the benchmark-provided binary completion predicate. A rollout is counted as successful if the task completion predicate is satisfied before the episode horizon or maximum step budget is exhausted. Primitive-level post-conditions, such as the return condition of \textsc{vla\_act} or \textsc{release}, only determine when an individual primitive returns control to the planner; they are not used as substitutes for the final task success predicate.

\subsection{LIBERO Evaluation Benchmark}
\label{app:benchmark:libero}

LIBERO~\citep{liu2023libero} is a language-conditioned manipulation benchmark organized into multiple task suites. We evaluate on four standard suites: \textsc{LIBERO-Spatial}, \textsc{LIBERO-Object}, \textsc{LIBERO-Goal}, and \textsc{LIBERO-10}. \textsc{LIBERO-Spatial} contains tasks that vary spatial relations, \textsc{LIBERO-Object} varies the target object identity, \textsc{LIBERO-Goal} varies the goal predicate under related scenes, and \textsc{LIBERO-10} contains longer-horizon compositional manipulation tasks.

Each suite contains 10 language-conditioned tasks. For each task, seed $s_0$ (seed 0) is used only to explore the task and construct Task Specific Memory. Reported evaluation uses ten held-out seeds, denoted $s_1$--$s_{10}$, which retrieve this Task Specific Memory and ground it under new initial states. Thus, each LIBERO suite contains $10$ tasks $\times$ $10$ evaluation seeds $=100$ reported rollouts, and the four suites contain 400 reported rollouts in total (Table~\ref{tab:app_libero_protocol}).

\begin{table}[h]
    \centering
    \caption{LIBERO evaluation protocol.}
    \label{tab:app_libero_protocol}
    \begin{tabular}{lccc}
        \toprule
        Suite & Tasks & Eval seeds per task & Reported rollouts \\
        \midrule
        \textsc{LIBERO-Spatial} & 10 & 10 & 100 \\
        \textsc{LIBERO-Object}  & 10 & 10 & 100 \\
        \textsc{LIBERO-Goal}    & 10 & 10 & 100 \\
        \textsc{LIBERO-10}      & 10 & 10 & 100 \\
        \midrule
        Total & 40 & -- & 400 \\
        \bottomrule
    \end{tabular}
\end{table}

Success rates are computed using the predicate-based rule defined at the beginning of this appendix.

\subsection{LIBERO-Pro Evaluation Benchmark}
\label{app:benchmark:liberopro}

LIBERO-Pro~\citep{Lagrange1788} extends the LIBERO task families with controlled perturbations. We evaluate four task families: \textsc{Spatial}, \textsc{Object}, \textsc{Goal}, and \textsc{LIBERO-10}. Each task family is evaluated under two perturbation settings, denoted \textbf{T} and \textbf{S}. \textbf{T} refers to the task or instruction-redirection setting, where the instruction is redirected to another valid target object or goal condition. \textbf{S} refers to the swap or position-swap setting, where object initial positions are swapped or rearranged while the instruction remains fixed.

We evaluate eight LIBERO-Pro cells: \textsc{Spatial-T}, \textsc{Spatial-S}, \textsc{Object-T}, \textsc{Object-S}, \textsc{Goal-T}, \textsc{Goal-S}, \textsc{LIBERO-10-T}, and \textsc{LIBERO-10-S}. Each cell contains 10 tasks. As in LIBERO, seed $s_0$ (seed 0) is used only to explore the task and construct Task Specific Memory; reported evaluation uses seeds $s_1$--$s_{10}$, which retrieve the stored Task Specific Memory and ground it under new initial states. Each cell therefore contains $10$ tasks $\times$ $10$ evaluation seeds $=100$ reported rollouts, for a total of 800 reported rollouts.

\begin{table}[h]
    \centering
    \caption{LIBERO-Pro evaluation protocol.}
    \label{tab:app_liberopro_protocol}
    \begin{tabular}{lccc}
        \toprule
        Evaluation cell & Tasks & Eval seeds per task & Reported rollouts \\
        \midrule
        \textsc{Spatial-T}   & 10 & 10 & 100 \\
        \textsc{Spatial-S}   & 10 & 10 & 100 \\
        \textsc{Object-T}    & 10 & 10 & 100 \\
        \textsc{Object-S}    & 10 & 10 & 100 \\
        \textsc{Goal-T}      & 10 & 10 & 100 \\
        \textsc{Goal-S}      & 10 & 10 & 100 \\
        \textsc{LIBERO-10-T} & 10 & 10 & 100 \\
        \textsc{LIBERO-10-S} & 10 & 10 & 100 \\
        \midrule
        Total & 80 & -- & 800 \\
        \bottomrule
    \end{tabular}
\end{table}

Table~\ref{tab:app_liberopro_protocol} summarizes the evaluation matrix. Success rates are computed using the predicate-based rule defined at the beginning of this appendix.

\subsection{RoboCasa Evaluation Benchmark}
\label{app:benchmark:robocasa}

RoboCasa365 extends the evaluation to kitchen household manipulation. We use the RoboCasa365 \texttt{target50} split, which consists of three task groups: \textsc{Atomic-Seen}, \textsc{Composite-Seen}, and \textsc{Composite-Unseen}. \textsc{Atomic-Seen} contains 18 atomic tasks corresponding to short-horizon kitchen operations. \textsc{Composite-Seen} contains 16 composite tasks whose templates are also present in the pretraining set. \textsc{Composite-Unseen} contains 16 composite tasks whose templates are held out from pretraining and appear only in target evaluation. Here, ``seen'' and ``unseen'' refer to whether the task template appears in the pretraining set, not whether the exact episode, trajectory, or scene has been observed.

RoboCasa365 uses five held-out evaluation seeds, $s_1$--$s_5$, for every task in all three splits. Seed $s_0$ (seed 0) is used only for exploration and Task Specific Memory construction and is excluded from evaluation. This gives $18 \times 5 = 90$ rollouts for \textsc{Atomic-Seen} and $16 \times 5 = 80$ for each composite split, totaling $250$ rollouts over $50$ tasks (Table~\ref{tab:app_robocasa_protocol}). All three planners, GPT-5.5, Opus-4.7, and GPT-6 Astra, use this protocol. Overall success gives equal weight to each task and, since all tasks have five evaluation seeds, equals the success rate pooled over all $250$ rollouts.

\begin{table}[h]
    \centering
    \caption{RoboCasa365 evaluation protocol.}
    \label{tab:app_robocasa_protocol}
    \begin{tabular}{lccc}
        \toprule
        Split & Tasks & Eval seeds per task & Reported rollouts \\
        \midrule
        \textsc{Atomic-Seen}      & 18 & 5 & 90 \\
        \textsc{Composite-Seen}   & 16 & 5 & 80 \\
        \textsc{Composite-Unseen} & 16 & 5 & 80 \\
        \midrule
        Total & 50 & 5 & 250 \\
        \bottomrule
    \end{tabular}
\end{table}

Success rates are computed using the predicate-based rule defined at the beginning of this appendix.

\subsection{RoboTwin Clean-to-Randomized Evaluation Benchmark}
\label{app:benchmark:robotwin}

RoboTwin C2R is a bimanual manipulation benchmark with 50 tasks. The task set covers pick-and-place, stacking, ordering, handover, dual-arm transport, articulated-object interaction, pressing and clicking, rotation, scanning, and container-placement behaviors.

RoboTwin C2R uses a separate Clean-to-Randomized protocol. For each task, the Task Specific Memory trace is obtained from one official scripted-expert-verified seed in the \texttt{demo\_clean} setting. Evaluation is then performed directly in the official \texttt{demo\_randomized} setting on five scripted-expert-verified randomized seeds. The expert verification step is used only to ensure that the sampled task instances are feasible under the official task definition; it is independent of our method and does not use Harness VLA rollouts for seed selection. No additional trace search, fine-tuning, or task-level adaptation is performed in the randomized setting. This protocol evaluates zero-shot transfer from a clean-setting trace to randomized task instances.

We evaluate all 50 RoboTwin C2R tasks. Each task is evaluated on 5 seeds in the \texttt{demo\_randomized} setting, resulting in $50 \times 5 = 250$ reported rollouts (Table~\ref{tab:app_robotwin_protocol}).

\begin{table}[h]
    \centering
    \caption{RoboTwin clean-to-randomized evaluation protocol.}
    \label{tab:app_robotwin_protocol}
    \begin{tabular}{lccc}
        \toprule
        Protocol & Tasks & Eval seeds per task & Reported rollouts \\
        \midrule
        RoboTwin C2R & 50 & 5 & 250 \\
        \bottomrule
    \end{tabular}
\end{table}

Success rates are computed using the predicate-based rule defined at the beginning of this appendix, with the completion predicate provided by the official RoboTwin C2R task-specific evaluator.

\subsection{Benchmark Summary}
\label{app:benchmark_summary}

Table~\ref{tab:app_benchmark_summary} summarizes the scale of each evaluation benchmark. LIBERO and LIBERO-Pro use seed $s_0$ (seed 0) only to construct Task Specific Memory and report evaluation on ten held-out seeds $s_1$--$s_{10}$ per task. RoboCasa365 uses the same reference-seed convention and evaluates every task on five held-out seeds $s_1$--$s_5$ across all three splits. RoboTwin C2R uses clean-to-randomized evaluation where the Task Specific Memory is obtained from one expert-verified \texttt{demo\_clean} seed and evaluated on expert-verified \texttt{demo\_randomized} seeds.

\begin{table}[h]
    \centering
    \caption{Summary of evaluation benchmarks.}
    \label{tab:app_benchmark_summary}
    \begin{tabular}{lccc}
        \toprule
        Benchmark & Tasks & Trials/task & Reported rollouts \\
        \midrule
        LIBERO & 40 & 10 & 400 \\
        LIBERO-Pro & 80 & 10 & 800 \\
        RoboCasa365 & 50 & 5 & 250 \\
        RoboTwin C2R & 50 & 5 & 250 \\
        \bottomrule
    \end{tabular}
\end{table}

\section{VLA Model Instantiations}
\label{app:vla_instantiations}

We instantiate different vision-language-action models across benchmarks, which are uniformly abstracted as the single contact-rich primitive \textsc{vla\_act} within the Harness VLA framework.
\subsection{\texorpdfstring{$\piRLinf$}{pi\_RLinf}: RLinf-released LIBERO checkpoint}

For LIBERO and LIBERO-Pro, we use the RLinf-released \texttt{pi05\_libero130\_fullshot} checkpoint, denoted $\piRLinf$, as the frozen vision-language-action policy. It is based on the $\pi_{0.5}$ architecture, and we directly adopt this official $\pi_{0.5}$-SFT checkpoint as a frozen \textsc{vla\_act} contact-rich execution primitive within the Harness VLA framework.

\paragraph{Architecture.}
$\piRLinf$ follows the $\pi_{0.5}$ vision-language-action architecture, which encodes multimodal inputs including visual observations $I_t$, language instructions $\ell$, and robot state $q_t$ into a unified transformer representation. The model is initialized from a pretrained vision-language backbone and aligned to robot action spaces via supervised learning.

Consistent with $\pi_{0.5}$, $\piRLinf$ supports hierarchical inference, where high-level semantic subtask prediction and low-level action generation are jointly modeled within a single policy.

\paragraph{Action Modeling.}
$\piRLinf$ adopts the two-stage inference paradigm introduced in $\pi_{0.5}$. Given an observation and language instruction, the model first predicts a high-level subtask $\hat{\ell}$ (e.g., “pick up the plate”), which is then used to condition low-level action generation.

The low-level policy produces continuous action chunks $a_{t:t+H}$, represented either via FAST tokenization or flow-based continuous modeling, enabling stable contact-rich manipulation.

\paragraph{Training.}
The model is supervised fine-tuned on the LIBERO-130 dataset following the $\pi_{0.5}$ training protocol, resulting in the official $\pi_{0.5}$-SFT checkpoint. In this work, no additional training or adaptation is performed, and the model is used in a fully frozen manner during evaluation.

\paragraph{Performance.}
On the LIBERO benchmark, $\piRLinf$ achieves a success rate of 95.3\%, demonstrating strong in-distribution manipulation capability. On LIBERO-Pro, which introduces instruction perturbations and compositional variations, performance drops to 50.0\%, indicating sensitivity to distribution shifts.

\paragraph{Role in This Work.}
In this paper, $\piRLinf$ is used as a frozen low-level execution module within the Harness VLA framework, serving as the contact-rich manipulation primitive for LIBERO and LIBERO-Pro tasks.

\subsection{RLDX-1}

RLDX-1 is used in RoboCasa365 for kitchen manipulation tasks. It is a large-scale vision-language-action (VLA) foundation model designed for general dexterous manipulation across diverse robotic embodiments. In this work, we directly use the official RLDX-1 checkpoint and keep the model fully frozen during evaluation, treating it as a \textsc{vla\_act} contact-rich execution primitive within the Harness VLA framework.

\paragraph{Architecture.}
RLDX-1 adopts a Multi-Stream Action Transformer (MSAT) as its core action modeling architecture. The system first encodes multi-frame video observations and language instructions using a Vision-Language Model (VLM) based on Qwen3-VL 8B, and extracts action-relevant representations via cognition tokens.

A memory module is further introduced to aggregate historical cognition features, producing history-aware representations. The action model builds on MSAT, which decouples cognition and action streams and optionally introduces a physics stream when physical signals are available. These streams are jointly modeled via cross-stream self-attention, enabling unified processing of vision, language, state, and physical signals.

\paragraph{Action Modeling.}
RLDX-1 is trained using a flow-matching diffusion transformer for continuous action prediction. The model learns a velocity field that maps noisy action trajectories to clean action sequences, and generates future actions via iterative denoising.

During inference, the model produces action chunks in a chunk-wise manner and executes partial chunks sequentially to enable stable closed-loop control. The model also jointly models physical signals when available, improving contact-rich manipulation capability.

\paragraph{Training.}
This work directly uses the official RLDX-1 checkpoint and keeps all parameters frozen during evaluation, without any additional training or fine-tuning. The model has already undergone multi-stage training in its original pipeline and is used here as a unified execution policy.

\paragraph{Performance.}
On the RoboCasa365 benchmark, RLDX-1 achieves 60.0\% on Atomic-Seen tasks, 21.3\% on Composite-Seen tasks, and 5.0\% on Composite-Unseen tasks, with an overall weighted success rate of 30.0\%. These results show strong performance on atomic contact-rich manipulation tasks, while performance significantly degrades on compositional and out-of-distribution settings.

\paragraph{Role in This Work.}
In this paper, RLDX-1 is used as a frozen low-level execution module within the Harness VLA framework, serving as the contact-rich manipulation primitive for RoboCasa365 kitchen manipulation.

\subsection{LingBot-VLA}

LingBot-VLA~\citep{wu2026pragmatic} is the vision-language-action model behind our RoboTwin backend for bimanual manipulation. It is a large-scale VLA foundation model designed for continuous robotic control across diverse real-world embodiments. In this work, we use our RoboTwin-post-trained LingBot-VLA checkpoint as a frozen low-level execution module within the Harness VLA framework.

\paragraph{Architecture.}
The model is built upon a pre-trained Qwen2.5-VL vision-language backbone and is extended with a Mixture-of-Transformers (MoT) architecture that separates vision-language reasoning and action generation into dedicated transformer pathways. These pathways are coupled via shared self-attention, enabling unified multimodal sequence modeling while mitigating cross-modal interference. An action expert module is introduced to predict continuous control signals conditioned on multimodal embeddings.

\paragraph{Action Modeling.}
LingBot-VLA adopts a flow-matching formulation for continuous action prediction. To improve temporal consistency in long-horizon manipulation, it employs chunked action decoding, where a fixed-length action sequence is predicted autoregressively in a single forward pass. The chunk size is set to $T = 50$, enabling stable and temporally coherent control.

\paragraph{Training.}
The model is first pre-trained on large-scale real-world dual-arm teleoperation data collected across 9 robotic embodiments, providing broad cross-embodiment generalization. It is then further adapted via supervised fine-tuning (SFT) on RoboTwin manipulation trajectories to specialize in bimanual manipulation tasks. After this post-training stage, the checkpoint is kept frozen in all direct VLA and Harness VLA evaluations.

\paragraph{Training Configuration.}
We summarize the key hyperparameters governing post-training in Table~\ref{tab:lingbot_hyperparams}. These parameters correspond to the configuration used for all LingBot-VLA post-training experiments reported in this work.

\begin{table}[h]
\centering
\caption{Post-training configuration of LingBot-VLA on RoboTwin.}
\label{tab:lingbot_hyperparams}
\begin{tabular}{ll}
\toprule
\textbf{Category} & \textbf{Configuration} \\
\midrule
\multicolumn{2}{l}{\textit{Optimization}} \\
Optimizer & AdamW \\
Learning rate & $1 \times 10^{-4}$ \\
Vision encoder LR & $1 \times 10^{-6}$ \\
Weight decay & 0 \\
Loss function & L1 Flow Matching (L1\_FM) \\
\midrule
\multicolumn{2}{l}{\textit{Sequence Modeling}} \\
Chunk size & 50 \\
Max sequence length & 2048 \\
Flow steps & 10 \\
Max action dimension & 75 \\
Max state dimension & 75 \\
\midrule
\multicolumn{2}{l}{\textit{Training Setup}} \\
Global batch size & 256 \\
Image resolution & $224 \times 224$ \\
Camera views & top + wrist left + wrist right \\
\midrule
\multicolumn{2}{l}{\textit{System}} \\
Precision & mixed precision (bf16/fp32) \\
Distributed training & FSDP2 \\
\bottomrule
\end{tabular}
\end{table}

\paragraph{Performance.}
Under the RoboTwin randomized evaluation setting, LingBot-VLA achieves a 50.4\% success rate in the direct frozen-agent configuration (i.e., as a standalone policy without agent-level decomposition or external planning). This direct baseline differs from the external $\pi_{0.5}$ comparison in the main table: LingBot-VLA is the RoboTwin-specialized frozen VLA backend used by Harness VLA, whereas $\pi_{0.5}$ is a representative external VLA baseline. The result indicates that LingBot-VLA already provides a strong and stable contact-rich manipulation capability before agent-level decomposition.

\paragraph{Role in This Work.}
In this paper, LingBot-VLA is used as a frozen execution module within Harness VLA, serving as the low-level contact-rich manipulation primitive for RoboTwin bimanual control.

\subsection{Unified Abstraction}

Across all benchmarks, heterogeneous vision-language-action models are uniformly abstracted as interchangeable contact-rich execution primitives. The LLM planner is responsible for semantic grounding, spatial decomposition, and long-horizon task planning, while each VLA is invoked solely for localized interaction execution conditioned on the current observation.

\section{Agent Prompt Specification}
\label{app:agent_prompt}

This appendix specifies the task prompts used by the LLM planner in Harness VLA. The
prompt is not merely a natural-language task instruction. It is the operating manual
given to the agent before each rollout: it defines the file-mediated interaction
protocol, the observation files that may be used for perception, the allowed primitive
vocabulary, the interface to the frozen VLA primitive, the use of Task Specific Memory,
and the output artifacts that must be written for reproducibility.

All benchmark prompts follow a shared-core design. A single benchmark-independent
prompt template defines the agent's responsibilities, and each benchmark instantiates
the slots corresponding to the success predicate, robot embodiment, camera files,
primitive schemas, VLA backend, Task Specific Memory paths, and known recovery rules. This shared
structure is important because the empirical comparison in the paper evaluates the
same agentic harness across LIBERO / LIBERO-Pro, RoboCasa365, and RoboTwin C2R rather
than hand-crafting unrelated controllers for each environment.

\subsection{Shared Prompt Core}
\label{app:agent_prompt:shared}

The shared prompt is written in the second person because it is directly addressed to
the agent. Its first paragraph defines the agent role:

\begin{lstlisting}[frame=single,framerule=0.5pt]
You are an LLM-in-the-loop hybrid manipulation agent for {BENCHMARK}.
A benchmark driver is already running and waiting for your commands.
Your job is to complete the task by reading the task state, localizing
objects from perception, choosing and executing available primitives,
invoking the VLA when contact-rich behavior is needed, and writing a
reproducible audit.
\end{lstlisting}

The remainder of the shared prompt is organized into the modules summarized in
Table~\ref{tab:app_prompt_shared_modules}. Each module is present in all benchmark
prompts, while benchmark-specific prompts fill in concrete fields such as
\texttt{state.libero\_terminated}, \texttt{state.success}, \texttt{eval\_success},
camera names, and primitive schemas.

\begin{table}[h]
\centering
\caption{Shared modules in the agent task prompt. Each benchmark-specific prompt
keeps this structure and fills in environment-specific details.}
\label{tab:app_prompt_shared_modules}
\small
\setlength{\tabcolsep}{4pt}
\begin{tabular}{p{0.22\linewidth}p{0.68\linewidth}}
\toprule
Prompt module & Information given to the agent \\
\midrule
Role and success signal &
Closed-loop control; optimize the benchmark predicate, not a visual guess. \\
Perception isolation &
No ground-truth poses or simulator internals; localize from RGB-D and world maps. \\
File-based REPL &
Write one JSON command, wait for execution, read refreshed artifacts, then iterate. \\
Primitive vocabulary &
Allowed primitive schemas and controller semantics, including gripper, arm, and step
conventions. \\
VLA division of labor &
VLA for contact-rich phases; analytic primitives for grounding, staging, transport,
release, and recovery. \\
Task language &
State-file task language is authoritative; do not infer tasks from filenames or
indices. \\
Seed 0 Task Specific Memory &
JSON audit for strategy and failure modes; JSONL trace for primitive execution order. \\
Global Memory &
Reusable success rules and failure observations that provide additional context beyond
seed 0 Task Specific Memory. \\
Closed-loop recovery &
Verify state, logs, RGB, and geometry after every primitive; diagnose before retrying. \\
Budget and reset policy &
Track budget and reset policy; reset is disabled in strict evaluation. \\
Output discipline &
Write audit and command trace for both successful and failed rollouts. \\
\bottomrule
\end{tabular}
\end{table}

\subsection{Perception and File-Mediated Control}
\label{app:agent_prompt:perception}

The shared prompt makes perception isolation explicit, explicitly prohibiting access to privileged information (e.g., ground-truth object poses or simulator internal states) to enforce a realistic partial-observation setting and prevent any reliance on oracle-level environment access during decision making. The agent receives object names and proprioception from the state file, but not object coordinates. It must localize entities by choosing pixels in RGB images and indexing the corresponding precomputed world map, thereby grounding all spatial reasoning in perceptual inputs rather than hidden state variables. The common localization instruction is:

\begin{samepage}
\begin{lstlisting}[frame=single,framerule=0.5pt]
1. Identify the relevant object, fixture, target surface, or relation
   landmark from RGB.
2. Pick pixels on the visible surface of that entity.
3. Index the matching precomputed world map at those pixels.
4. Sample multiple stable pixels and use a robust statistic, typically
   the median.
5. Avoid rims, object edges, table gaps, holes, reflections, and
   background pixels.
6. Re-localize whenever the robot, camera, object, base, fixture, or
   grasp state changes.
\end{lstlisting}
\end{samepage}

This perception rule is paired with the same REPL-style execution contract used
throughout the paper:

\begin{samepage}
\begin{lstlisting}[frame=single,framerule=0.5pt]
1. Write one JSON command to {WORKDIR}/command.json.
2. Wait until the driver finishes that primitive, typically via
   done_NN.flag, log_NN.json, or a benchmark-specific terminal file.
3. Read the new state_NN.json, log_NN.json, images, depth maps, and
   world maps.
4. Decide the next command from the new evidence.
\end{lstlisting}
\end{samepage}

Thus, the prompt enforces the same closed-loop behavior used in the framework
description: every primitive call is treated as an experiment whose result must be
observed before the next command is issued.

\subsection{Seed 0 Task Specific Memory}
\label{app:agent_prompt:task_specific_memory}

The most important memory-related part of the prompt is the instruction for using
seed 0 Task Specific Memory. Task Specific Memory is not a plain demonstration to replay. It is a structured
memory object that separates semantic strategy from concrete primitive execution.
Across benchmarks, the prompt tells the agent to look for two complementary files:
\begin{lstlisting}[frame=single,framerule=0.5pt]
  {TASK}\_s0.json (Task Specific Memory audit JSON)
  {TASK}\_s0.jsonl (Task Specific Memory command JSONL)
  \end{lstlisting}

The JSON file is the audit and strategy summary for the seed 0 rollout. It records the outcome of the reference run and provides high-level notes about the solution strategy, useful primitive choices, recovery decisions, and failure modes observed during exploration. It is not replayed as an action trace; instead, the agent uses it to interpret the reference solution before consulting the JSONL command trace for the concrete primitive order. Depending on the benchmark, this JSON includes the rollout outcome,  success status, command or step counts, strategy notes, failure observations, and a summary of the final state.

The JSONL file is the executable command trace. Each line stores one JSON primitive
  issued by the agent during the reference rollout. The agent reads this trace to recover
  the procedural structure of the solution: the ordering of primitive calls, the choice
  of analytic versus VLA-backed actions, the number and placement of VLA invocations,
  and the transition points between perception, staging, contact-rich execution,
  transport, release, and verification. The trace is used as a structural prior rather
  than a trajectory to replay; all spatial arguments are re-grounded from the current
  observation before execution.

The prompt gives the agent the following rule:

\begin{samepage}
\begin{lstlisting}[frame=single,framerule=0.5pt]
Use the JSON audit to understand why the strategy worked and what to
avoid. Use the JSONL trace to understand what was executed and in what
order. Reuse the Task Specific Memory procedural structure, but never replay literal
coordinates. Previous xyz, xy, quat, pixel locations, base poses, and
fixture coordinates belong to the seed 0 scene. Re-localize every current
object, destination, support surface, relation landmark, and fixture from
the current images and world maps.
\end{lstlisting}
\end{samepage}

This rule is the prompt-level implementation of Task Specific Memory in
Section~\ref{sec:method:harness}. It lets the planner transfer the structure of a
successful solution while grounding all geometry in the current rollout.

\subsection{Global Memory}
  \label{app:agent_prompt:global_memory}

  Global Memory complements seed 0 Task Specific Memory. Task Specific Memory is task-specific
  procedural context: it records the JSON audit and JSONL command trace for one
  reference rollout. Global Memory is task-independent. It stores reusable success rules
  and failure models for the fixed primitive library, including known VLA operating
  conditions, empty-grasp failures, false visual success, unstable staging, and recovery
  patterns. It is used as contextual guidance during closed-loop execution, not as an
  action trace to replay.

  \begin{samepage}
  \begin{lstlisting}[frame=single,framerule=0.5pt]
  Use Global Memory to check:
  1. known success rules for VLA and analytic primitives;
  2. known failure models before repeating or repairing an action;
  3. empty grasps, wrong-object attempts, false visual success, and
     unstable staging;
  \end{lstlisting}
  \end{samepage}

\subsection{Benchmark-Specific Prompt Instantiations}
\label{app:agent_prompt:instantiations}

Tables~\ref{tab:app_prompt_instantiations} and
\ref{tab:app_prompt_env_content} summarize how the shared prompt is instantiated for
each benchmark. We separate the instantiation into interface-level fields and
environment-context fields. The former specifies the success predicate, frozen VLA
entry point, and required audit artifacts; the latter records the embodiment and
control assumptions that specialize the shared prompt for each benchmark.

 \begin{table}[h]
  \centering
  \caption{Interface-level prompt instantiations across benchmarks. In the main text,
  the heterogeneous VLA primitive names are abstracted as the unified
  \textsc{vla\_act} interface.}
  \label{tab:app_prompt_instantiations}
  \footnotesize
  \setlength{\tabcolsep}{3pt}
  \begin{tabular}{@{}p{0.18\linewidth}p{0.22\linewidth}p{0.28\linewidth}p{0.22\linewidth}@{}}
  \toprule
  Benchmark & Success signal & VLA interface & Output artifacts \\
  \midrule
  LIBERO / LIBERO-Pro &
  \texttt{libero\_terminated} &
  \textsc{vla\_act} &
  Command JSONL; audit JSON \\
  \midrule
  RoboCasa365 &
  \texttt{success} &
  \textsc{vla\_act} &
  Command JSONL; audit JSON \\
  \midrule
  RoboTwin C2R &
  \texttt{eval\_success} &
  \textsc{vla\_act} &
  Command JSONL; audit JSON \\
  \bottomrule
  \end{tabular}
  \end{table}

  \begin{table}[h]
  \centering
  \caption{Environment context supplied by the benchmark-specific prompts.}
  \label{tab:app_prompt_env_content}
  \footnotesize
  \setlength{\tabcolsep}{3pt}
  \begin{tabular}{@{}p{0.18\linewidth}p{0.76\linewidth}@{}}
  \toprule
  Benchmark & Environment-specific prompt specialization \\
  \midrule
  LIBERO / LIBERO-Pro &
  Single-arm tabletop manipulation; fixed agentview and moving wrist RGB-D/world-map
  observations; agent-side transport,
  release, and visual verification after contact-rich steps. \\
  \midrule
  RoboCasa365 &
  Mobile kitchen manipulation; base motion is available for reaching distant fixtures;
  the agent re-localizes after base movement; fixture-facing staging and continuation of
  capped but progressing contact attempts are treated as part of the operating policy. \\
  \midrule
  RoboTwin C2R &
  Dual-arm manipulation; manual motion commands are bound to a specified arm; head and
  left/right wrist observations provide perception; manual primitives support
  observation refresh, non-grasp motion, release, termination, and recovery around
  contact-rich attempts. \\
  \bottomrule
  \end{tabular}
  \end{table}

 \paragraph{LIBERO / LIBERO-Pro.}
  The LIBERO-family prompt is shared by standard LIBERO and LIBERO-Pro:

  \begin{lstlisting}[frame=single,framerule=0.5pt]
  You are an LLM-in-the-loop hybrid driver for the LIBERO PRO/LIBERO
  benchmark.
  \end{lstlisting}

  This instantiation specializes the shared prompt for single-arm tabletop manipulation.
  It defines the LIBERO observation files used for perception-based grounding, including
  the fixed agentview RGB-D/world-map files and the moving wrist-camera files used for
  close-range re-localization. The unified \textsc{vla\_act} interface is used for
  contact-rich steps, including grasping and closed-loop articulated-object, button, or knob
  manipulation. After contact is established, the agent remains responsible for target
  identification, scene re-localization, free-space transport, release, and progress
  verification.

  \paragraph{RoboCasa365.}
  The RoboCasa prompt instantiates the shared structure for mobile kitchen manipulation:

  \begin{lstlisting}[frame=single,framerule=0.5pt]
  You are an LLM-in-the-loop hybrid driver for the RoboCasa365 kitchen
  benchmark.
  \end{lstlisting}

  This instantiation adds mobile-base staging to the shared manipulation loop. The agent
  grounds objects and fixtures from RGB-D/world-map observations, uses
  \texttt{navigate\_to} for coarse base placement, and uses \texttt{move\_base} for
  small local corrections. Because base motion changes the robot viewpoint and the
  relative arm workspace, the prompt emphasizes re-localization after navigation before
  continuing manipulation.

  The unified \textsc{vla\_act} interface provides the contact-rich primitive. The VLA
  instruction is the full task language, and the planner decides whether to invoke it
  after local staging or during broader whole-body interaction. A capped but
  still-progressing VLA call is handled as a continuation case rather than an immediate
  failure, so the same VLA call can be
  continued instead of being interrupted by manual commands.

  \paragraph{RoboTwin C2R.}
  The RoboTwin prompt instantiates the shared core for dual-arm manipulation:

  \begin{lstlisting}[frame=single,framerule=0.5pt]
  You are an LLM-in-the-loop hybrid manipulation agent for the RoboTwin
  benchmark.
  \end{lstlisting}

  This instantiation specializes the prompt for a dual-arm setting. Manual motion
  commands include explicit left/right arm binding, and the observation stream includes
  head and left/right wrist views with corresponding depth and world-map files. The
  driver reports the official RoboTwin success signal through \texttt{eval\_success} and
  writes a terminal \texttt{final.json} when the rollout exits.

  The contact-rich interface is exposed as \textsc{vla\_act}. This primitive is used
  for grasp formation, re-grasping, handover grasps, bimanual grasp
  formation, and other contact-rich phases. Manual primitives are used around these VLA
  attempts for observation refresh, non-grasp motion, release, termination, and recovery.
  RoboTwin additionally requires a diagnosis Markdown file for post-hoc analysis.

\subsection{Compact Prompt Skeleton}
\label{app:agent_prompt:skeleton}

For completeness, the following listing shows the compact shared skeleton underlying
all four prompt files. The benchmark-specific prompts fill the bracketed slots with
the concrete values described above.

\begin{samepage}
\begin{lstlisting}[frame=single,framerule=0.5pt]
You are an LLM-in-the-loop hybrid manipulation agent for {BENCHMARK}.
The benchmark driver is already running in {WORKDIR}. Complete the task
by reading state and perception files, localizing task entities from RGB
and world maps, invoking only the allowed primitives, using the frozen VLA
for contact-rich behavior, and writing a reproducible audit.

1. ROLE AND SUCCESS SIGNAL
You are a closed-loop controller. Optimize {SUCCESS_SIGNAL}, not a visual
guess. Continue until success, budget exhaustion, or unrecoverability.


2. PERCEPTION ISOLATION
Do not query simulator object poses or hidden task initialization. Use
RGB for semantic identity and depth/world maps for metric localization.
Re-localize after every object, camera, robot, base, or grasp change.

3. FILE-BASED REPL
Write one JSON command to {WORKDIR}/command.json. Wait for the driver
result. Read state_NN.json, log_NN.json, images, depth maps, and world
maps. Then decide the next command.

4. PRIMITIVE VOCABULARY
Use only {PRIMITIVE_SCHEMAS}. Preserve exact syntax and controller
semantics, including gripper sign, arm binding, chunk budgets, and step
costs.

5. DIVISION OF LABOR BETWEEN YOU AND THE VLA
Use {VLA_PRIMITIVE} for grasping, re-grasping, articulated contact,
insertion, pressing, seating, and other contact-rich phases. Use analytic
primitives for grounding, staging, free-space transport, release,
verification, and recovery.

6. TASK LANGUAGE
Read task_language from the state file. It is authoritative. Do not infer
the task from filenames, object lists, task indices, or neighboring
Task Specific Memory files.

7. SEED 0 TASK SPECIFIC MEMORY AND GLOBAL MEMORY
Read the task-matched seed 0 Task Specific Memory audit JSON to understand
why the task-specific strategy worked and what failed. Read the seed 0
Task Specific Memory JSONL to recover what was executed and in what
order. Read Global Memory for cross-task success rules and failure models.
Use Task Specific Memory as the task-specific procedural skeleton, but use
Global Memory and current perception to decide when to re-ground, verify,
recover, or stop. Never replay literal coordinates.

8. CLOSED-LOOP VERIFICATION AND RECOVERY
After every command, inspect state, logs, RGB, and world maps. Diagnose
wrong-object selection, poor stance, VLA miss, short placement, hidden
predicate failure, or unrecoverable displacement before acting again.

9. BUDGET, RESET, AND TERMINATION
Track the benchmark budget and reset policy. Do not reset in strict
evaluation. Stop only on success, budget exhaustion, or unrecoverability.

10. OUTPUT DISCIPLINE
Write the required audit JSON and command-trace JSONL. The audit JSON
records the benchmark success status and the final outcome fields used
for evaluation and success-rate computation. The JSONL trace records the
agent's executed primitive commands in order, enabling inspection and
analysis of the agent's decision process.

11. OPERATING LOOP
Read prompt, state, task language, perception, Task Specific Memory, and
Global Memory. Localize entities. Execute one primitive. Observe. Recover.
Repeat. Write outputs.
\end{lstlisting}
\end{samepage}

This structured prompt design keeps the common Harness VLA operating protocol separate from benchmark-specific assumptions, providing a reusable template for constructing agent prompts in additional manipulation benchmarks.

\section{Primitive Usage Statistics}
\label{app:primitive_stats}

Table~\ref{tab:primitive_usage_by_env} aggregates the primitive calls issued by Harness VLA (CC) across LIBERO Pro-family, RoboTwin C2R, and RoboCasa365 runs. We report canonical primitive names following the taxonomy and availability summary in Table~\ref{tab:app_primitive_availability}: all backend-specific VLA calls are merged into the unified \textsc{vla\_act} primitive, implementation-level motion macros are folded into their exposed analytic primitive, and non-manipulation helpers such as rendering, reset, notes, and no-ops are excluded. Percentages are computed within each environment's total manipulation-primitive calls.

The usage pattern supports the intended asymmetric decomposition. In LIBERO, analytic primitives dominate: \textsc{move\_to} alone accounts for $61.8\%$ of calls, while \textsc{vla\_act} accounts for $15.8\%$. This matches the tabletop structure of the tasks: the VLA is used primarily to establish contact-rich grasps or fixture interactions, after which analytic transport, gripper control, and release complete much of the rollout. RoboCasa365 shifts the mix toward mobile staging and longer-horizon interaction: \textsc{navigate\_to} and \textsc{move\_base} together account for $19.4\%$ of calls, while \textsc{vla\_act} rises to $35.3\%$ because kitchen tasks require learned grasps, fixture actuation, and constrained placements across larger scenes. RoboTwin C2R has the highest VLA share ($47.4\%$), reflecting bimanual grasping and handover-like contact, yet analytic primitives still provide a slight majority of calls for planned arm motion, release, and final arrangement.

Table~\ref{tab:primitive_usage_class} summarizes the same evidence at the class level. Across all three embodiments, the frozen VLA is not used as a monolithic end-to-end controller; it is invoked as a contact-rich primitive inside a larger analytic scaffold. The exact ratio changes with embodiment and task family, but the qualitative division remains stable: analytic primitives handle reproducible geometry and staging, while \textsc{vla\_act} supplies the learned local interactions that are difficult to script.

\begin{table}[h]
\centering
\caption{Canonical primitive usage across benchmark environments. Each cell reports count and percentage of manipulation-primitive calls within that environment. A dash indicates that the primitive is not exposed in the corresponding environment.}
\label{tab:primitive_usage_by_env}
\small
\setlength{\tabcolsep}{4pt}
\resizebox{\textwidth}{!}{%
\begin{tabular}{llccc}
\toprule
\textbf{Canonical primitive} & \textbf{Kind} & \textbf{LIBERO} & \textbf{RoboTwin C2R} & \textbf{RoboCasa365} \\
\midrule
\textsc{move\_to} & Analytic composite & $6263$ ($61.8\%$) & $685$ ($40.9\%$) & $3004$ ($38.7\%$) \\
\textsc{move\_pose} & Analytic composite & $203$ ($2.0\%$) & -- & -- \\
\textsc{navigate\_to} & Analytic composite & -- & -- & $701$ ($9.0\%$) \\
\textsc{rotate\_wrist} & Analytic atomic & $44$ ($0.4\%$) & $1$ ($0.1\%$) & -- \\
\textsc{rotate\_pitch} & Analytic atomic & $58$ ($0.6\%$) & -- & $66$ ($0.8\%$) \\
\textsc{set\_gripper} & Analytic atomic & $1137$ ($11.2\%$) & $71$ ($4.2\%$) & $371$ ($4.8\%$) \\
\textsc{release} & Analytic atomic & $831$ ($8.2\%$) & $124$ ($7.4\%$) & $76$ ($1.0\%$) \\
\textsc{move\_base} & Analytic atomic & -- & -- & $808$ ($10.4\%$) \\
\textsc{vla\_act} & VLA & $1598$ ($15.8\%$) & $794$ ($47.4\%$) & $2746$ ($35.3\%$) \\
\midrule
\textbf{Total} & & \textbf{$10134$ ($100.0\%$)} & \textbf{$1675$ ($100.0\%$)} & \textbf{$7772$ ($100.0\%$)} \\
\bottomrule
\end{tabular}
}
\end{table}

\begin{table}[h]
\centering
\caption{Primitive usage grouped by class. Analytic primitives include both composite goal-reaching controllers and atomic set-point commands.}
\label{tab:primitive_usage_class}
\small
\setlength{\tabcolsep}{12pt}
\begin{tabular}{lccc}
\toprule
\textbf{Primitive class} & \textbf{LIBERO} & \textbf{RoboTwin C2R} & \textbf{RoboCasa365} \\
\midrule
Analytic primitives & $8536$ ($84.2\%$) & $881$ ($52.6\%$) & $5026$ ($64.7\%$) \\
VLA primitive, \textsc{vla\_act} & $1598$ ($15.8\%$) & $794$ ($47.4\%$) & $2746$ ($35.3\%$) \\
\bottomrule
\end{tabular}
\end{table}

\section{Flash Mode: Generation and Selection}
\label{app:flash_mode}

\paragraph{Candidate generation.}
For each task, we compile one card from every available successful agent trajectory. The compiler verifies the benchmark success flag and matches the audit and command trace by suite, task, and seed. It parses the instruction into object relations, aligns them with ordered acquisition, release, and articulated-object interactions, and retains the primitive sequence, policy prompts, and non-spatial arguments. Saved segmentation observations supply reference anchors; missing anchors are reconstructed from the task relations and manipulation transactions. Each candidate consists of an action plan and an anchor specification. Compilation processes one trajectory; candidate evaluation is a separate stage.

\paragraph{Spatial binding and replay.}
In the LIBERO backend, a \textsc{move\_to} or \textsc{move\_pose} waypoint within $0.20$\,m of an anchor in the horizontal plane can be stored relative to that anchor. For reference waypoint $\mathbf{x}^{\mathrm{ref}}_i\in\mathbb{R}^2$ and anchor $\mathbf{a}^{\mathrm{ref}}_{j(i)}$, replay uses
\begin{equation}
    \Delta\mathbf{x}_i=\mathbf{x}^{\mathrm{ref}}_i-\mathbf{a}^{\mathrm{ref}}_{j(i)},
    \qquad
    \mathbf{x}^{\mathrm{run}}_i=\widehat{\mathbf{a}}_{j(i)}+\Delta\mathbf{x}_i,
    \label{eq:flash_anchor_binding}
\end{equation}
where $\widehat{\mathbf{a}}_{j(i)}$ is obtained by identifying the anchor in an RGB image with a VLM and back-projecting its image location using depth and camera calibration. Recorded height and non-spatial arguments are preserved; unanchored waypoints retain their coordinates. Localization can be refined from a wrist image, with a held-object offset correction when available. Execution allows at most three grasp attempts, skips dependent transport after an unconfirmed grasp, and stops if a required anchor cannot be localized. No scene reset occurs within the rollout. These binding and retry rules describe the LIBERO implementation; hardware procedures use the corresponding robot interfaces.

\paragraph{Applicable tasks.}
Flash Mode assumes a fixed subgoal order, objects and destinations that can be localized directly, and failures recoverable by the predefined checks and retries. Sequence length alone does not determine applicability: a long fixed routine can be replayed, whereas even a short task requiring a new semantic decision or an unanticipated recovery needs online planning.

\paragraph{Max-card selection.}
Let $\mathcal{C}_{\tau}$ be the cards compiled from all successful runs of task $\tau$, and $\mathcal{S}_{\tau}$ its ten-seed batch. For benchmark outcome $Y(c,s)\in\{0,1\}$, we select
\begin{equation}
    \widehat{p}_{\tau}(c)=\frac{1}{|\mathcal{S}_{\tau}|}
       \sum_{s\in\mathcal{S}_{\tau}}Y(c,s),
    \qquad
    c^{\star}_{\tau}\in\operatorname*{arg\,max}_{c\in\mathcal{C}_{\tau}}
       \widehat{p}_{\tau}(c).
    \label{eq:flash_max_selection}
\end{equation}
One card $c^{\star}_{\tau}$ is fixed across all seeds of task $\tau$. The same seeds are used for selection and reporting: Flash Mode (max) is therefore a retrospective best-card result, not an independent held-out evaluation, and can be optimistic relative to fresh layouts. Across the $80$ task identities in LIBERO-Pro, $78$ have cards; \texttt{goal\_swap\_t0} and \texttt{10\_swap\_t9} have no successful source trace and contribute $0/10$. Aggregating all $800$ episodes gives $581$ successes ($72.63\%$). Both compilation and selection occur offline with frozen VLA weights.

\paragraph{Comparison and timing protocol.}
Table~\ref{tab:flash_libero_pro} aggregates the public RPent Flash comparison records, whose Codex runs are a separate batch from Table~\ref{tab:libero_pro_perturbation}. All three success rates use $800$ episodes. Codex time averages the ten-seed mean planner durations over $80$ tasks, giving $529.23$\,s with high reasoning and $344.17$\,s without reasoning. Flash time averages the $78$ nonempty source-episode tool-duration entries, giving $56.17$\,s; the two published zero entries are retained, and the two missing plans have no timing entry. Model and service startup are excluded.

\section{Additional Planner Results: GPT-6 Astra}
\label{app:astra_results}

We follow the experimental setup in Section~\ref{sec:exp:setup}, using GPT-6 Astra as the Codex planner at low reasoning effort on LIBERO-Pro and RoboCasa365. Tables~\ref{tab:astra_libero_pro} and~\ref{tab:astra_robocasa} report the additional results, with the main-text planner results included for reference.

\paragraph{LIBERO-Pro.}
Table~\ref{tab:astra_libero_pro} reports $92.63\%$ overall success ($741/800$) with GPT-6 Astra.

\begin{table}[htbp]
\centering
\caption{LIBERO-Pro success rates (\%) with GPT-6 Astra, following the protocol and notation of Table~\ref{tab:libero_pro_perturbation}. Bold marks the best method in each column.}
\label{tab:astra_libero_pro}
\small
\setlength{\tabcolsep}{3pt}
\begin{tabular}{l cccccccc c}
\toprule
Planner & Spat-T & Spat-S & Obj-T & Obj-S & Goal-T & Goal-S & L10-T & L10-S & Overall \\
\midrule
Codex (GPT-5.5) & 81.0 & 69.0 & 94.0 & 91.0 & 75.0 & 66.0 & 52.0 & 49.0 & 72.1 \\
Claude Code (Opus-4.7) & 94.0 & 80.0 & 88.0 & 90.0 & 87.0 & 87.0 & 71.0 & 62.0 & 82.4 \\
Codex (GPT-6 Astra) & \textbf{100.0} & \textbf{98.0} & \textbf{100.0} & \textbf{99.0} & \textbf{88.0} & \textbf{99.0} & \textbf{85.0} & \textbf{72.0} & \textbf{92.63} \\
\bottomrule
\end{tabular}
\end{table}

\paragraph{RoboCasa365.}
Table~\ref{tab:astra_robocasa} reports $59.20\%$ overall success ($148/250$) with GPT-6 Astra, including $42.50\%$ on \textsc{Composite-Unseen}.

\begin{table}[htbp]
\centering
\caption{RoboCasa365 success rates (\%) with GPT-6 Astra, following the protocol of Table~\ref{tab:robocasa_main}. Overall averages over tasks. Bold marks the best method in each column.}
\label{tab:astra_robocasa}
\small
\setlength{\tabcolsep}{6pt}
\begin{tabular}{l cccc}
\toprule
Planner & Atomic-Seen & Composite-Seen & Composite-Unseen & Overall \\
\midrule
Codex (GPT-5.5) & \textbf{92.2} & \textbf{61.3} & 13.8 & 57.2 \\
Claude Code (Opus-4.7) & 80.0 & 47.5 & 15.0 & 48.8 \\
Codex (GPT-6 Astra) & 87.78 & 43.75 & \textbf{42.50} & \textbf{59.20} \\
\bottomrule
\end{tabular}
\end{table}

\end{document}